\RequirePackage{fix-cm}
\documentclass[twocolumn, natbib]{svjour3}          % twocolumn
\smartqed  % flush right qed marks, e.g. at end of proof
\usepackage{times}
\usepackage{epsfig}
\usepackage{graphicx}
\usepackage[cmex10]{amsmath}
\usepackage{amssymb}
\usepackage{tabulary,color}
\usepackage{tabularx}
\usepackage{multirow}
\usepackage{booktabs, siunitx, dcolumn}
\usepackage{colortbl}
\usepackage{pifont}
\usepackage{dsfont}
\usepackage{enumitem} % noindent for itemize
\usepackage[para]{threeparttable} % footnote for table
\usepackage[export]{adjustbox}
\usepackage[percent]{overpic}
\usepackage{array} % ALIGNMENT PACKAGES
\usepackage{url} % PDF, URL AND HYPERLINK PACKAGES
\usepackage{fixltx2e} % FLOAT PACKAGES
\usepackage{algorithm}     % algorithm
\usepackage{algorithmic}     % algorithm
 \usepackage[caption=false,font=footnotesize]{subfig}
\usepackage{natbib}
\setcitestyle{authoryear,open={(},close={)}} % cite style

\usepackage[pagebackref=true,breaklinks=true,colorlinks,citecolor=blue,bookmarks=false]{hyperref} % cite color

\usepackage[toc,page]{appendix} % appendix

\journalname{International Journal of Computer Vision}

\begin{document}\sloppy

\title{Deformable Kernel Networks for Joint Image Filtering}

\author{Beomjun~Kim$^{1}$ \and
		Jean~Ponce$^{2}$ \and 
		Bumsub~Ham$^{1}$}

\authorrunning{International Journal of Computer Vision}
\institute{Beomjun Kim \at
           \email{beomjun.kim@yonsei.ac.kr}   
           \and
           Jean Ponce \at
           \email{jean.ponce@inria.fr}   
		   \and
           Bumsub Ham (Corresponding author)\at
           \email{bumsub.ham@yonsei.ac.kr}
		   \and
	$^1$\;\; School of Electrical and Electronic Engineering, Yonsei University, Seoul, Korea. \\
	$^2$\;\; Inria and DI-ENS, D{\'e}partement d'Informatique de l'ENS, CNRS, PSL University, Paris, France. \\
}
% The e-mail address, and telephone number(s) of the corresponding author

\date{}

\maketitle

\begin{abstract}
Joint image filters are used to transfer structural details from a guidance picture used as a prior to a target image, in tasks such as enhancing spatial resolution and suppressing noise. Previous methods based on convolutional neural networks~(CNNs) combine nonlinear activations of spatially-invariant kernels to estimate structural details and regress the filtering result. In this paper, we instead learn explicitly sparse and spatially-variant kernels. We propose a CNN architecture and its efficient implementation, called the deformable kernel network~(DKN), that outputs sets of neighbors and the corresponding weights adaptively for each pixel. The filtering result is then computed as a weighted average. We also propose a fast version of DKN that runs about seventeen times faster for an image of size $640 \times 480$. We demonstrate the effectiveness and flexibility of our models on the tasks of depth map upsampling, saliency map upsampling, cross-modality image restoration, texture removal, and semantic segmentation. In particular, we show that the weighted averaging process with sparsely sampled $3 \times 3$ kernels outperforms the state of the art by a significant margin in all cases. \end{abstract}

% provide an abstract of 150 to 250 words. (177 words)

% provide 4 to 6 keywords ( 6 keywords )

\begin{figure*}[t]
  \centering
  \subfloat[RGB image.]{
    \begin{minipage}[b]{0.16\linewidth} 
      \includegraphics[width=\linewidth]{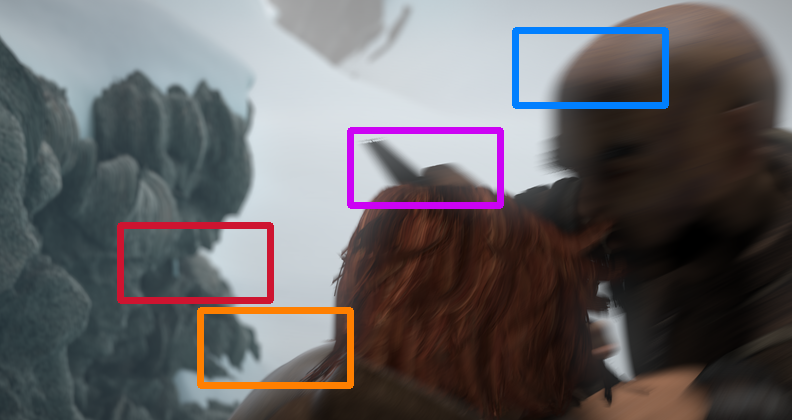}\vspace{0.03cm} \\
      \includegraphics[width=0.494\linewidth]{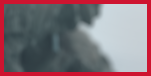}\hspace{-0.04cm}
      \includegraphics[width=0.494\linewidth]{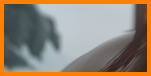} \\
      \includegraphics[width=0.494\linewidth]{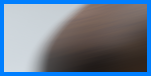}\hspace{-0.04cm}
      \includegraphics[width=0.494\linewidth]{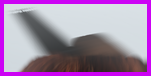}
    \end{minipage}
  }\hspace{-0.2cm}
  \subfloat[Depth image.]{
    \begin{minipage}[b]{0.16\linewidth} 
      \includegraphics[width=\linewidth]{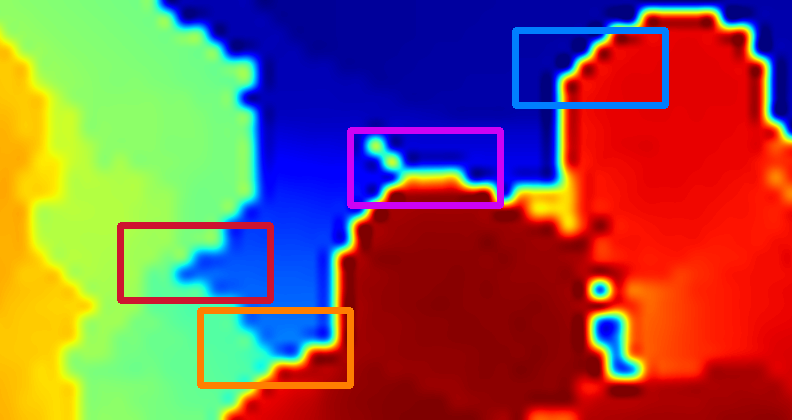}\vspace{0.03cm} \\
      \includegraphics[width=0.494\linewidth]{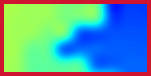}\hspace{-0.04cm}
      \includegraphics[width=0.494\linewidth]{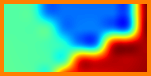} \\
      \includegraphics[width=0.494\linewidth]{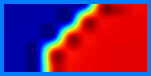}\hspace{-0.04cm}
      \includegraphics[width=0.494\linewidth]{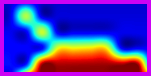}
    \end{minipage}
  }\hspace{-0.2cm}
  \subfloat[GF.]{
    \begin{minipage}[b]{0.16\linewidth} 
      \includegraphics[width=\linewidth]{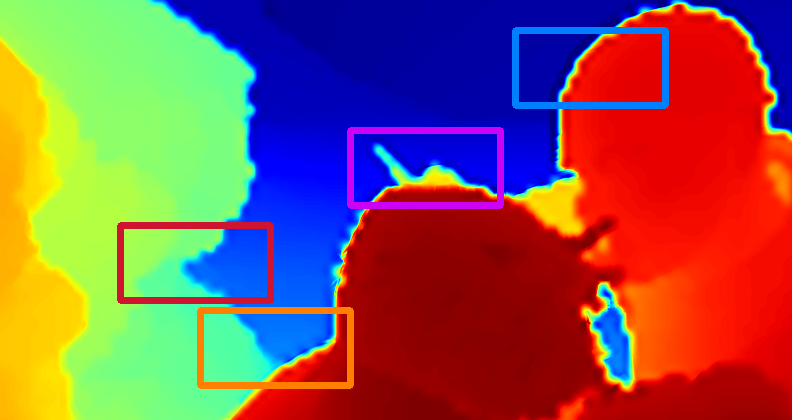}\vspace{0.03cm} \\
      \includegraphics[width=0.494\linewidth]{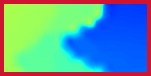}\hspace{-0.04cm}
      \includegraphics[width=0.494\linewidth]{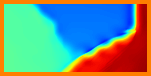} \\
      \includegraphics[width=0.494\linewidth]{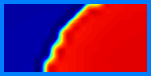}\hspace{-0.04cm}
      \includegraphics[width=0.494\linewidth]{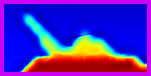}
    \end{minipage}
  }\hspace{-0.2cm}
  \subfloat[SDF.]{
    \begin{minipage}[b]{0.16\linewidth} 
      \includegraphics[width=\linewidth]{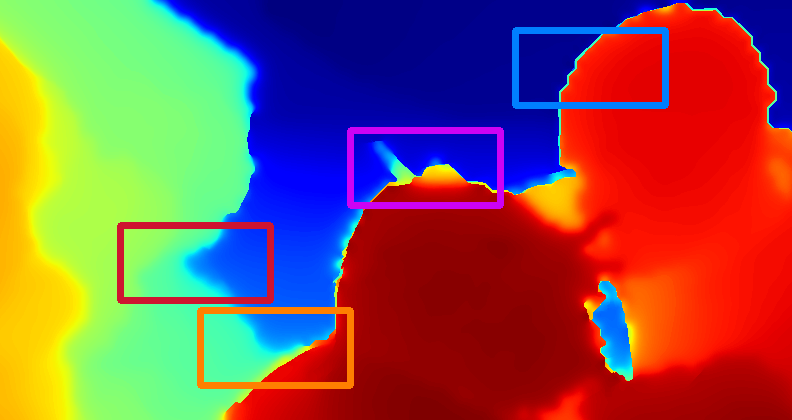}\vspace{0.03cm} \\
      \includegraphics[width=0.494\linewidth]{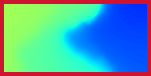}\hspace{-0.04cm}
      \includegraphics[width=0.494\linewidth]{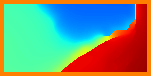} \\
      \includegraphics[width=0.494\linewidth]{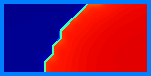}\hspace{-0.04cm}
      \includegraphics[width=0.494\linewidth]{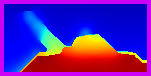}
    \end{minipage}
  }\hspace{-0.2cm}
  \subfloat[DJFR.]{
    \begin{minipage}[b]{0.16\linewidth} 
      \includegraphics[width=\linewidth]{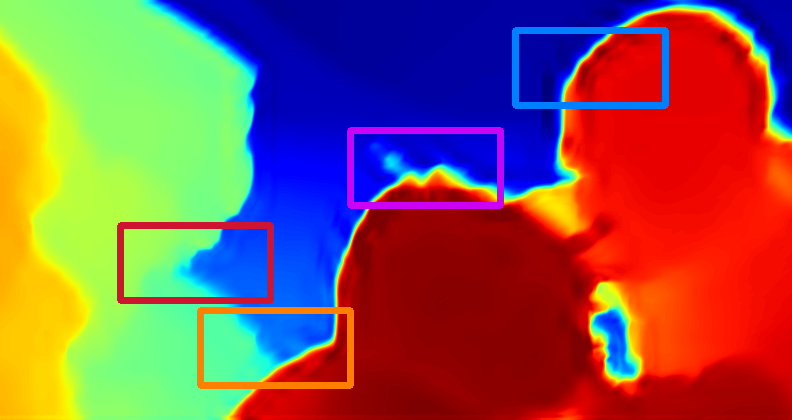}\vspace{0.03cm} \\
      \includegraphics[width=0.494\linewidth]{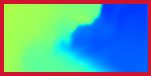}\hspace{-0.04cm}
      \includegraphics[width=0.494\linewidth]{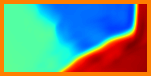} \\
      \includegraphics[width=0.494\linewidth]{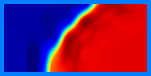}\hspace{-0.04cm}
      \includegraphics[width=0.494\linewidth]{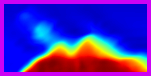}
    \end{minipage}
  }\hspace{-0.2cm}
  \subfloat[Ours~(DKN).]{
    \begin{minipage}[b]{0.16\linewidth} 
      \includegraphics[width=\linewidth]{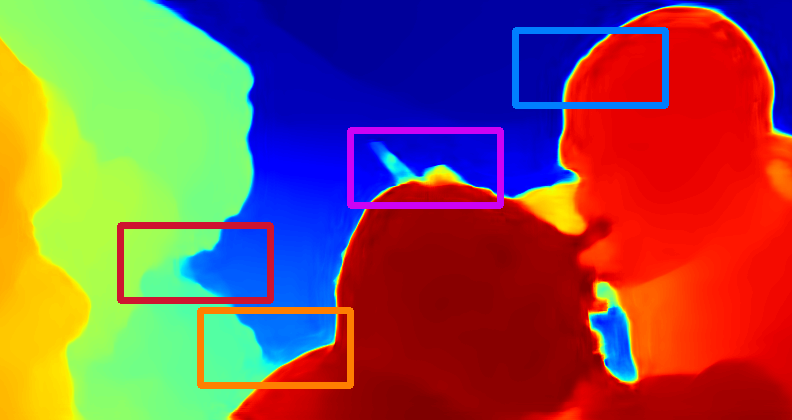}\vspace{0.03cm} \\
      \includegraphics[width=0.494\linewidth]{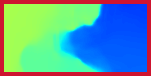}\hspace{-0.04cm}
      \includegraphics[width=0.494\linewidth]{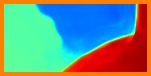} \\
      \includegraphics[width=0.494\linewidth]{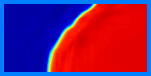}\hspace{-0.04cm}
      \includegraphics[width=0.494\linewidth]{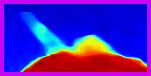}
    \end{minipage}
  }\hspace{-0.1cm}
  \begin{minipage}[b]{0.0222\linewidth}
%  \vspace*{1cm}
  \includegraphics[width=\linewidth]{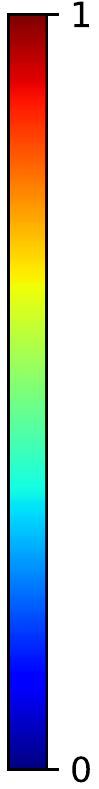}
  \end{minipage}
%\vspace{-0.2cm}
  \caption{Qualitative comparison of the state of the art and our model on depth map upsampling~($16 \times$). Given (a) a high-resolution color image and (b) a low-resolution depth image from the Sintel dataset~\citep{butler2012sintel}, we upsample the depth image using (c)~GF~\citep{he2013guided},~(d)~SDF~\citep{ham2018robust},~(e)~DJFR~\citep{li2017joint} and (f)~our method. The filtering results for GF and our model are obtained by the weighted average in~\eqref{eq:weighted_average}. We use filter kernels~$W$ of size $3 \times 3$ and $17 \times 17$ in our model and GF, respectively. We can see that our method using sparsely sampled $3 \times 3$ kernels outperforms GF and even the state of the art including the optimization-based SDF method~\citep{ham2018robust} and CNN-based one~\citep{li2017joint}. Note that applying GF with $3 \times 3$ kernels does not recover fine details.~(Best viewed in color.)}
  \label{fig:teaser}
\end{figure*}

\section{Introduction}\label{sec:introduction}
\vspace{-.2cm}
Image filtering with a guidance signal, a process called guided or joint filtering, has been used in a variety of computer vision and graphics tasks, including depth map upsampling~\citep{yang07,park2011,ferstl2013,kopf2007joint,li2016deep,ham2018robust}, cross-modality image restoration~\citep{he2013guided,shen2015mutual,yan2013cross}, texture removal~\citep{ham2018robust,xu2012structure,zhang2014rolling,karacan2013structure}, scale-space filtering~\citep{ham2018robust}, dense correspondence~\citep{ham2016proposal,hosni2013fast} and semantic segmentation~\citep{barron2016fast}. For example, high-resolution color images can be used as guidance to enhance the spatial resolution of depth maps~\citep{kopf2007joint}. The basic idea behind joint image filtering is to transfer structural details from the guidance image to the target one, typically by estimating spatially-variant kernels from the guidance. Concretely, given the target image $f$ and the guidance image $g$, the filtering output $\hat f$ at position ${\bf{p}}=(x,y)$ is expressed as a weighted average~\citep{he2013guided,kopf2007joint,tomasi1998bilateral}:
\begin{equation}\label{eq:weighted_average}
\hat f_{\bf{p}} = \sum_{{\bf{q}} \in \mathcal{N({\bf{p}})}} W_{{\bf{p}}{\bf{q}}}(f,g)f_{\bf{q}},
\end{equation}
where we denote by $\mathcal{N(\bf{p})}$ a set of neighbors (defined on a discrete regular grid) near the position $\bf{p}$. The filter kernel $W$ is a function of the guidance image $g$~\citep{park2011,ferstl2013,kopf2007joint,he2013guided}, the target image $f$ itself~\citep{zhang2014rolling,tomasi1998bilateral}, or both~\citep{li2016deep,ham2018robust}, normalized so that 
\vspace{-.1cm}
\begin{equation}\label{eq:constraint}
	\sum_{{\bf{q}} \in \mathcal{N({\bf{p}})}} W_{{\bf{p}}{\bf{q}}}(f,g) = 1.
\end{equation}

Classical approaches to joint image filtering mainly focus on designing the filter kernels $W$ and the set of neighbors $\mathcal{N}$~(i.e.,~sampling locations $\bf{q}$). They use hand-crafted kernels and sets of neighbors without learning~\citep{he2013guided,kopf2007joint,tomasi1998bilateral}. For example, the bilateral filter~\citep{tomasi1998bilateral} uses spatially-variant Gaussian kernels to encode local structures from the guidance image. The guided filter~\citep{he2013guided} also leverages the local structure of the guidance image, but uses matting Laplacian kernels~\citep{levin2008closed}, enabling a constant processing time. These filters use regularly sampled neighbors for aggregating pixels, and do not handle inconsistent structures in the guidance and target images~\citep{ham2018robust}. This causes texture-copying artifacts, especially in the case of data from different sensors~\citep{ferstl2013}. To address this problem, the SD filter~\citep{ham2018robust} constructs spatially-variant kernels from both guidance and target images to exploit common structures, and formulates joint image filtering as an optimization problem. The DG filter~\citep{Gu2017learning} uses a task-driven learning method to obtain the optimized guidance images tailored to depth upsampling. This type of approaches~(e.g.,~\citep{ferstl2013,xu2012structure,farbman2008edge}) computes a filtering output by optimizing an objective function that involves solving a large linear system. This is equivalent to filtering an image by an inverse matrix~\citep{he2013guided}, whose rows correspond to a filter kernel, leveraging global structures in the guidance image. Optimization-based methods can be considered as implicit weighted-average filters. Learning-based approaches using convolutional neural networks~(CNNs)~\citep{li2016deep,hui2016depth,xu2015deep} are also becoming increasingly popular. The networks are trained using large quantities of data, reflecting natural image priors and often outperforming traditional methods by large margins. These methods do not use a weighted averaging process with spatially-variant kernels as in \eqref{eq:weighted_average}. They combine instead nonlinear activations of spatially-invariant kernels learned from the networks. That is, they approximate spatially-variant kernels by mixing the activations of spatially-invariant ones nonlinearly~(e.g., via the ReLU function~\citep{krizhevsky2012imagenet}). 

In this paper, we revisit the guided weighted average framework in~\eqref{eq:weighted_average} for joint image filtering. We argue that leveraging spatially-invariant kernels in CNN-based methods~\citep{li2016deep,hui2016depth,xu2015deep} is limited to encoding structural details from  guidance and target images that typically change with image location. We exploit instead spatially-variant kernels explicitly, as in the classical approaches using the weighted average, but learn the kernel weights~($W$) and the set of neighbors~($\mathcal{N}$) in a completely data-driven way, building an adaptive and sparse neighborhood system for each pixel, which may be difficult to design by hand. To implement this idea, we propose a CNN architecture and its efficient implementation, called a \emph{deformable kernel network}~(DKN), for learning sampling locations of the neighboring pixels and their corresponding kernel weights at every pixel. We also propose a fast version of DKN~(FDKN), achieving a $17 \times$~speed-up compared to the plain DKN for an image of size $640 \times 480$, while retaining its superior performance. 
 We show that the weighted averaging process using sparsely sampled $3 \times 3$ kernels is sufficient to obtain a new state of the art in a variety of applications, including depth map upsampling~(Figs.~\ref{fig:teaser} and~\ref{fig:depth_upsampling}), saliency map upsampling~(Fig.~\ref{fig:saliency}), cross-modality image restoration~(Fig.~\ref{fig:noise-reduction}), texture removal~(Fig.~\ref{fig:texture_removal}), and semantic segmentation~(Fig.~\ref{fig:segmentation}).

\noindent {\textbf{Contributions.}} The main contributions of this paper can be summarized as follows:
\vspace{-.2cm}
\begin{itemize}[leftmargin=*]
  \item[$\bullet$] We introduce a novel variant of the classical guided weighted averaging process for joint image filtering and its implementation, the DKN, that computes the set of nonlocal neighbors and their corresponding weights adaptively for individual pixels~(Section~\ref{sec:proposed}).
  \item[$\bullet$] We propose a fast version of DKN~(FDKN) that runs about seventeen times faster than the DKN while retaining its superior performance~(Section~\ref{sec:proposed}).
  \item[$\bullet$] We achieve a new state of the art on several tasks, clearly demonstrating the advantage of our approach to learning both kernel weights and sampling locations~(Section~\ref{sec:exp}). We also provide an extensive experimental analysis to investigate the influence of all the components and parameters of our model~(Section~\ref{sec:discussion}). 
\end{itemize}
\vspace{-.2cm}
To encourage comparison and future work, our code and models are available at our project webpage\footnote{\url{https://cvlab.yonsei.ac.kr/projects/dkn}}.

\vspace{-.4cm}
\section{Related work}
\vspace{-.3cm}
Here we briefly describe the context of our approach, and review representative works related to ours. 
\vspace{-.6cm}
\subsection{Joint image filtering}
\vspace{-.2cm}
We categorize joint image filtering into explicit/implicit weighted-average methods and learning-based ones. 

First, explicit joint filters compute the output at each pixel by a weighted average of neighboring pixels in the target image, where the weights are estimated from the guidance and/or target image~\citep{kopf2007joint,he2013guided,zhang2014rolling}. The bilateral~\citep{tomasi1998bilateral} and guided~\citep{he2013guided} filters are representative methods that have been successfully adapted to joint image filtering. They use hand-crafted kernels to transfer fine-grained structures from the guidance image. It is, however, difficult to manually adapt the kernels to new tasks, and these methods may transfer erroneous structures to the target image~\citep{li2016deep}. 

Second, implicit weighted-average methods formulate joint filtering as an optimization problem, and minimize an objective function that usually involves fidelity and regularization terms~\citep{park2011,ferstl2013,ham2018robust,xu2012structure,farbman2008edge,xu2011image}. The fidelity term encourages the filtering output to be close to the target image, and the regularization term, typically modeled using a weighted L2 norm~\citep{farbman2008edge}, gives the output having a structure similar to that of the guidance image. Although, unlike explicit ones, implicit joint filters exploit global structures in the guidance image, hand-crafted regularizers may not reflect structural priors in the guidance image. Moreover, optimizing the objective function involves solving a large linear system, which is time consuming, even with preconditioning~\citep{szeliski2006locally} or multigrid methods~\citep{farbman2008edge}. 

Finally, learning-based methods can further be categorized into dictionary-~and CNN-based approaches. Dictionary-based methods exploit the relationship between paired target patches~(e.g., low-~and high-resolution patches for upsampling), additionally coupled with the guidance image~\citep{yang2010image,ferstl2015variational}. In CNN-based methods~\citep{hui2016depth,li2016deep,li2017joint}, an encoder-decoder architecture is used to learn features from the target and guidance images, and the filtering output is then regressed directly from the network. Learning-based techniques require a large number of ground-truth images for training. Other methods~\citep{riegler16gdsr,riegler16dsr} integrate a variational optimization into CNNs by unrolling the optimization steps of the primal-dual algorithm, which requires two stages in training and a number of iterations in testing. Similar to implicit weighted-average methods, they use hand-crafted regularizers, which may not capture structural priors.

Our method borrows from both explicit weighted-average methods and CNN-based ones. Unlike existing explicit weighted-average methods~\citep{he2013guided,kopf2007joint}, that use hand-crafted kernels and neighbors defined on a fixed regular grid, we leverage CNNs to learn the set of sparsely chosen neighbors and their corresponding weights adaptively. Our method differs from previous CNN-based ones~\citep{hui2016depth,li2016deep,li2017joint} in that we learn sparse and spatially-variant kernels for each pixel to obtain upsampling results as a weighted average. The bucketing stretch in single image super-resolution~\citep{getreuer2018blade,romano2017raisr} can be seen as a non-learning-based approach to filter selection. It assigns a single filter by solving a least-squares problem for a set of similar patches (buckets). In contrast, our model \emph{learns} different filters using CNNs even for similar RGB patches, since we learn them from a set of multi-modal images (i.e.,~pairs of RGB/D images).

\begin{figure*}
\centering
\includegraphics[width=0.95\linewidth]{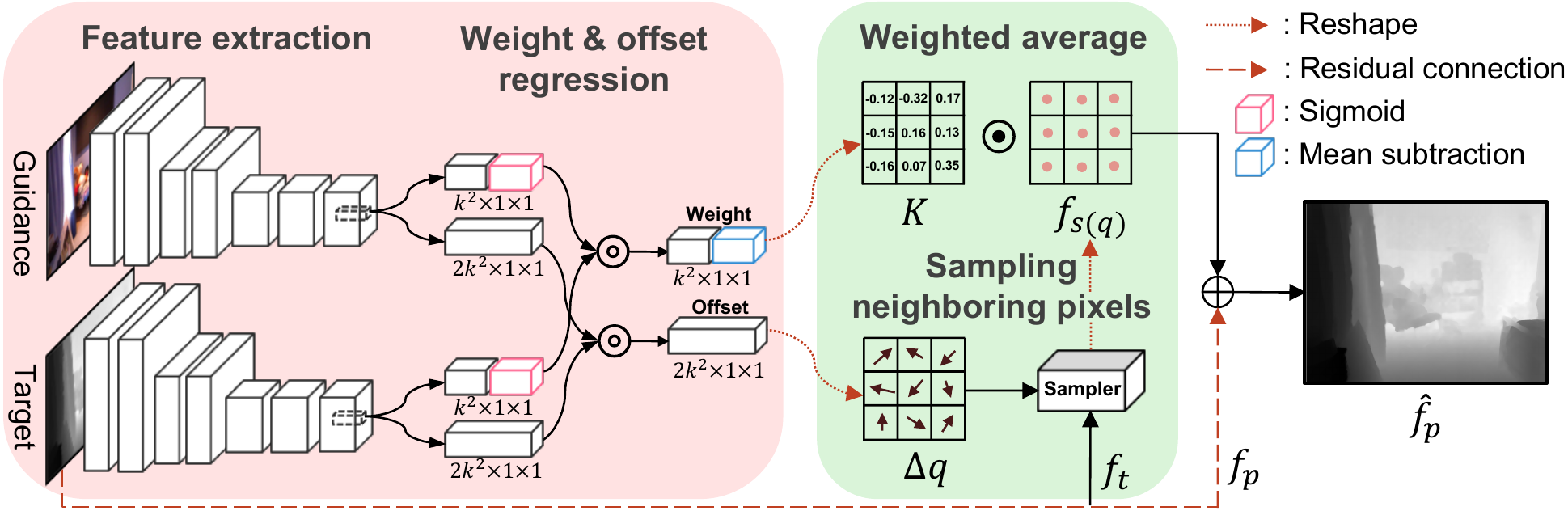}
\caption{The DKN architecture. We learn the kernel weights~$K$ and the spatial sampling offsets~$\Delta {\bf{q}}$ from the feature maps of guidance and target images. To obtain the residual image~$\hat f_{\bf{p}} -f_{\bf{p}}$, we then compute the weighted average with the kernel weights~$K$ and image values~$f_{{\bf{s}}({\bf{q}})}$ sampled at offset locations~$\Delta {\bf{q}}$ from the neighbors~$f_{{\bf{t}}}$. Finally, the result is combined with the target image~$f_{{\bf{p}}}$ to obtain the filtering result~$\hat f_{\bf{p}}$. Our model is fully convolutional and is learned end-to-end. We denote by $\circledcirc$ and $\odot$~element-wise multiplication and dot product, respectively. The reshaping operator and residual connection are drawn in dotted and dashed lines, respectively. See Table~\ref{table:architecture} for the detailed description of the network structure. (Best viewed in color.) }
\label{fig:overview}
\end{figure*}

\vspace{-.5cm}
\subsection{Variants of the spatial transformer}
\vspace{-.2cm}
Recent works introduce more flexible and effective CNN architectures. Jaderberg et al. propose a novel learnable module, the spatial transformer~\citep{jaderberg2015spatial}, that outputs the parameters of the desired spatial transformation~(e.g., affine or thin plate spline)~given a feature map or an input image. The spatial transformer makes a standard CNN network for classification invariant to a set of geometric transformation, but it has a limited capability of handling local transformations. Choy et al. introduce a convolutional version of the spatial transformer~\citep{choy2016universal}. They learn local transformation parameters for normalizing orientation and scale in feature matching. 

Most similar to ours are the dynamic filter network~\citep{jia2016dynamic} and its variants~(the adaptive convolution network~\citep{niklaus2017video} and the kernel prediction networks~\citep{bako2017kernel,mildenhall2018burst,vogels2018denoising}), where a set of local transformation parameters is generated adaptively conditioned on the input image. The main differences between our model and these works are three-fold. First, our network is not limited to learning spatially-variant kernels, but it also learns the sampling locations of neighbors. This allows us to aggregate sparse but highly related samples only, enabling an efficient implementation in terms of speed and memory and achieving state-of-the-art results even with kernels of size~$3 \times 3$ on several tasks. For comparison, the adaptive convolution and kernel prediction networks require much larger neighbors~(e.g.,~$21 \times 21$ in~\citep{bako2017kernel,vogels2018denoising}, $41 \times 41$ in~\citep{niklaus2017video}, and $8 \times 5 \times 5$ in~\citep{mildenhall2018burst}). As will be seen in our experiments, learning sampling locations of neighbors clearly boosts the performance significantly compared to learning kernel weights only. Second, our model is generic and generalizes well to other tasks. It can handle multi-modal images~(e.g.,~RGB/D images in depth map upsampling and an RGB image/a cost volume storing the costs for choosing labels in semantic segmentation), and thus is applicable to various tasks including depth and saliency map upsampling, cross-modality image restoration, texture removal, and semantic segmentation. In contrast, the adaptive convolution network is specialized to video frame interpolation, and kernel prediction networks are applicable to denoising Monte Carlo renderings~\citep{bako2017kernel,vogels2018denoising} or burst denoising~\citep{mildenhall2018burst} only. Finally, our model learns spatially-variant kernels to compute residual images, not a final output as in~\citep{bako2017kernel,jia2016dynamic,mildenhall2018burst,niklaus2017video,vogels2018denoising}, with constraints on weight regression. This allows the use of residual connections for adaptive convolution and kernel prediction networks, and achieves better results. 

Our work is also related to the deformable convolutional network~\citep{dai2017deformable}. The basic idea of deformable convolutions is to add  offsets to the sampling locations defined on a regular grid in standard CNNs. The deformable convolutional network samples features directly from learned offsets, but shares the same weights for different sets of offsets as in standard CNNs. In contrast, we use spatially-variant weights for each sampling location. Another difference is that we use the learned offset explicitly to obtain the final result, while the deformable convolutional network uses it to compute intermediate feature maps. 

\begin{table*}[t]
\centering
\caption{Network architecture details. ``BN" and ``Res." denote the batch normalization~\citep{ioffe2015batch} and residual connection, respectively. We denote by ``DownConv" convolution with stride 2. The inputs of our network are 3-channel guidance and 1-channel target images~(denoted by $D$). For the model without the residual connection, we use an L1 normalization layer~(denoted by ``L1 norm.") instead of subtracting mean values for weight regression. We apply a zero padding technique to input images to handle boundary pixels.}
\label{table:architecture}
\addtolength{\tabcolsep}{-3.0pt}
\renewcommand{\arraystretch}{0.0}
\newcolumntype{L}[1]{>{\raggedright\arraybackslash}p{#1}}
\newcolumntype{C}[1]{>{\centering\arraybackslash}p{#1}}
\newcolumntype{R}[1]{>{\raggedleft\arraybackslash}p{#1}}
\begin{tabular}{L{4.5cm} R{2.5cm} L{4.5cm} R{2.5cm}}
\midrule
\multicolumn{2}{c}{Feature extraction} & \multicolumn{2}{c}{Weight regression}\\
\cmidrule(lr){1-2}
\cmidrule(lr){3-4}
\multicolumn{1}{c}{Type} & \multicolumn{1}{c}{Output} & \multicolumn{1}{c}{Type} & \multicolumn{1}{c}{Output} \\ 
\cmidrule(lr){1-1}
\cmidrule(lr){2-2}
\cmidrule(lr){3-3}
\cmidrule(lr){4-4}
Input & $ D \times 51 \times 51$ & Conv($1\times1$) & $k^2 \times 1 \times 1$ \\
\addlinespace[0.5em]
Conv($7\times7$)-BN-ReLU & $32\times 45 \times 45$ & Sigmoid & $k^2 \times 1 \times 1$\\
\addlinespace[0.5em]
DownConv($2\times2$)-ReLu & $ 32 \times 22 \times 22$ & Mean subtraction or & \multirow{2}{*}{$k^2 \times 1 \times 1$} \\
\addlinespace[0.5em]
Conv($5\times5$)-BN-ReLU & $ 64 \times 18 \times 18$ & L1 norm. (w/o Res.) \\

\cmidrule(lr){3-4} 
DownConv($2\times2$)-ReLU & $ 64 \times 9 \times 9$ &\multicolumn{2}{c}{Offset regression} \\
\cmidrule(lr){3-4} 
Conv($5\times5$)-BN-ReLU & $ 128 \times 5 \times 5$ & \multicolumn{1}{c}{Type} & \multicolumn{1}{c}{Output} \\
\cmidrule(lr){3-3}
\cmidrule(lr){4-4} 
Conv($3\times3$)-ReLU & $ 128 \times 3 \times 3$ & \multirow{2}{*}{Conv($1\times1$)} & \multirow{2}{*}{$2k^2 \times 1 \times 1$} \\
\addlinespace[0.5em]
Conv($3\times3$)-ReLU & $ 128 \times 1 \times 1$ \\
\midrule
\end{tabular}
\end{table*}

\vspace{-.2cm}
\section{Proposed approach}\label{sec:proposed}
\vspace{-.2cm}
In this section, we briefly describe our approach to learning both kernel weights and sampling locations for joint image filtering~(Section~\ref{subsec:overview}), and present a concrete network architecture and its efficient implementation using a shift-and-stitch approach~(Section~\ref{subsec:dkn}). We then describe a fast version of DKN~(Section~\ref{subsec:fdkn}).

\vspace{-.3cm}
\subsection{Overview}\label{subsec:overview}
\vspace{-.2cm}
Our network mainly consists of two parts~(Fig.~\ref{fig:overview}): We first learn spatially-variant kernel weights and spatial sampling offsets w.r.t the regular grid. Motivated by the SD filter~\citep{ham2018robust} and DJF~\citep{li2016deep}, we extract features from individual guidance and target images. Specifically, we use a two-stream CNN~\citep{simonyan2014two}, where each sub-network is in charge of one of the two images, with different feature maps used to estimate the corresponding kernel weights and offsets. We then compute a weighted average using the learned kernel weights and sampling locations computed from the offsets to obtain a residual image. Finally, the filtering result is obtained by combining the residuals with the target image. Our network is fully convolutional, does not require fixed-size input images, and it is trained end-to-end. 

%Specifically, a two-stream CNN~\citep{simonyan2014two}, where each sub-network has the same structure~(but different parameters), takes the guidance and target images to extract feature maps that are used to estimate the kernel weights and the offsets.

\noindent {\textbf{Weight and offset learning.}} Dual supervisory information for the weights and offsets is typically not available. We learn instead these parameters by minimizing directly the discrepancy between the output of the network and a reference image~(e.g.,~a ground-truth depth map for depth upsampling). In particular, constraints on weight and offset regression~(sigmoid and mean subtraction layers in Fig.~\ref{fig:overview}) specify how the kernel weights and offsets behave and guide the learning process. For weight regression, we apply a sigmoid layer that makes all elements larger than 0 and smaller than 1. We then subtract the mean value from the output of the sigmoid layer. This makes the sum of kernel weights to be 0, so that the regressed weights act similar to a high-pass filter. For offset regression, we do not apply the sigmoid layer, since relative offsets~(for~$x$,~$y$ positions)~from locations on a regular grid can have negative values.

\noindent {\textbf{Residual connection.}} The main reason behind using a residual connection is that the filtering result is largely correlated with the target image, and both share low-frequency content~\citep{li2017joint,kim2016accurate,he2016residual,Zhang2017}. Focussing on learning the residuals also accelerates training speed while achieving better performance~\citep{kim2016accurate}. Note that contrary to~\citep{li2017joint,kim2016accurate,he2016residual,Zhang2017}, we obtain the residuals by a weighted averaging process with the learned kernels, instead of regressing them directly from the network output. Empirically, the kernels learned with the residual connection have the same characteristics as the high-pass filters widely used to extract important structures~(e.g., object boundaries)~from images. Note also that we can train DKN without the residual connection. In this case, we compute the filtering result as the weighted average in~\eqref{eq:weighted_average}.

\vspace{-.6cm}
\subsection{DKN architecture}\label{subsec:dkn}
\vspace{-.2cm}

We design a fully convolutional network to learn the kernel weights and the sampling offsets for individual pixels. A detailed description of the network structure is shown in Table~\ref{table:architecture}.  

\noindent
\textbf{Feature extraction.}
We adapt the architecture of~\citep{niklaus2017video} for feature extraction, with 7 convolutional layers. We input the guidance and target images to each of the sub-networks, which gives a feature map of size~$128 \times 1 \times 1$ for a receptive field of size $51 \times 51$. We use the ReLU~\citep{krizhevsky2012imagenet} as an activation function. Batch normalization~\citep{ioffe2015batch} is used for speeding up training and regularization. In case of joint upsampling tasks~(e.g.,~depth map upsampling), we input a high-resolution guidance image~(e.g.,~a color image) and a low-resolution target image~(e.g.,~a depth map) upsampled using bicubic interpolation.

\noindent
\textbf{Weight regression.}
For each sub-network, we add a $1 \times 1$ convolutional layer on top of the feature extraction layer. It gives a feature map of size~$k^2 \times 1 \times 1$, where $k$ is the size of the filter kernel, which is used to regress the kernel weights. To estimate the weights, we apply a sigmoid layer to each feature map of size~$k^2 \times 1 \times 1$, and then combine the outputs by element-wise multiplication~(see Fig.~\ref{fig:overview}). We could use a softmax layer as in~\citep{bako2017kernel,niklaus2017video,vogels2018denoising}, but empirically find that it does not perform as well as the sigmoid layer. The softmax function encourages the estimated kernel to have only a few non-zero elements, which is not appropriate for image filtering. The estimated kernels should be similar to high-pass filters, with kernel weights adding to 0. To this end, we subtract the mean value from the combined output of size $k^2 \times 1 \times 1$. For our model without a residual connection, we apply instead L1 normalization to the output of size $k^2 \times 1 \times 1$. Since the sigmoid layer makes all elements in the combined output larger than 0, applying L1 normalization forces the kernel weights to add to 1 as in~\eqref{eq:constraint}. 

\noindent
\textbf{Offset regression.}
Similar to the weight regression case, we add a $1 \times 1$ convolutional layer on top of the feature extraction layer. The resulting two feature maps of size~$2k^2 \times 1 \times 1$ are combined by element-wise multiplication. The final output contains relative offsets (for $x$, $y$ positions) from locations on a regular grid. In our implementation, we use $3\times 3$ kernels, but the filtering result is computed by aggregating 9 samples sparsely chosen from a much larger neighborhood. {The two main reasons behind the use of small-size kernels are as follows:~(1)~This enables an efficient implementation in terms of speed and memory. (2) The reliability of samples is more important than the total number of samples aggregated. A similar finding is noted in~\citep{wang2007optimized}, which shows that only high-confidence samples should be chosen when estimating foreground and background images in image matting. Offset regression is closely related to nonlocal means~\citep{buades2005non} in that both aggregate pixels not located at immediate neighbors. Note that learning offsets can be seen as estimating correspondences \emph{within} input images. The sampling positions computed by the offsets point to pixels whose intensity values are similar to that of the center of the kernel. That is, our model with filter kernels of size $k\times k$ computes $k^2$ matching points for each pixel. The kernel weights tell us how much information is aggregated from these matching points, indicating that the weights correspond to matching confidence. Note also that we use a larger receptive field than the filter kernel. This better handles the aperture problem occurring in finding the matching points by considering contextual information~\citep{niklaus2017video}.

\begin{figure}[t]
\vspace{-.4cm}
  \centering
  \subfloat[]{
    \begin{minipage}[b]{0.3\linewidth}
      \includegraphics[width=\linewidth]{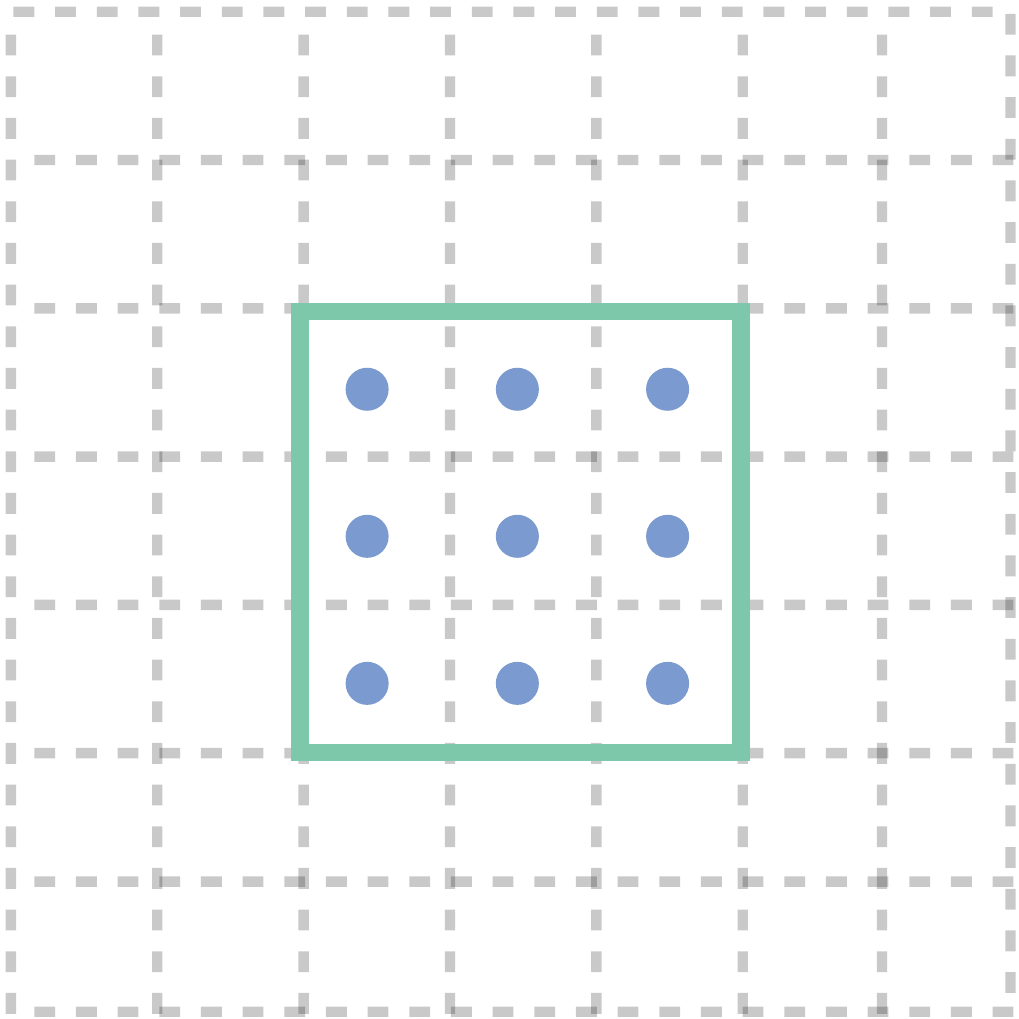} 
    \end{minipage}
  }
  \subfloat[]{
    \begin{minipage}[b]{0.3\linewidth}
      \includegraphics[width=\linewidth]{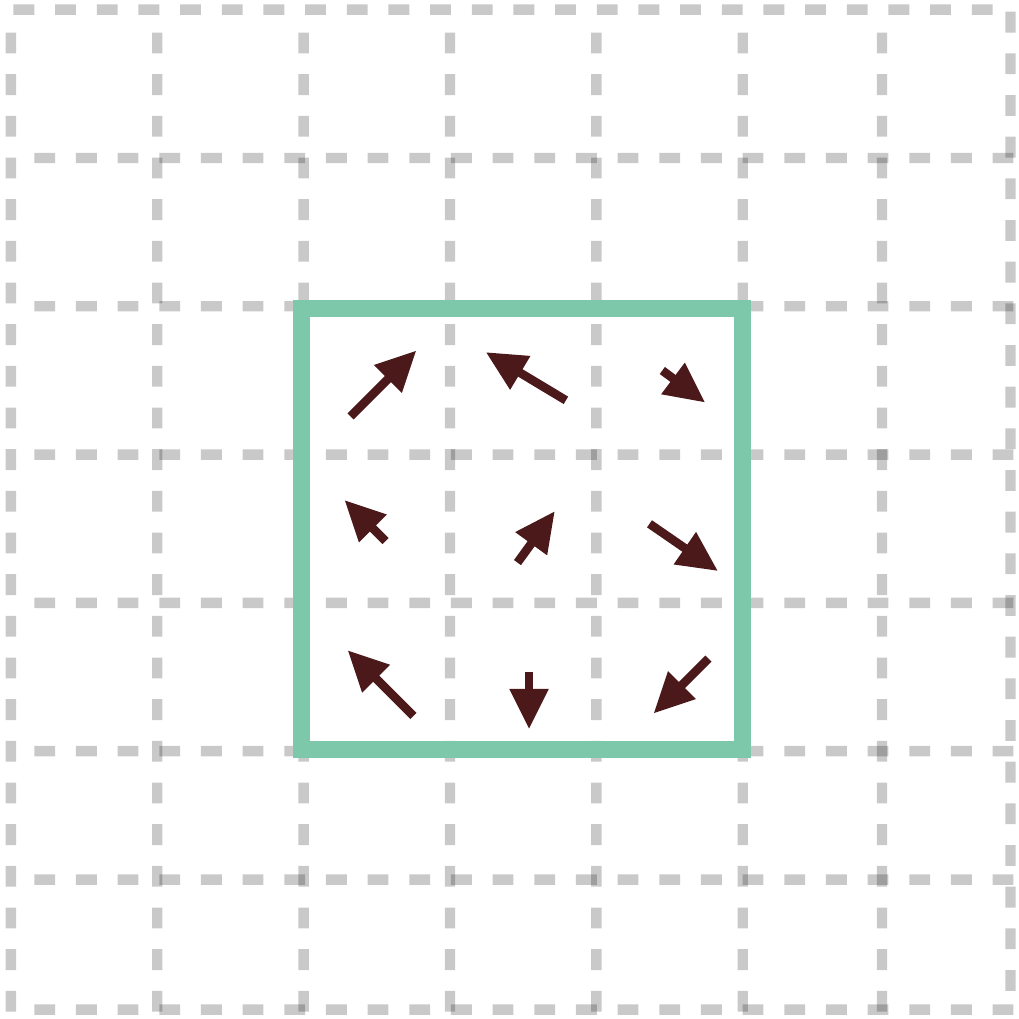} 
    \end{minipage}
  }
    \subfloat[]{
    \begin{minipage}[b]{0.3\linewidth}
      \includegraphics[width=\linewidth]{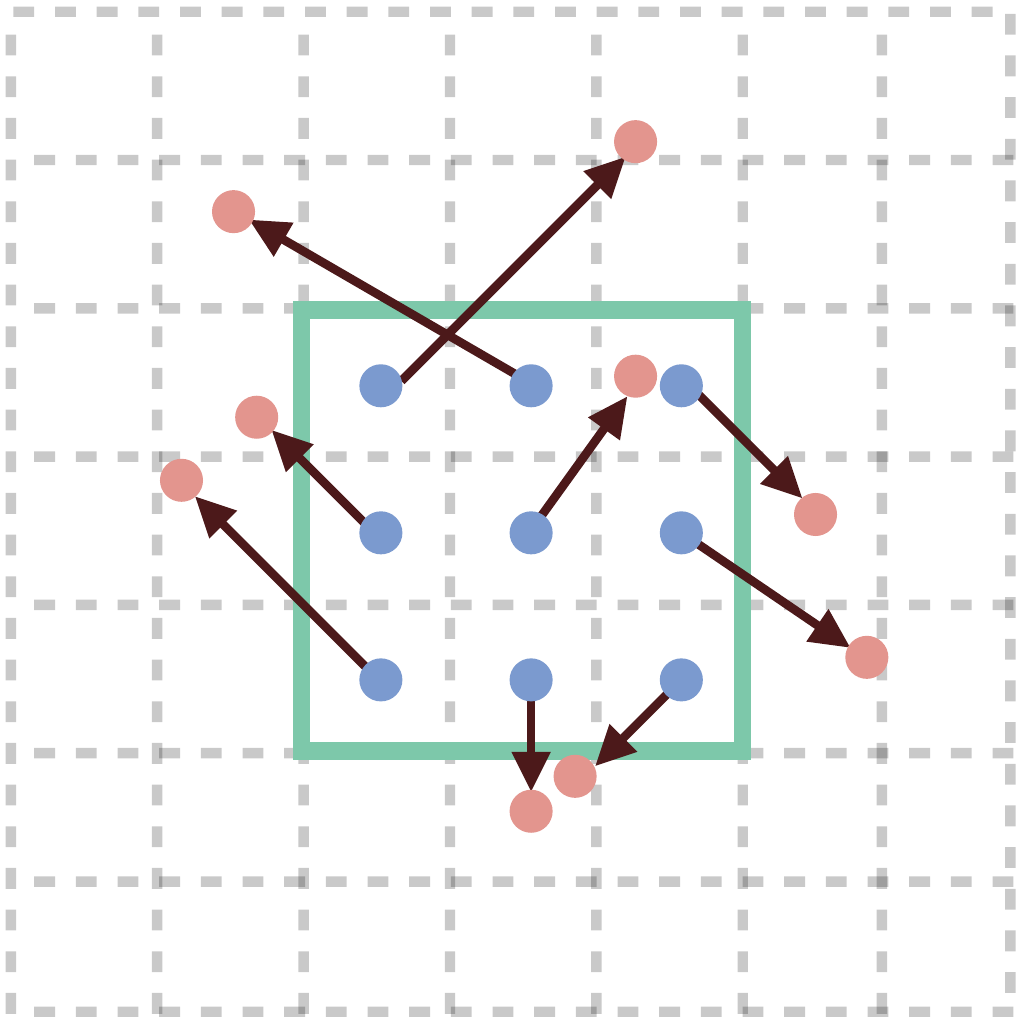}   
    \end{minipage}
  }
  \caption {Illustration of irregular sampling of neighboring pixels using offsets: (a) regular sampling~${\bf{q}}$ on discrete grid; (b) learned offsets~$\Delta {\bf{q}}$; (c) deformable sampling locations~${\bf{s}}({\bf{q}})$ with the offsets~$\Delta {\bf{q}}$. The learned offsets are fractional and the corresponding pixel values are obtained by bilinear interpolation.}
  \vspace{-.2cm}
  \label{fig:grid}
\end{figure}

\noindent
\textbf{Weighted average.}
Given the learned kernel~$K$ and sampling offsets~$\Delta {{\bf{q}}}$, we compute the residuals~$\hat f_{\bf{p}} - f_{\bf{p}}$ as a weighted average: 
\begin{equation}\label{eq:dkn}
\hat f_{\bf{p}}  =  f_{\bf{p}} + \sum_{{\bf{q}} \in \mathcal{N({\bf{p}})}} K_{{\bf{p}}{{\bf{s}}}}(f,g)f_{{\bf{s}}({\bf{q}})},
\end{equation}
where $\mathcal{N({\bf{p}})}$ is a local $3 \times 3$ window centered at the location $\bf{p}$ on a regular grid~(Fig.~\ref{fig:grid}(a)). We denote by ${{\bf{s}}}({\bf{q}})$ the sampling position computed from the offset~$\Delta {{\bf{q}}}$~(Fig.~\ref{fig:grid}(b)) of the location ${\bf{q}}$ as follows.
\begin{equation}
{\bf{s}}({\bf{q}}) = {\bf{q}} + \Delta {\bf{q}}.
\end{equation}
The sampling position ${{\bf{s}}}({\bf{q}})$ predicted by the network is irregular and typically fractional~(Fig.~\ref{fig:grid}(c)). We use bilinear interpolation~\citep{jaderberg2015spatial} to sample corresponding (sub) pixels~$f_{{\bf{s}}({\bf{q}})}$ as 
\begin{equation}
	f_{{\bf{s}}({\bf{q}})} = \sum_{{\bf{t}} \in \mathcal{R({\bf{s}}({\bf{q}}))}} G({\bf{s}}, {\bf{t}}) f_{{\bf{t}}},
\end{equation}
where $\mathcal{R({{\bf{s}}}({\bf{q}}))}$ enumerates all integer locations in a local 4-neighborhood system to the fractional position ${\bf{s}}({\bf{q}})$, and $G$ is a sampling kernel. Following~\citep{dai2017deformable,jaderberg2015spatial}, we use a two-dimensional bilinear kernel, and split it into two one-dimensional ones as
\begin{equation}
	G({\bf{s}}, {\bf{t}}) = g(s_x, t_x) g(s_y, t_y),
\end{equation}
where $g(a,b) = \operatorname{max}(0,1-|a-b|)$. Note that the residual term in~\eqref{eq:dkn} is exactly the same as the explicit joint filters in~\eqref{eq:weighted_average}, but we aggregate pixels from the sparsely chosen locations~${\bf{s}}({\bf{q}})$ with the learned kernels~$K$. Note that aggregating pixels from a fractional grid is not feasible in current joint filters including DJF~\citep{li2017joint}.

When we do not use a residual connection, we compute the filtering result~$\hat f_{\bf{p}}$ directly as a weighted average using the learned kernels and offsets:
\begin{equation}\label{eq:dkn_wo_res}
\hat f_{\bf{p}} = \sum_{{\bf{q}} \in \mathcal{N({\bf{p}})}} K_{{\bf{p}}{{\bf{s}}}}(f,g)f_{{\bf{s}}({\bf{q}})}.
\end{equation}

\noindent
\textbf{Loss.}
We train our model by minimizing the $L_1$ norm of the difference between the network output~$\hat{f}$ and ground truth $f^{\text{gt}}$ as follows.
\begin{equation}
	L(f^{\text{gt}}, \hat{f}) =  \sum_{\bf{p}}{| f^{\text{gt}}_{\bf{p}} -  \hat f_{\bf{p}} |_{1}}.
\end{equation}

\begin{figure*}[t]
\centering
\includegraphics[width=0.95\linewidth]{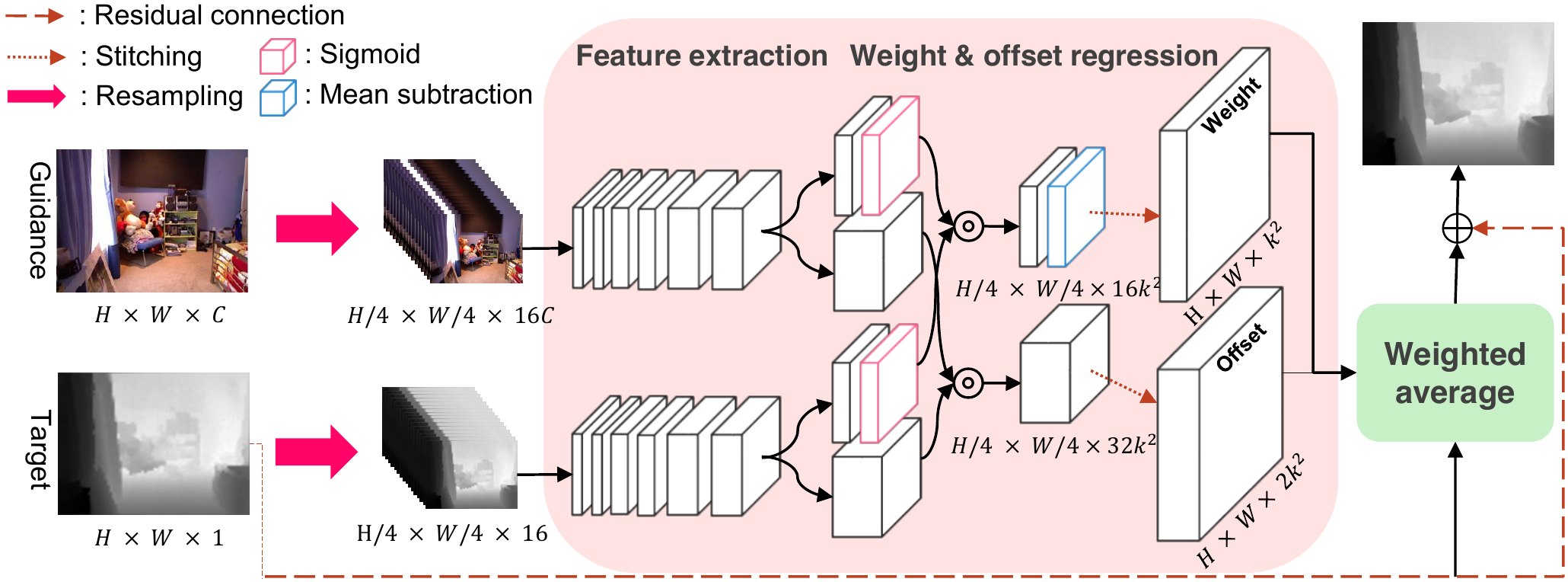}
\caption{The FDKN architecture. We resample a guidance image of size~$H \times W \times C$ with stride 4 in each dimension~(Fig.~\ref{fig:resampling}), where~$H$,~$W$ and~$C$ are height, width and the number of channels, respectively. This gives resampled guidance images of size~$H/4 \times W/4 \times 16C$. The target image is also resampled to the size of $H/4 \times W/4 \times 16$. This allows the FDKN to maintain a receptive field size comparable with the DKN. The FDKN inputs resampled guidance and target images, and then computes the kernel weight and offset locations of size~$H \times W \times k^2$ and~$H \times W \times 2k^2$, respectively, making it possible to get filtering results in a single forward pass without loss of resolution. Finally, the residuals computed by the weighted average and the target image are combined to obtain the filtering result. See Table~\ref{table:fastarchitecture} for the detailed description of the network structure. (Best viewed in color.)} 
\label{fig:fast-architecture}
\vspace{-0.3cm}
\end{figure*}

\begin{table*}[t]
\centering
\caption{FDKN architecture details. The inputs of FDKN are $16C$-channel guidance and 16-channel target images~(denoted by~$D$). Note that the receptive field of size~$13 \times 13$ in the resampled images is comparable to that of size~$51 \times 51$ in the DKN. We apply a zero padding technique to input images to handle boundary pixels.}
\label{table:fastarchitecture}
\addtolength{\tabcolsep}{-3.0pt}
\renewcommand{\arraystretch}{0.0}
\newcolumntype{L}[1]{>{\raggedright\arraybackslash}p{#1}}
\newcolumntype{C}[1]{>{\centering\arraybackslash}p{#1}}
\newcolumntype{R}[1]{>{\raggedleft\arraybackslash}p{#1}}
\begin{tabular}{L{4.5cm} R{2.5cm} L{4.5cm} R{2.5cm}}
\midrule
\multicolumn{2}{c}{Feature extraction} & \multicolumn{2}{c}{Weight regression}\\
\cmidrule(lr){1-2}
\cmidrule(lr){3-4}
\multicolumn{1}{c}{Type} & \multicolumn{1}{c}{Output} & \multicolumn{1}{c}{Type} & \multicolumn{1}{c}{Output} \\ 
\cmidrule(lr){1-1}
\cmidrule(lr){2-2}
\cmidrule(lr){3-3}
\cmidrule(lr){4-4}
Input & $ D \times 13 \times 13$ & Conv($1\times1$) & $16k^2 \times 1 \times 1$ \\
\addlinespace[0.5em]
Conv($3\times3$)-BN-ReLU & $32\times 11 \times 11$ & Sigmoid & $16k^2 \times 1 \times 1$\\
\addlinespace[0.5em]
Conv($3\times3$)-ReLU & $ 32 \times 9 \times 9$ & Mean subtraction or & \multirow{2}{*}{$16k^2 \times 1 \times 1$} \\
\addlinespace[0.5em]
Conv($3\times3$)-BN-ReLU & $ 64 \times 7 \times 7$ & L1 norm. (w/o Res.) \\
\cmidrule(lr){3-4} %\addlinespace[-0.4em]
Conv($3\times3$)-ReLU & $ 64 \times 5 \times 5$ &\multicolumn{2}{c}{Offset regression} \\
\cmidrule(lr){3-4} %\addlinespace[-0.4em]
Conv($3\times3$)-BN-ReLU & $ 128 \times 3 \times 3$ & \multicolumn{1}{c}{Type} & \multicolumn{1}{c}{Output} \\
\cmidrule(lr){3-3}
\cmidrule(lr){4-4} %\addlinespace[-0.4em]
Conv($3\times3$)-ReLU & $ 128 \times 1 \times 1$ & Conv($1\times1$) & $32k^2 \times 1 \times 1$ \\
\addlinespace[0.5em]
\midrule
\end{tabular}
\vspace{-0.4cm}
\end{table*}

\noindent
\textbf{Testing.}
Two principles have guided the design of our learning architecture:
(1) Points from a large receptive field in the original guidance and
target images should be used to compute the weighted averages
associated with the value of the upsampled depth map at each one of
its pixels; and (2) inference should be fast. The second principle is
rather self-evident. We believe that the first one is also rather
intuitive, and it is justified empirically by the ablation study
presented later. In fine, it is also the basis for our approach, since
our network learns where, and how to sample a small number of points
in a large receptive field.

A reasonable compromise between receptive field size and speed is to
use one or several convolutional layers with a multi-pixel stride,
which enlarges the image area pixels are drawn from without increasing
the number of weights in the network. This is the approach we have
followed in our base architecture, DKN, with two stride-2 ``DownConv"
layers. The price to pay is a loss in spatial resolution for the final
feature map, with only $1/16$ of the total number $N$ of pixels in the
input images. One could of course give as input to our network the
receptive fields associated with all $N$ of the original guidance and
target image pixels, at the cost of $N$ forward passes during
inference. DKN implements a much more efficient method where $16$
shifted copies of the two images are used in turn as input to the
network, and the corresponding network outputs are then stitched
together in a single image, at the cost of only $16$ forward
passes. The details of this {\em shift-and-stitch} approach~\citep{long2015fully,niklaus2017video} can be found in Appendix~\ref{app:implementation}.

\vspace{-0.2cm}
\begin{figure}
\centering
\includegraphics[width=0.9\linewidth]{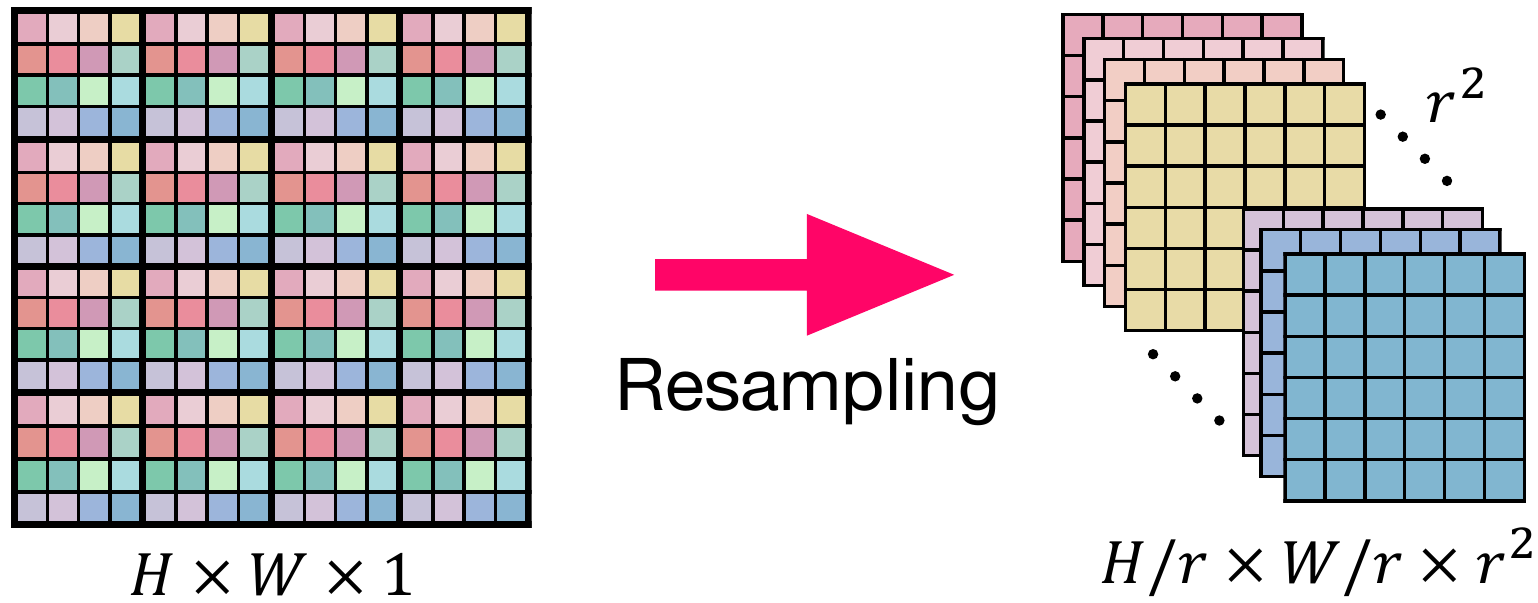}
\caption{Illustration of resampling. An image of size~$H \times W \times 1$ is reshaped with stride~$r$ in each dimension, resulting a resampled one of size~$H/r \times W/r \times r^2$.}
\vspace{-0.4cm}
\label{fig:resampling}
\end{figure}

\begin{table*}[t]
\centering
\caption{Quantitative comparison with the state of the art on depth map upsampling in terms of average RMSE. Numbers in bold indicate the best performance and underscored ones are the second best. Following~\citep{li2016deep,li2017joint}, the average RMSE are measured in centimeter for the NYU v2 dataset~\citep{silberman2012indoor}. For other datasets, we compute RMSE with upsampled depth maps scaled to the range $[0, 255]$. $\dagger$: Our models trained with the depth map only without any guidance.}
\label{table:depth-upsampling}
\addtolength{\tabcolsep}{-2.0pt}
\newcolumntype{L}[1]{>{\raggedright\arraybackslash}p{#1}}
\newcolumntype{C}[1]{>{\centering\arraybackslash}p{#1}}
\newcolumntype{R}[1]{>{\raggedleft\arraybackslash}p{#1}}
\begin{tabular}[c]{L{4cm} R{0.75cm} R{0.75cm} R{0.75cm} R{0.75cm} R{0.75cm} R{0.75cm} R{0.75cm} R{0.75cm} R{0.75cm} R{0.75cm} R{0.75cm} R{0.75cm}} 
\midrule
\multicolumn{1}{c}{Datasets} & \multicolumn{3}{c}{Middlebury} & \multicolumn{3}{c}{Lu} & \multicolumn{3}{c}{NYU v2} & \multicolumn{3}{c}{Sintel} \\
\cmidrule(lr){1-13}
\multicolumn{1}{c}{Methods} &  \multicolumn{1}{c}{$4\times$} &  \multicolumn{1}{c}{$8\times$} &  \multicolumn{1}{c}{$16\times$} &  
\multicolumn{1}{c}{$4\times$} &  \multicolumn{1}{c}{$8\times$} &  \multicolumn{1}{c}{$16\times$} & 
\multicolumn{1}{c}{$4\times$} &  \multicolumn{1}{c}{$8\times$} &  \multicolumn{1}{c}{$16\times$} & 
\multicolumn{1}{c}{$4\times$} &  \multicolumn{1}{c}{$8\times$} &  \multicolumn{1}{c}{$16\times$} \\

\cmidrule{1-13}

Bicubic Int. & 4.44 & 7.58 & 11.87 & 5.07 & 9.22 & 14.27 & 8.16 & 14.22 & 22.32 & 6.54 & 8.80 & 12.17 \\
MRF~\citep{diebel2006application} & 4.26 & 7.43 & 11.80 & 4.90 & 9.03 & 14.19 & 7.84 & 13.98 & 22.20 & 8.81 & 11.77 & 15.75 \\
GF~\citep{he2013guided} & 4.01 & 7.22 & 11.70 & 4.87 & 8.85 & 14.09 & 7.32 & 13.62 & 22.03 & 6.10 & 8.22 & 11.22 \\

JBU~\citep{kopf2007joint} & 2.44 & 3.81 & 6.13 & 2.99 & 5.06 & 7.51 & 4.07 & 8.29 & 13.35  & 5.88 & 7.63 & 10.97 \\

TGV~\citep{ferstl2013} & 3.39 & 5.41 & 12.03 & 4.48 & 7.58 & 17.46 & 6.98 & 11.23 & 28.13 & 32.01 & 36.78 & 43.89 \\

Park~\citep{park2011} & 2.82 & 4.08 & 7.26 & 4.09 & 6.19 & 10.14 & 5.21 & 9.56 & 18.10 & 9.28 & 12.22 & 16.51 \\

SDF~\citep{ham2018robust} & 3.14 & 5.03 & 8.83 & 4.65 & 7.53 & 11.52 & 5.27 & 12.31 & 19.24 & 6.52 & 7.98 & 11.36 \\

FBS~\citep{barron2016fast} & 2.58 & 4.19 & 7.30 & 3.03 & 5.77 & 8.48 & 4.29 & 8.94 & 14.59 & 11.96 & 12.29 & 13.08 \\
\cmidrule{1-13}
\multicolumn{13}{c}{Results with downsampling based on bicubic interpolation}\\
\cmidrule{1-13}
DMSG~\citep{hui2016depth} & 1.88 & 3.45 & 6.28 & 2.30 & 4.17 & 7.22 & 3.02 & 5.38 & 9.17 & 5.32 & 7.24 & 10.11 \\
{DJF}~\citep{li2016deep} & {1.68} & {3.24} & {5.62} & {1.65} & {3.96} & {6.75} & {2.80} & {5.33} & {9.46} & {4.58} & {6.57} & {9.38} \\
{DG}~\citep{Gu2017learning} & {1.97} & {4.16} & {5.27} & {2.06} & {4.19} & {6.90} & {3.68} & {5.78} & {10.08} & {4.11} & {6.61} & {9.24} \\
{DJFR}~\citep{li2017joint} & {1.32} & {3.19} & {5.57} & {1.15} & {3.57} & {6.77} & {2.38} & {4.94} & {9.18} & {3.97} & {6.32} & {9.19} \\

PAC~\citep{su2019pixel}  & 1.32 & 2.62  & 4.58  & 1.20  & 2.33 & 5.19 & 1.89 &  3.33 & 6.78 & 3.38  & 4.89 & 7.65 \\
\cmidrule{1-13}

FDKN$^\dagger$ & \textbf{1.07} & 2.23 & 5.09 & \underline{0.85} & \underline{1.90} & 5.33 & 2.05 & 4.10 & 8.10 & \underline{3.31} & 5.08 & 8.51\\

DKN$^\dagger$ & 1.12 & \underline{2.13} & 5.00 & 0.90 & \textbf{1.83} & \textbf{4.99} & 2.11 & 4.00 & 8.24 & 3.40 & 4.90 & 8.18 \\
\cmidrule{1-13}
FDKN w/o Res. & 1.12 & 2.23 & 4.52 & \underline{0.85} & 2.19 & 5.15 & 1.88 & 3.67 & 7.13 & 3.38 & 5.02 & 7.74 \\

DKN w/o Res. & 1.26 & 2.16 & \underline{4.32} & 0.99 & 2.21 & 5.12 & \underline{1.66} & \underline{3.36} & \underline{6.78} & 3.36 & \underline{4.82} & \textbf{7.48} \\

FDKN & \underline{1.08} & 2.17 & 4.50 & \textbf{0.82} & 2.10 & \underline{5.05} & 1.86 & 3.58 & 6.96 & 3.36 & 4.96 & 7.74 \\

DKN& 1.23 & \textbf{2.12} & \textbf{4.24} & 0.96 & 2.16 & 5.11 & \textbf{1.62} & \textbf{3.26} & \textbf{6.51} & \textbf{3.30} & \textbf{4.77} & \underline{7.59} \\

\cmidrule{1-13}
\multicolumn{13}{c}{Results with downsampling based on nearest-neighbor interpolation}\\
\cmidrule{1-13}

DMSG~\citep{hui2016depth} &  2.11 & 3.74  & 6.03  & 2.48  & 4.74 & 7.51 & 3.37 & 6.20 & 10.05 & 5.49 & 7.48 & 10.52\\
DJF~\citep{li2016deep} & 2.14  & 3.77  & 6.12  & 2.54  & 4.71 & 7.66 & 3.54 & 6.20 & 10.21 & 5.51 & 7.52 & 10.63 \\
DJFR~\citep{li2017joint} & 1.98  & 3.61  & 6.07  &  2.22 & 4.54 & 7.48 & 3.38 & 5.86 & 10.11 & 5.50 & 7.43 & 10.48\\

PAC~\citep{su2019pixel} & \textbf{1.91} & 3.20 & 5.60 & 2.48 & 4.37 & 6.60 &  2.82 &  5.01 & 8.64 & 5.41 & 6.98 &  9.64 \\

\cmidrule{1-13}
FDKN$^\dagger$ &  2.89 & 3.92  & 6.75  & 2.85  & 4.64 & 7.62 &  3.64 & 5.43 & 8.96 & 5.74 & 7.31 & 10.31\\
DKN$^\dagger$ & 2.94  & 4.14  & 8.12  & 2.88  & 5.13 & 8.24 & 3.67 & 6.68 & 12.15 & 5.78 & 7.59 & 10.95 \\
\midrule
FDKN w/o Res. &  2.11 & 3.63  & 6.29  & 2.50  & 4.52 & 7.37 & 2.65 & 5.02 & 8.69  & 5.48 & 7.27 & 10.22\\
DKN w/o Res. &  1.95 &  \underline{3.18} & \underline{5.50}  & \underline{2.37} & \underline{4.17} & \underline{6.37} & \underline{2.48} & \underline{4.78} & \underline{8.52} & \textbf{5.29} & \underline{6.95} & \textbf{9.53} \\
FDKN &  2.21  &  3.64 & 6.15  &  2.64 & 4.55 & 7.20 &  2.62 &  4.99 & 8.67 & 5.33 & 7.25 & 9.86\\
DKN & \underline{1.93} &  \textbf{3.17} & \textbf{5.49}  & \textbf{2.35} & \textbf{4.16} & \textbf{6.33} & \textbf{2.46} & \textbf{4.76} & \textbf{8.50} & \textbf{5.29} & \textbf{6.92} & \underline{9.56}  \\
\midrule
\end{tabular}
\vspace{-.4cm}
\end{table*}

\vspace{-.5cm}
\subsection{FDKN architecture}\label{subsec:fdkn}
\vspace{-.3cm}
We now present a fast version of DKN~(FDKN) that achieves a $17 \times$ speed-up over the DKN for an image of size~$640 \times 480$ while retaining the same level of quality~(Fig.~\ref{fig:fast-architecture}). The basic idea is to remove DownConv layers while maintaining the same receptive field size as the DKN, making it possible to obtain the filtering output in a single forward pass. Simply removing these layers leads to an increase in the total number of convolutions~(see Section~\ref{sec:discussion}). A more efficient alternative to DKN is to split the input images into the same 16 subsampled and shifted parts as before, but this time {\em stack} them into new target and guidance images~(Fig.~\ref{fig:resampling}), with $16$ channels for the former, and $16C$ channels for the latter, e.g.,~$C=3$ when the RGB image is used. The effective receptive field for FDKN is comparable to that of DKN\footnote{We could also use dilated convolutions~\citep{yu2016multi} that support large receptive fields without loss of resolution, but empirically find that runtime gain is marginal~(see Section~\ref{sec:discussion}).}, but FDKN involves much fewer parameters because of the reduced input image resolution and the shared weights across channels. The individual channels are then recomposed into the final upsampled image~\citep{shi2016real}, at the cost of only one forward pass. Specifically, we use a series of 6 convolutional layers of size~$3 \times 3$ for feature extraction. For weight and offset regression, we apply a $1\times 1$ convolution on top of the feature extraction layers similar to DKN, but using more network parameters. For example, FDKN and DKN compute feature maps of size $16k^2 \times 1 \times 1$ and $k^2 \times 1 \times 1$, respectively, for weight regression, from each feature of size $128 \times 1 \times 1$. This allows FDKN to estimate kernel weights and offsets for all pixels simultaneously. In practice, FDKN gives a 17 times speed-up over DKN. Because it involves fewer parameters~($0.6$M vs. $1.1$M for DKN), one might expect somewhat degraded results. Our experiments demonstrate that FDKN remains in the ballpark of that of DKN, still significantly better than competing approaches, and in one case even overperforming DKN. We show in Table~\ref{table:fastarchitecture} the detailed description of the network structure. 

\vspace{-.55cm}
\section{Experiments}\label{sec:exp}
\vspace{-.2cm}
In this section we present a detailed analysis and evaluation of our approach. We apply our models to the tasks of joint image upsampling~(Section~\ref{exp:joint}), cross-modality image restoration and texture removal~(Section~\ref{exp:other}) and semantic segmentation~(Section~\ref{exp:ss}), and compare them to the state of the art in each case. The inputs of our network are 3-channel guidance and 1-channel target images. In case of a 1-channel guidance image~(e.g., RGB/NIR image restoration), we create a 3-channel image by duplicating the single channel three times. For multi-channel target images~(e.g., in texture removal), we apply our model separately in each channel and combine the outputs. The results for comparisons have been obtained from the source code provided by the authors.

\begin{figure*}[t]
\vspace{-0.2cm}
  \centering
  \footnotesize	
\begin{flushright}
	\includegraphics[width=0.5\linewidth]{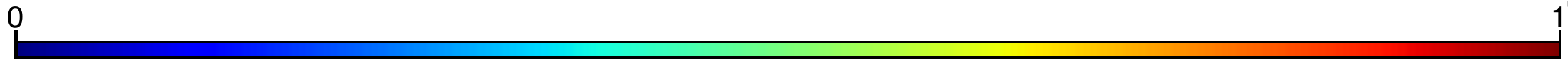}\vspace{-.5cm}
	\end{flushright}
  \subfloat[{RGB image.}]{
    \begin{minipage}[b]{0.141\linewidth} 
      \includegraphics[width=\linewidth, height=1.891cm]{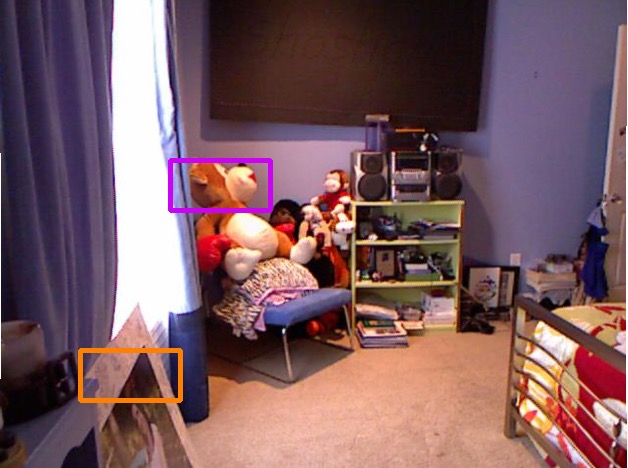}\vspace{0.05cm} 
      \includegraphics[width=\linewidth, height=1.891cm]{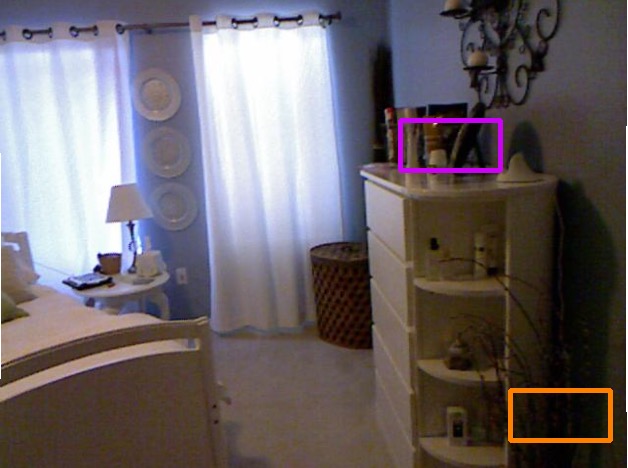}\vspace{0.05cm}
      \includegraphics[width=\linewidth, height=1.891cm]{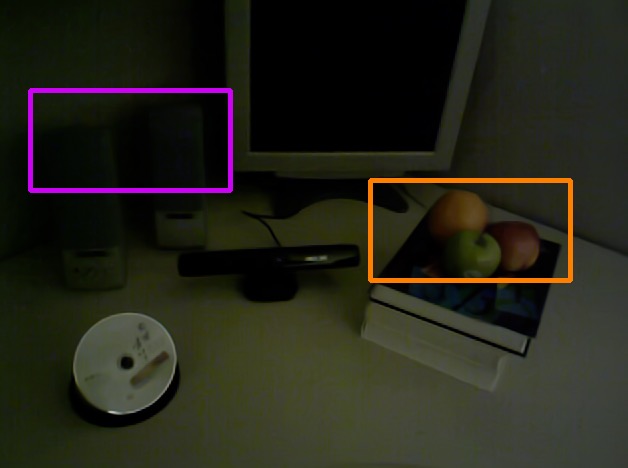}\vspace{0.05cm}
      \includegraphics[width=\linewidth, height=1.891cm]{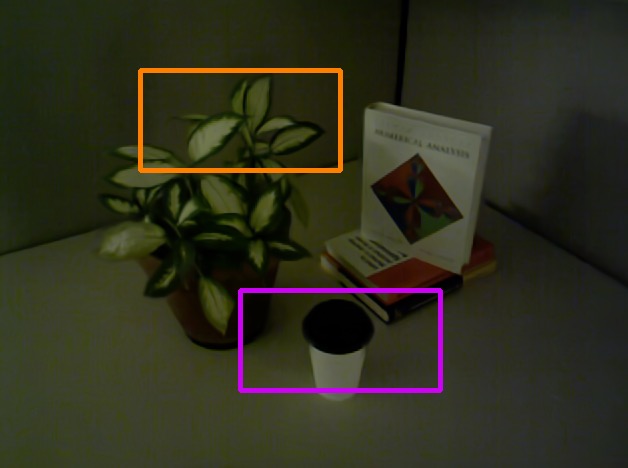}\vspace{0.05cm}
      \includegraphics[width=\linewidth, height=1.891cm]{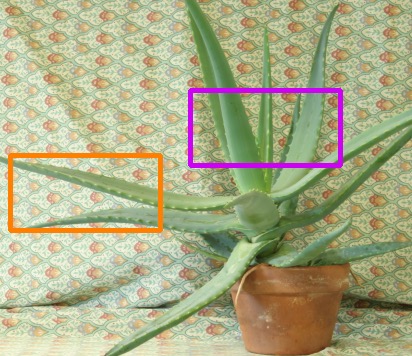}\vspace{0.05cm}
      \includegraphics[width=\linewidth, height=1.891cm]{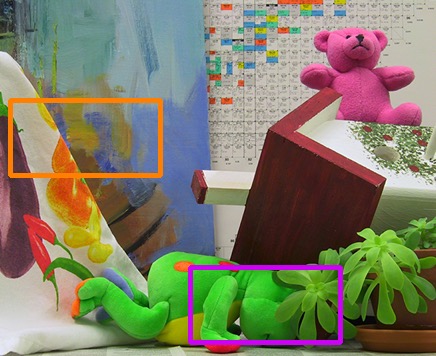}\vspace{0.05cm}
      \includegraphics[width=\linewidth, height=1.891cm]{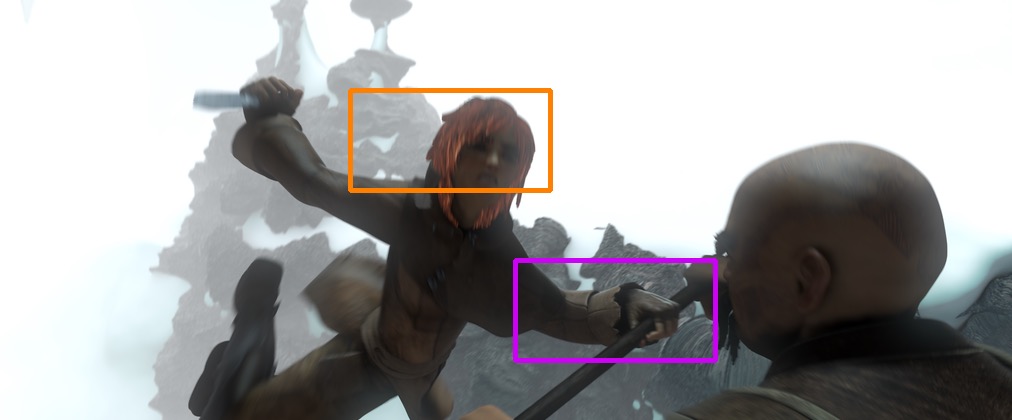}\vspace{0.05cm}
      \includegraphics[width=\linewidth, height=1.891cm]{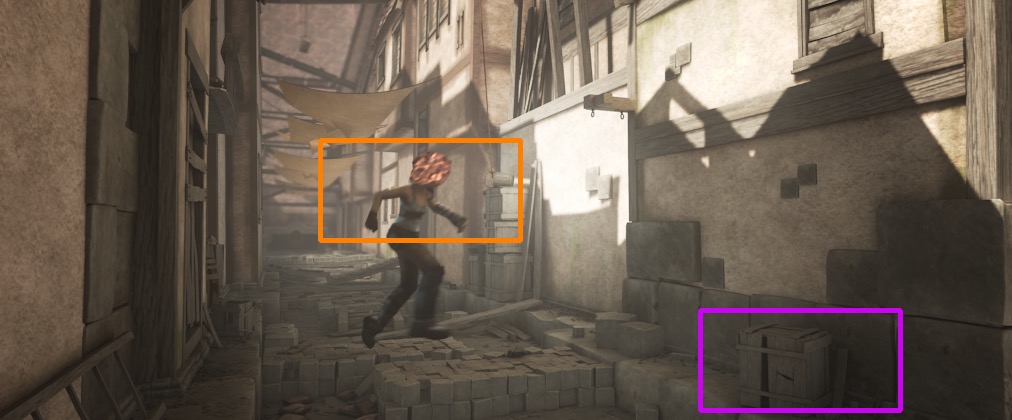}
    \end{minipage}
  }\hspace{-0.22cm}
  \subfloat[{GF}.]{
    \begin{minipage}[b]{0.106\linewidth} 
      \includegraphics[width=\linewidth]{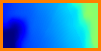}
      \includegraphics[width=\linewidth]{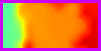}\vspace{0.05cm}    
      \includegraphics[width=\linewidth]{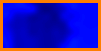}
      \includegraphics[width=\linewidth]{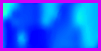}\vspace{0.05cm}
      \includegraphics[width=\linewidth]{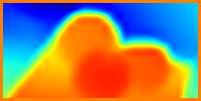}
      \includegraphics[width=\linewidth]{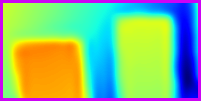}\vspace{0.05cm}
      \includegraphics[width=\linewidth]{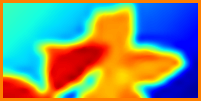}
      \includegraphics[width=\linewidth]{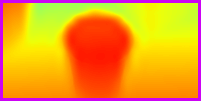}\vspace{0.05cm}
      \includegraphics[width=\linewidth]{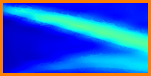}
      \includegraphics[width=\linewidth]{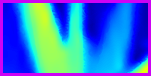}\vspace{0.05cm}
      \includegraphics[width=\linewidth]{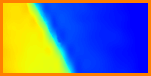}
      \includegraphics[width=\linewidth]{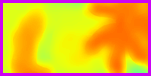}\vspace{0.05cm}
      \includegraphics[width=\linewidth]{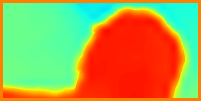}
      \includegraphics[width=\linewidth]{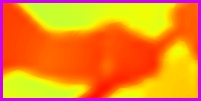}\vspace{0.05cm}
      \includegraphics[width=\linewidth]{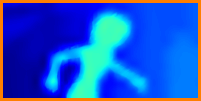}
      \includegraphics[width=\linewidth]{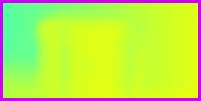}
    \end{minipage}
  }\hspace{-0.22cm}
  \subfloat[{TGV}.]{
    \begin{minipage}[b]{0.106\linewidth} 
      \includegraphics[width=\linewidth]{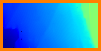}
      \includegraphics[width=\linewidth]{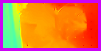}\vspace{0.05cm}
      \includegraphics[width=\linewidth]{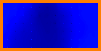}
      \includegraphics[width=\linewidth]{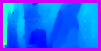}\vspace{0.05cm} 
      \includegraphics[width=\linewidth]{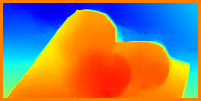}
      \includegraphics[width=\linewidth]{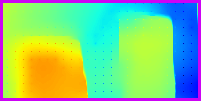}\vspace{0.05cm} 
      \includegraphics[width=\linewidth]{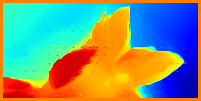}
      \includegraphics[width=\linewidth]{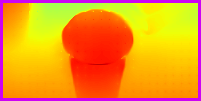}\vspace{0.05cm} 
      \includegraphics[width=\linewidth]{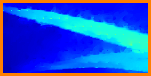}
      \includegraphics[width=\linewidth]{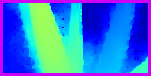}\vspace{0.05cm} 
      \includegraphics[width=\linewidth]{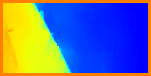}
      \includegraphics[width=\linewidth]{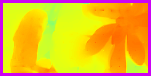}\vspace{0.05cm} 
      \includegraphics[width=\linewidth]{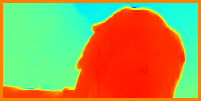}
      \includegraphics[width=\linewidth]{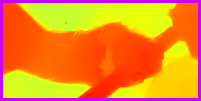}\vspace{0.05cm} 
      \includegraphics[width=\linewidth]{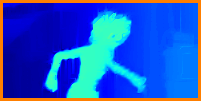}
      \includegraphics[width=\linewidth]{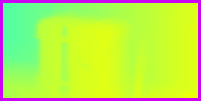}
    \end{minipage}
  }\hspace{-0.22cm}
  \subfloat[{SDF}.]{
    \begin{minipage}[b]{0.106\linewidth} 
      \includegraphics[width=\linewidth]{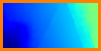}
      \includegraphics[width=\linewidth]{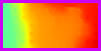}\vspace{0.05cm}
      \includegraphics[width=\linewidth]{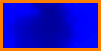}
      \includegraphics[width=\linewidth]{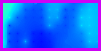}\vspace{0.05cm}
      \includegraphics[width=\linewidth]{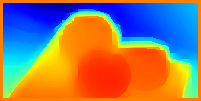}
      \includegraphics[width=\linewidth]{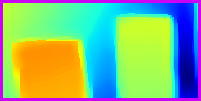}\vspace{0.05cm}
      \includegraphics[width=\linewidth]{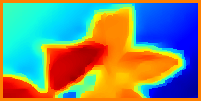}
      \includegraphics[width=\linewidth]{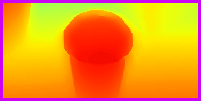}\vspace{0.05cm}
      \includegraphics[width=\linewidth]{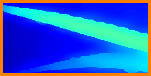}
      \includegraphics[width=\linewidth]{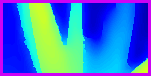}\vspace{0.05cm}
      \includegraphics[width=\linewidth]{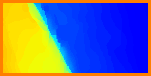}
      \includegraphics[width=\linewidth]{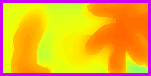}\vspace{0.05cm}
      \includegraphics[width=\linewidth]{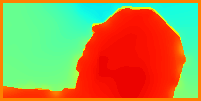}
      \includegraphics[width=\linewidth]{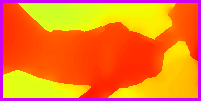}\vspace{0.05cm}
      \includegraphics[width=\linewidth]{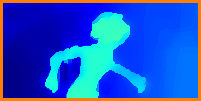}
      \includegraphics[width=\linewidth]{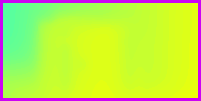}
    \end{minipage}
  }\hspace{-0.22cm}
  \subfloat[{DJFR}.]{
    \begin{minipage}[b]{0.106\linewidth} 
      \includegraphics[width=\linewidth]{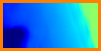}
      \includegraphics[width=\linewidth]{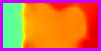}\vspace{0.05cm}
      \includegraphics[width=\linewidth]{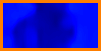}
      \includegraphics[width=\linewidth]{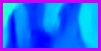}\vspace{0.05cm}  
      \includegraphics[width=\linewidth]{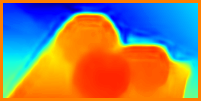}
      \includegraphics[width=\linewidth]{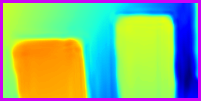}\vspace{0.05cm} 
      \includegraphics[width=\linewidth]{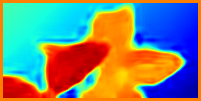}
      \includegraphics[width=\linewidth]{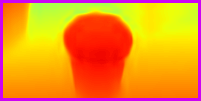}\vspace{0.05cm} 
      \includegraphics[width=\linewidth]{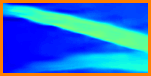}
      \includegraphics[width=\linewidth]{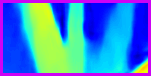}\vspace{0.05cm} 
      \includegraphics[width=\linewidth]{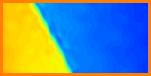}
      \includegraphics[width=\linewidth]{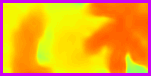}\vspace{0.05cm} 
      \includegraphics[width=\linewidth]{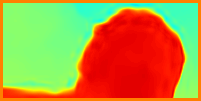}
      \includegraphics[width=\linewidth]{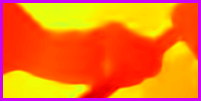}\vspace{0.05cm} 
      \includegraphics[width=\linewidth]{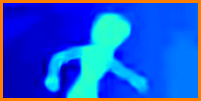}
      \includegraphics[width=\linewidth]{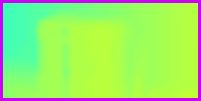}
    \end{minipage}
  }\hspace{-0.22cm}
  \subfloat[PAC.]{
    \begin{minipage}[b]{0.106\linewidth} 
      \includegraphics[width=\linewidth]{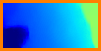}
      \includegraphics[width=\linewidth]{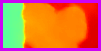}\vspace{0.05cm}    
      \includegraphics[width=\linewidth]{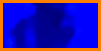}
      \includegraphics[width=\linewidth]{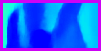}\vspace{0.05cm}
      \includegraphics[width=\linewidth]{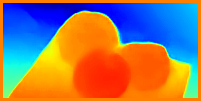}
      \includegraphics[width=\linewidth]{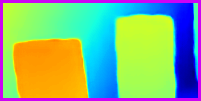}\vspace{0.05cm} 
      \includegraphics[width=\linewidth]{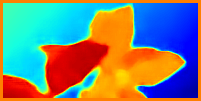}
      \includegraphics[width=\linewidth]{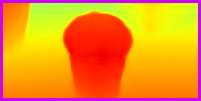}\vspace{0.05cm} 
      \includegraphics[width=\linewidth]{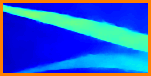}
      \includegraphics[width=\linewidth]{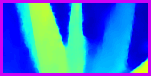}\vspace{0.05cm} 
      \includegraphics[width=\linewidth]{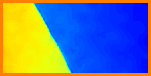}
      \includegraphics[width=\linewidth]{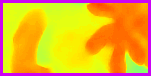}\vspace{0.05cm} 
      \includegraphics[width=\linewidth]{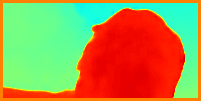}
      \includegraphics[width=\linewidth]{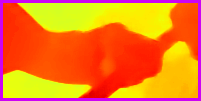}\vspace{0.05cm} 
      \includegraphics[width=\linewidth]{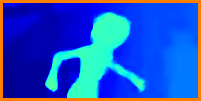}
      \includegraphics[width=\linewidth]{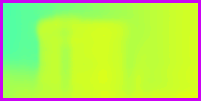}
    \end{minipage}
  }\hspace{-0.22cm}
  \subfloat[{DKN}.]{
    \begin{minipage}[b]{0.106\linewidth} 
      \includegraphics[width=\linewidth]{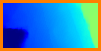}
      \includegraphics[width=\linewidth]{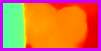}\vspace{0.05cm}
      \includegraphics[width=\linewidth]{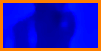}
      \includegraphics[width=\linewidth]{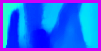}\vspace{0.05cm}
      \includegraphics[width=\linewidth]{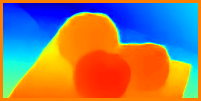}
      \includegraphics[width=\linewidth]{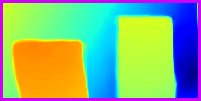}\vspace{0.05cm} 
      \includegraphics[width=\linewidth]{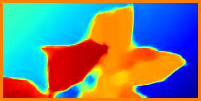}
      \includegraphics[width=\linewidth]{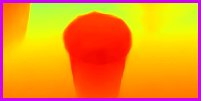}\vspace{0.05cm} 
      \includegraphics[width=\linewidth]{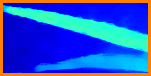}
      \includegraphics[width=\linewidth]{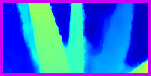}\vspace{0.05cm} 
      \includegraphics[width=\linewidth]{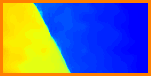}
      \includegraphics[width=\linewidth]{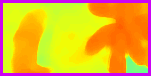}\vspace{0.05cm} 
      \includegraphics[width=\linewidth]{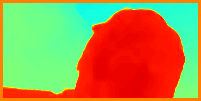}
      \includegraphics[width=\linewidth]{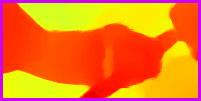}\vspace{0.05cm} 
      \includegraphics[width=\linewidth]{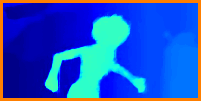}
      \includegraphics[width=\linewidth]{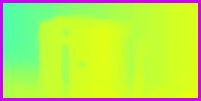}
    \end{minipage}
  }\hspace{-0.22cm}
  \subfloat[{FDKN}.]{
    \begin{minipage}[b]{0.106\linewidth} 
      \includegraphics[width=\linewidth]{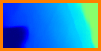}
      \includegraphics[width=\linewidth]{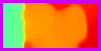}\vspace{0.05cm}
      \includegraphics[width=\linewidth]{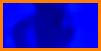}
      \includegraphics[width=\linewidth]{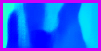}\vspace{0.05cm}
      \includegraphics[width=\linewidth]{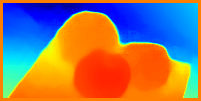}
      \includegraphics[width=\linewidth]{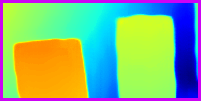}\vspace{0.05cm} 
      \includegraphics[width=\linewidth]{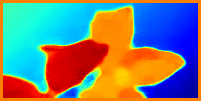}
      \includegraphics[width=\linewidth]{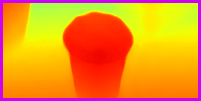}\vspace{0.05cm} 
      \includegraphics[width=\linewidth]{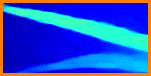}
      \includegraphics[width=\linewidth]{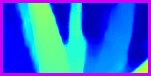}\vspace{0.05cm} 
      \includegraphics[width=\linewidth]{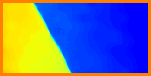}
      \includegraphics[width=\linewidth]{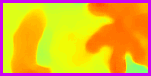}\vspace{0.05cm} 
      \includegraphics[width=\linewidth]{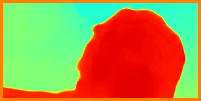}
      \includegraphics[width=\linewidth]{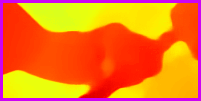}\vspace{0.05cm} 
      \includegraphics[width=\linewidth]{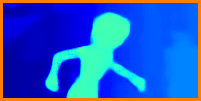}
      \includegraphics[width=\linewidth]{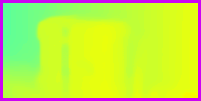}
    \end{minipage}
  }\hspace{-0.22cm}
  \subfloat[{Ground truth}.]{
    \begin{minipage}[b]{0.106\linewidth} 
      \includegraphics[width=\linewidth]{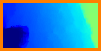}
      \includegraphics[width=\linewidth]{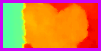}\vspace{0.05cm}
      \includegraphics[width=\linewidth]{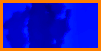}
      \includegraphics[width=\linewidth]{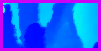}\vspace{0.05cm}
      \includegraphics[width=\linewidth]{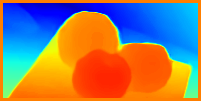}
      \includegraphics[width=\linewidth]{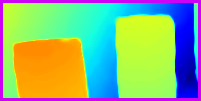}\vspace{0.05cm} 
      \includegraphics[width=\linewidth]{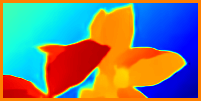}
      \includegraphics[width=\linewidth]{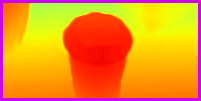}\vspace{0.05cm} 
      \includegraphics[width=\linewidth]{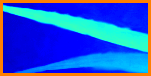}
      \includegraphics[width=\linewidth]{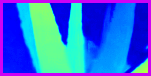}\vspace{0.05cm} 
      \includegraphics[width=\linewidth]{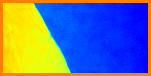}
      \includegraphics[width=\linewidth]{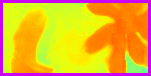}\vspace{0.05cm} 
      \includegraphics[width=\linewidth]{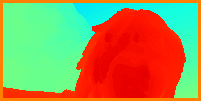}
      \includegraphics[width=\linewidth]{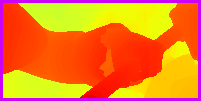}\vspace{0.05cm} 
      \includegraphics[width=\linewidth]{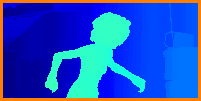}
      \includegraphics[width=\linewidth]{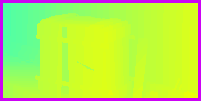}
    \end{minipage}
  }
  \caption{Visual comparison of upsampled depth images~($8 \times$): (a)~an RGB image, (b)~GF~\citep{he2013guided}, (c)~TGV~\citep{ferstl2013}, (d)~SDF~\citep{ham2018robust}, (e)~DJFR~\citep{li2017joint}, (f)~PAC~\citep{su2019pixel}, (g)~DKN, (h)~FDKN, and (i)~ground truth. Top to bottom: Each two rows show upsampled images on the NYU v2~\citep{silberman2012indoor}, Lu~\citep{lu2014depth}, Middlebury~\citep{hirschmuller2007evaluation} and Sintel~\citep{butler2012sintel} datasets, respectively. Note that we train our models with the NYU v2 dataset, and do not fine-tune them to other datasets.}
  \label{fig:depth_upsampling}
  \vspace{-0.3cm}
\end{figure*}

%We obtain filtering results of DKN by stitching coarse outputs, which is far more efficient than a brute-force implementation. Filtering results for FDKN are obtained in a single pass. 
\vspace{-0.6cm}
\subsection{Joint image upsampling experiments}\label{exp:joint}
\vspace{-0.3cm}
Following the experimental protocol of~\citep{li2016deep,li2017joint}, we train our models on the task of depth map upsampling using a large number of RGB/D image pairs, and test them on the tasks of (noisy) depth map upsampling, and saliency image upsampling that require selectively transferring the structural details from the guidance image to the target one. 

To our knowledge, there is no standard protocol for simulating low-resolution depth images. For example, DJF~\citep{li2016deep}, DJFR~\citep{li2017joint}, and PAC~\citep{su2019pixel} use nearest-neighbor downsampling, e.g.,~selecting a single pixel from a $8 \times 8$ window for a scale factor of $\times 8$. In particular, PAC keeps the center pixel, while DJF and DJFR select the right-bottom one. DMSG~\citep{hui2016depth} and DG~\citep{Gu2017learning} exploit a bicubic downsampling technique. For fair comparison, we report the results obtained by all methods, DJF, DJFR, PAC, DMSG\footnote{It uses the Middlebury and Sintel datasets for training the network. For fair comparison of DMSG with other CNN-based methods including ours, we retrain the DMSG model using the same image pairs from the NYU v2 dataset as in~\citep{li2016deep} }, DG, using bicubic downsampling, unless otherwise specified. We also perform the same comparison using nearest-neighbor downsampling as in DJF~\citep{li2016deep} and DJFR~\citep{li2017joint}~(i.e.,~selecting the right-bottom pixel for each sampling grid) for all methods, except for DG, as it does not provide a source code for training. 

\vspace{-0.2cm}
\subsubsection{Experimental details}\label{exp:detail}
\vspace{-0.3cm}

\begin{figure}
\vspace{-.3cm}
\centering
  \footnotesize		
  \subfloat[RGB image.]{
    \begin{minipage}[b]{0.4\linewidth} 
      \includegraphics[width=\linewidth]{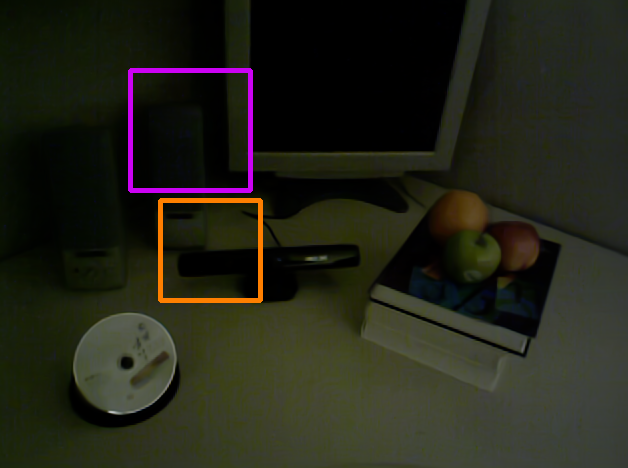}\\
      \includegraphics[width=0.494\linewidth]{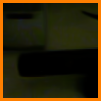}\hspace{-0.04cm}
      \includegraphics[width=0.494\linewidth]{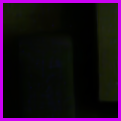}
    \end{minipage}
  }\hspace{-0.2cm}
  \subfloat[GT.]{
    \begin{minipage}[b]{0.4\linewidth} 
      \includegraphics[width=\linewidth]{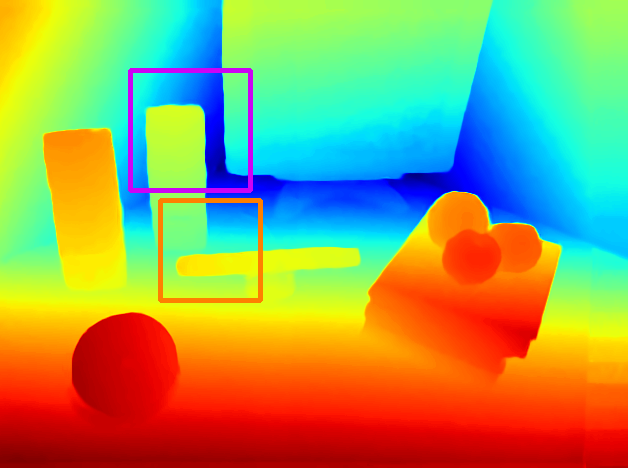}\\
      \includegraphics[width=0.494\linewidth]{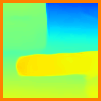}\hspace{-0.04cm}
      \includegraphics[width=0.494\linewidth]{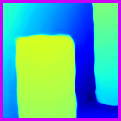}
    \end{minipage}
  }\hspace{-0.15cm}
  \begin{minipage}[b]{0.1\linewidth}
%  \vspace*{1cm}
  \includegraphics[width=0.64\linewidth]{images/vert_colorbar}
  \end{minipage}
  \hspace{-0.2cm} \\ \vspace{-0.3cm}
  \subfloat[DKN.]{
    \begin{minipage}[b]{0.4\linewidth} 
      \includegraphics[width=\linewidth]{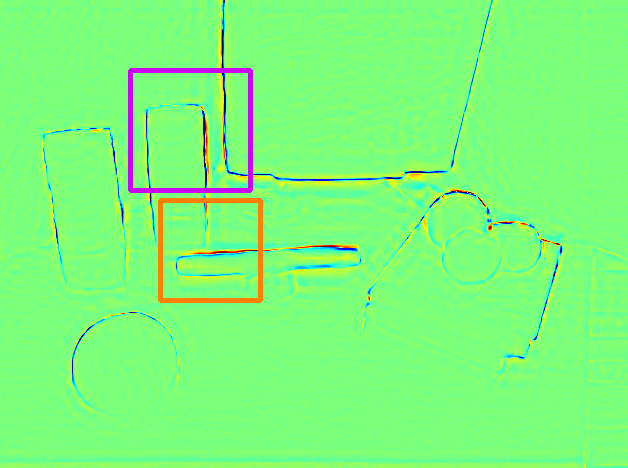}\\
       \includegraphics[width=0.494\linewidth]{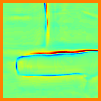}\hspace{-0.04cm}
      \includegraphics[width=0.494\linewidth]{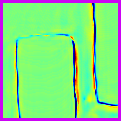}
    \end{minipage}
  }\hspace{-0.2cm}
  \subfloat[DKN$^\dagger$.]{
    \begin{minipage}[b]{0.4\linewidth} 
      \includegraphics[width=\linewidth]{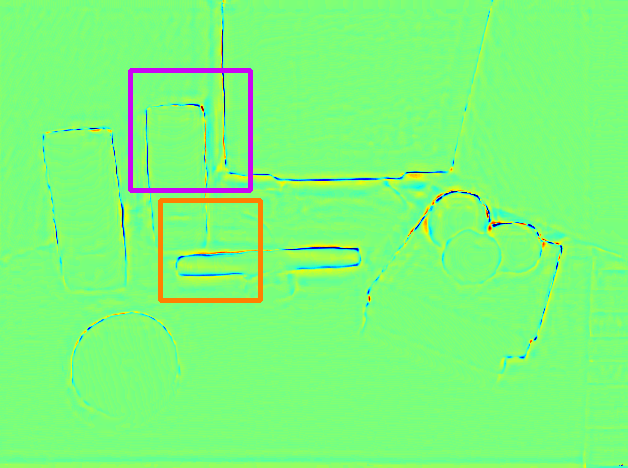}
             \includegraphics[width=0.494\linewidth]{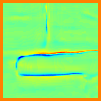}\hspace{-0.04cm}
      \includegraphics[width=0.494\linewidth]{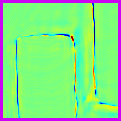}
    \end{minipage}
  }\hspace{-0.15cm}
  \begin{minipage}[b]{0.1\linewidth}
%  \vspace*{1cm}
  \includegraphics[width=0.96\linewidth]{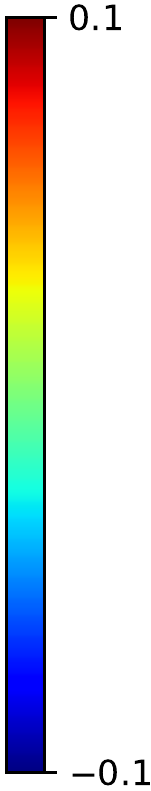}
  \end{minipage}
\caption{An example from the Lu dataset~\citep{lu2014depth} where the RGB guidance image does not help. From left to right: (a) an RGB image, (b) ground-truth depth, and results from (c) DKN and (d) DKN$^\dagger$, rendered as difference maps between upsampling results and ground truth. In this example, the poor contrast of color edges in dark regions hampers the ability of the RGB guidance to assist upsampling, and DKN$^\dagger$ performs better than DKN.}
\vspace{-.4cm}
\label{fig:limitation}
\end{figure}

We sample 1,000 RGB/D image pairs of size $640 \times 480$ from the NYU v2 dataset~\citep{silberman2012indoor}. We use the same image pairs as in~\citep{li2016deep,li2017joint} to train the networks. We train different models to upsample depth map with a batch size of 1 for 40k iterations, giving roughly 20 epochs over the training data. We use the Adam optimizer~\citep{kingma2015adam} with $\beta_1 = 0.9$ and $\beta_2 = 0.999$. As learning rate we use $0.001$ and divide it by 5 every 10k iterations. Data augmentation and regularization techniques such as weight decay and dropout~\citep{krizhevsky2012imagenet} are not used, since 1,000 RGB/D image pairs from the NYU dataset have proven to be sufficient to train our models. All networks are trained end-to-end using \texttt{PyTorch}~\citep{paszke2017automatic}.

%\vspace{-1.1cm}

\vspace{-0.4cm}
\subsubsection{Results}
\vspace{-0.3cm}
\noindent{\textbf{Depth map upsampling.}} We test our models on depth map upsampling with the following four benchmark datasets. These datasets feature aligned color and depth images. The inputs to our models are a high-resolution color image and a low-resolution depth image upsampled using bicubic or nearest-neighbor interpolation.

\vspace{-0.3cm}
\begin{itemize}[leftmargin=*]
	\item[$\bullet$] Middlebury dataset~\citep{hirschmuller2007evaluation,scharstein2007learning}: We use the 30 RGB/D image pairs from the 2001-2006 datasets provided by Lu~\citep{lu2014depth}.
	\item[$\bullet$] Lu dataset~\citep{lu2014depth}: This provides 6 RGB/D image pairs acquired by the ASUS Xtion Pro camera. %~\citep{asus2018xtion}. 
	\item[$\bullet$] NYU v2 dataset~\citep{silberman2012indoor}: It consists of 1,449 RGB/D image pairs captured with the Microsoft Kinect~\citep{zhang2012kinect} using structured light, similar to the Middlebury dataset. We exclude the 1,000 pairs used for training, and use the rest (449 pairs) for evaluation.
	\item[$\bullet$] Sintel dataset~\citep{butler2012sintel}: This dataset provides 1,064 RGB/D image pairs created from an animated 3D movie. It contains realistic scenes including fog and motion blur. We use 864 pairs from a final-pass dataset for testing\footnote{We discard the 200 pairs that provide RGB images only.}.
\end{itemize}
%\vspace{-.4cm}

\begin{table*}[t]
\centering
\caption{Quantitative comparison with the state of the art on noisy depth map upsampling in terms of average RMSE on the noisy Middlebury dataset~\citep{park2011}. $\dagger$: The model trained with a synthetic dataset.}\label{table:noise-middlebury}
\addtolength{\tabcolsep}{-2.0pt}
\newcolumntype{L}[1]{>{\raggedright\arraybackslash}p{#1}}
\newcolumntype{C}[1]{>{\centering\arraybackslash}p{#1}}
\newcolumntype{R}[1]{>{\raggedleft\arraybackslash}p{#1}}
\begin{tabular}[c]{L{3.6cm} R{0.75cm} R{0.75cm} R{0.75cm} R{0.75cm} R{0.75cm} R{0.75cm} R{0.75cm} R{0.75cm} R{0.75cm} } 
\midrule
\multicolumn{1}{c}{} & \multicolumn{3}{c}{Art} & \multicolumn{3}{c}{Books} & \multicolumn{3}{c}{Moebius} \\
\cmidrule(lr){1-10}
\multicolumn{1}{c}{Methods} & \multicolumn{1}{c}{$4\times$} &  \multicolumn{1}{c}{$8\times$} &  \multicolumn{1}{c}{$16\times$} &\multicolumn{1}{c}{$4\times$} &  \multicolumn{1}{c}{$8\times$} &  \multicolumn{1}{c}{$16\times$} & \multicolumn{1}{c}{$4\times$} &  \multicolumn{1}{c}{$8\times$} &  \multicolumn{1}{c}{$16\times$} \\

\cmidrule{1-10}

Bicubic  & 6.07 & 7.27 & 9.59  & 5.15 & 5.45 & 5.97  & 5.51 & 5.68 & 6.11 \\
DMSG~\citep{hui2016depth}  & 6.19 & 7.26 & 9.53 & 5.38 & 5.18 & 5.20 & 5.48 & 5.06 & 5.36 \\
DJFR~\citep{li2017joint} & 4.25 & 6.43 & 9.05  & 2.20 & 3.35 & 4.94 & 2.39 & 3.51 & 4.56 \\ 
PAC~\citep{su2019pixel}  & 5.34 & 7.69 & 10.66 & 2.11 & 3.12 & 4.60 & 2.21 & 3.38 & 4.72 \\
PDN$^\dagger$~\citep{riegler16gdsr} & 3.11 & 4.48 & \underline{7.35} & 1.56 & 2.24 & 3.46 & 1.68 & 2.48 & \underline{3.62} \\
%DG~\citep{Gu2017learning} & \bf{2.96} & 4.41 & \underline{7.06} & 1.64 & 2.35 & 3.50 & 1.74 & 2.57 & 3.79 \\

\cmidrule{1-10}

FDKN w/o Res. & 3.24 & 4.57 & 7.67 & 1.55 & 2.18 & \underline{3.32} & 1.76 & 2.57 & 3.92\\
FDKN & 3.14  & 4.47 & 7.61 & 1.49 & \underline{2.13} & 3.40 & 1.70 & 2.53 & 3.91 \\

DKN w/o Res. & \bf{2.98} & \underline{4.25} & 7.78 & \underline{1.47} & 2.16 & 3.62 & \underline{1.64} & \underline{2.43} & 4.03 \\
DKN  & \underline{3.01} & \bf{4.14} & \bf{7.01} & \bf{1.44} & \bf{2.10} & \bf{3.09} & \bf{1.63} & \bf{2.39} & \bf{3.55} \\ 
\midrule
\end{tabular}
\vspace{-0.3cm}
\end{table*}

We compare our method with the state of the art in Table~\ref{table:depth-upsampling}. It shows the average RMSE between upsampling results and ground truth. From this table, we observe three things:~(1) Our models outperform the state of the art including CNN-based methods~\citep{hui2016depth,li2016deep,li2017joint,su2019pixel} by significant margins in terms of RMSE, even without the residual connection~(DKN w/o Res. and FDKN w/o Res.). For example, DKN decreases the average RMSE by $32\%$~($4 \times$), $34\%$~($8 \times$) and $29\%$~($16 \times$) compared to DJFR. (2)~We can clearly see that our models perform well on both synthetic and real datasets~(e.g., the Sintel and NYU v2 datasets), and generalize well to other images~(e.g., on the Middlebury dataset)~outside the training dataset. Note that we train our models with the NYU v2 dataset, and do not fine-tune them to other datasets. (3)~FDKN retains the superior performance of DKN. Similar conclusions can be seen for the results of nearest-neighbor downsampling.

%, and even outperforms DKN for the Lu dataset. 

Figure~\ref{fig:depth_upsampling} shows a visual comparison of the upsampled depth images~(8$\times$). The better ability to extract common structures from the target and guidance images by our models here is clearly visible. Specifically, our results show a sharp depth transition without the texture-copying artifacts. In contrast, artifacts are clearly visible even in the results of DJFR, which tends to over-smooth the results and does not recover fine details. This confirms once more the advantage of using the weighted average with spatially-variant kernels and an adaptive neighborhood system in joint image filtering.

\begin{figure*}[t]
  \centering
  
  \subfloat[Input.]{
  \includegraphics[width=0.19\linewidth]{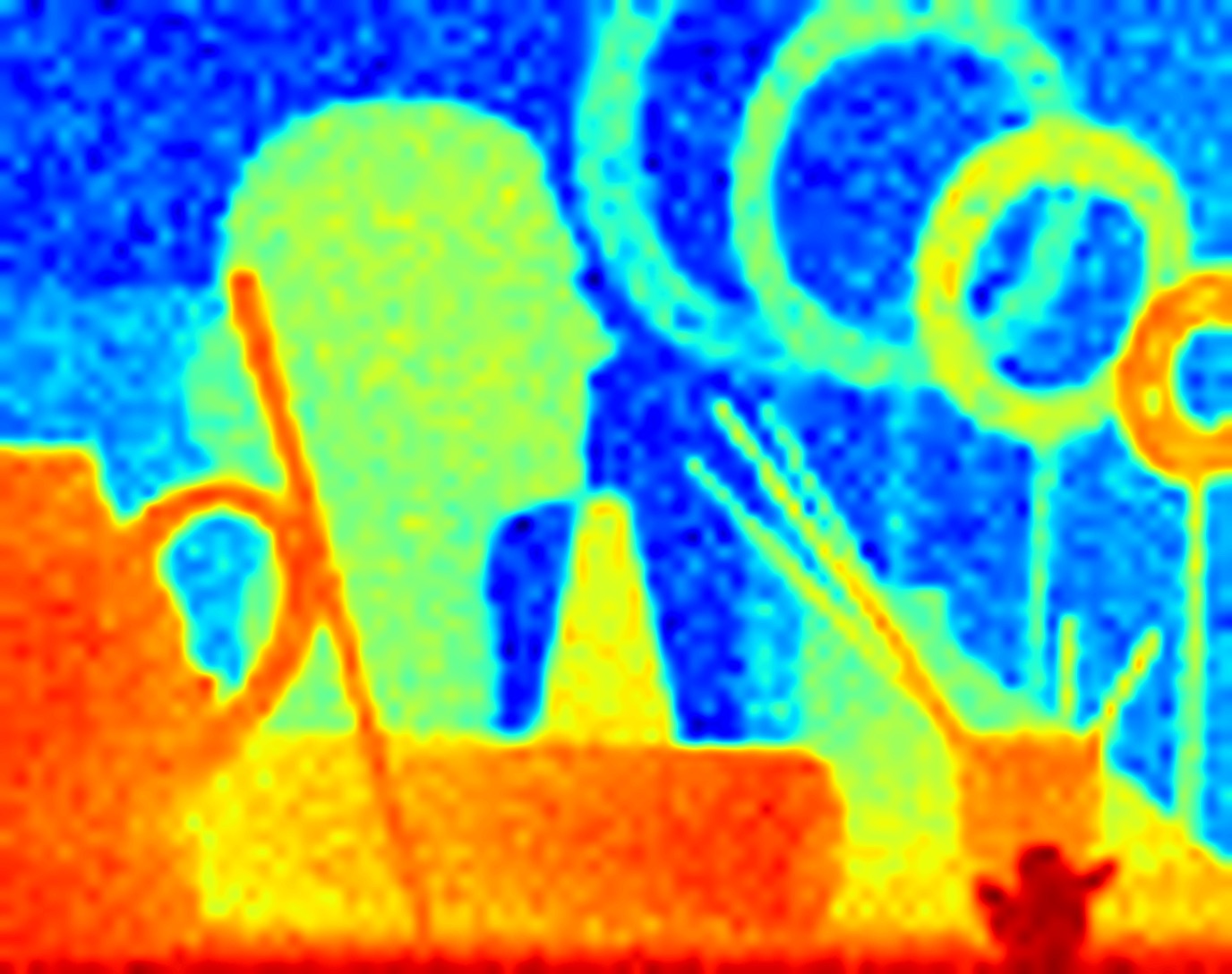} 
  }
  \subfloat[DMSG.]{
  \includegraphics[width=0.19\linewidth]{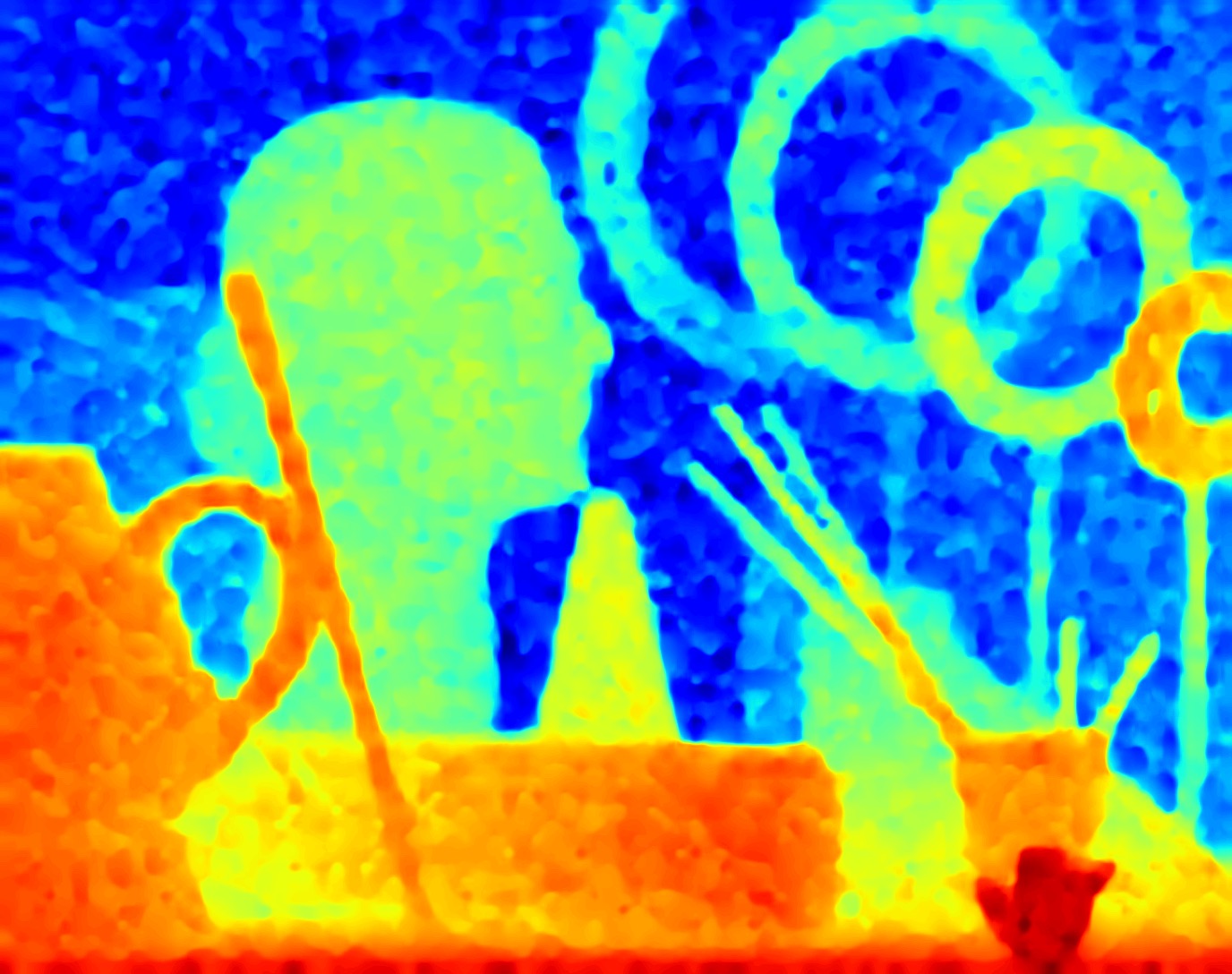} 
  }
  \subfloat[PAC.]{
  \includegraphics[width=0.19\linewidth]{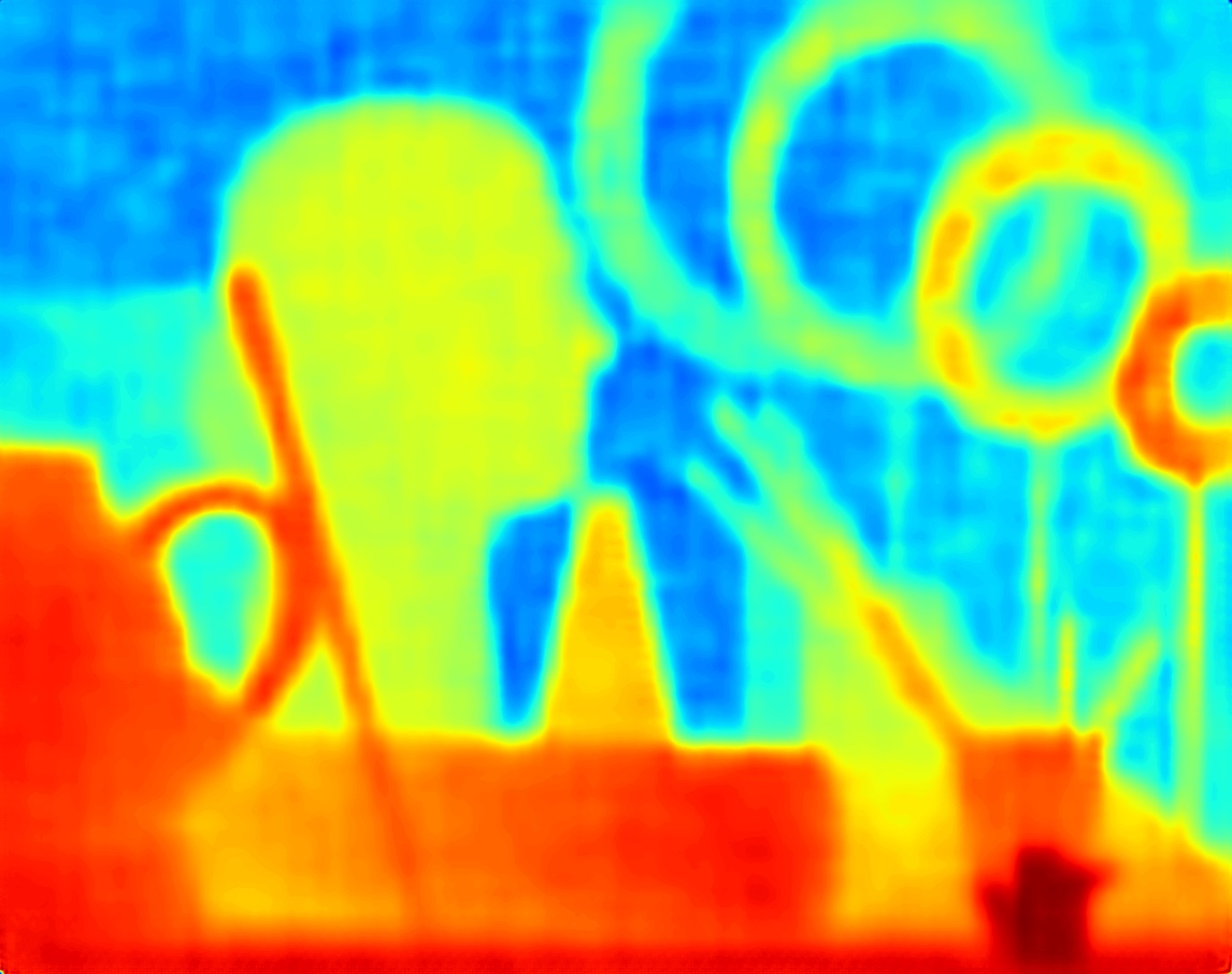} 
  }
  \subfloat[DJFR.]{
  \includegraphics[width=0.19\linewidth]{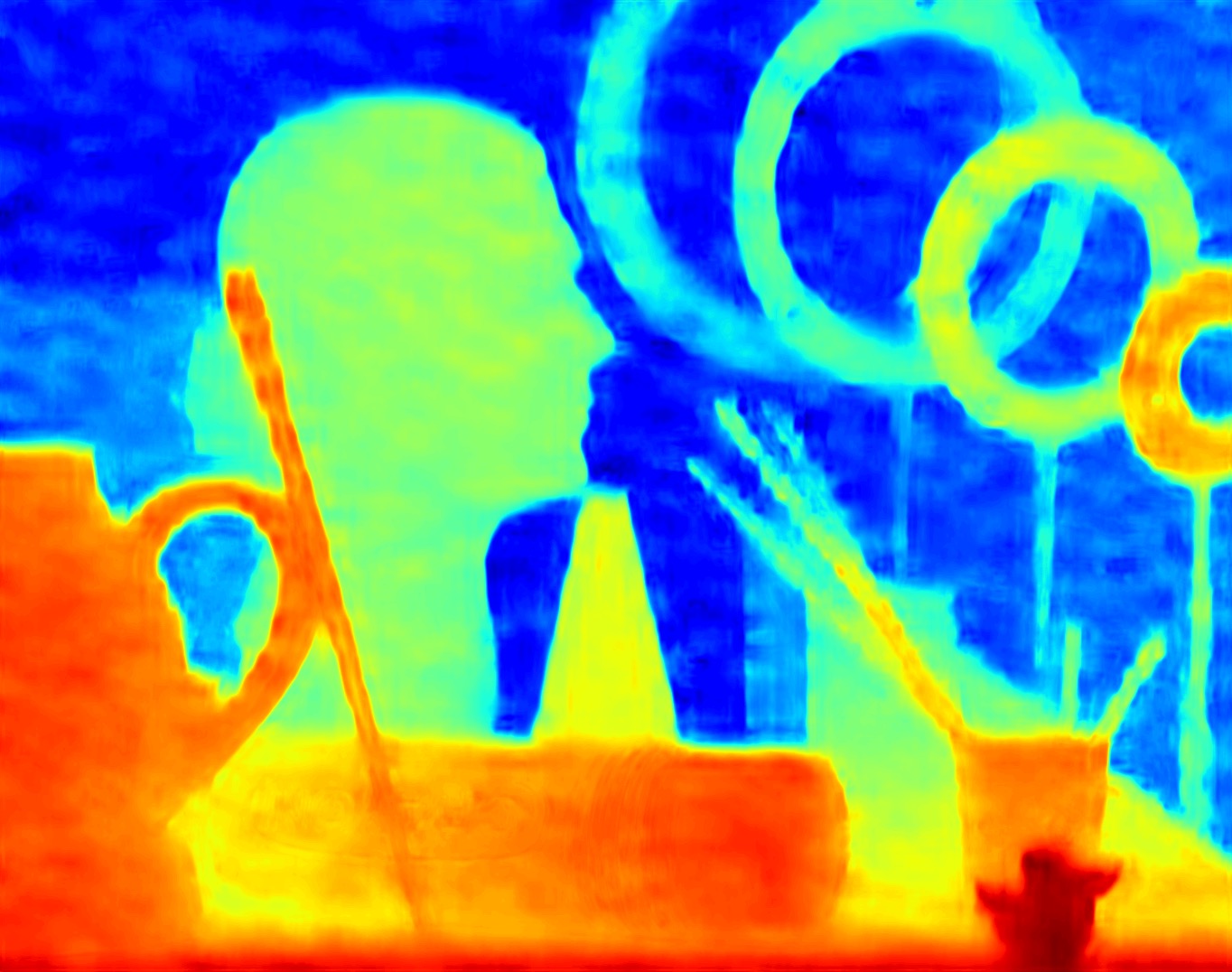} 
  }
  \subfloat[DKN.]{
  \includegraphics[width=0.19\linewidth]{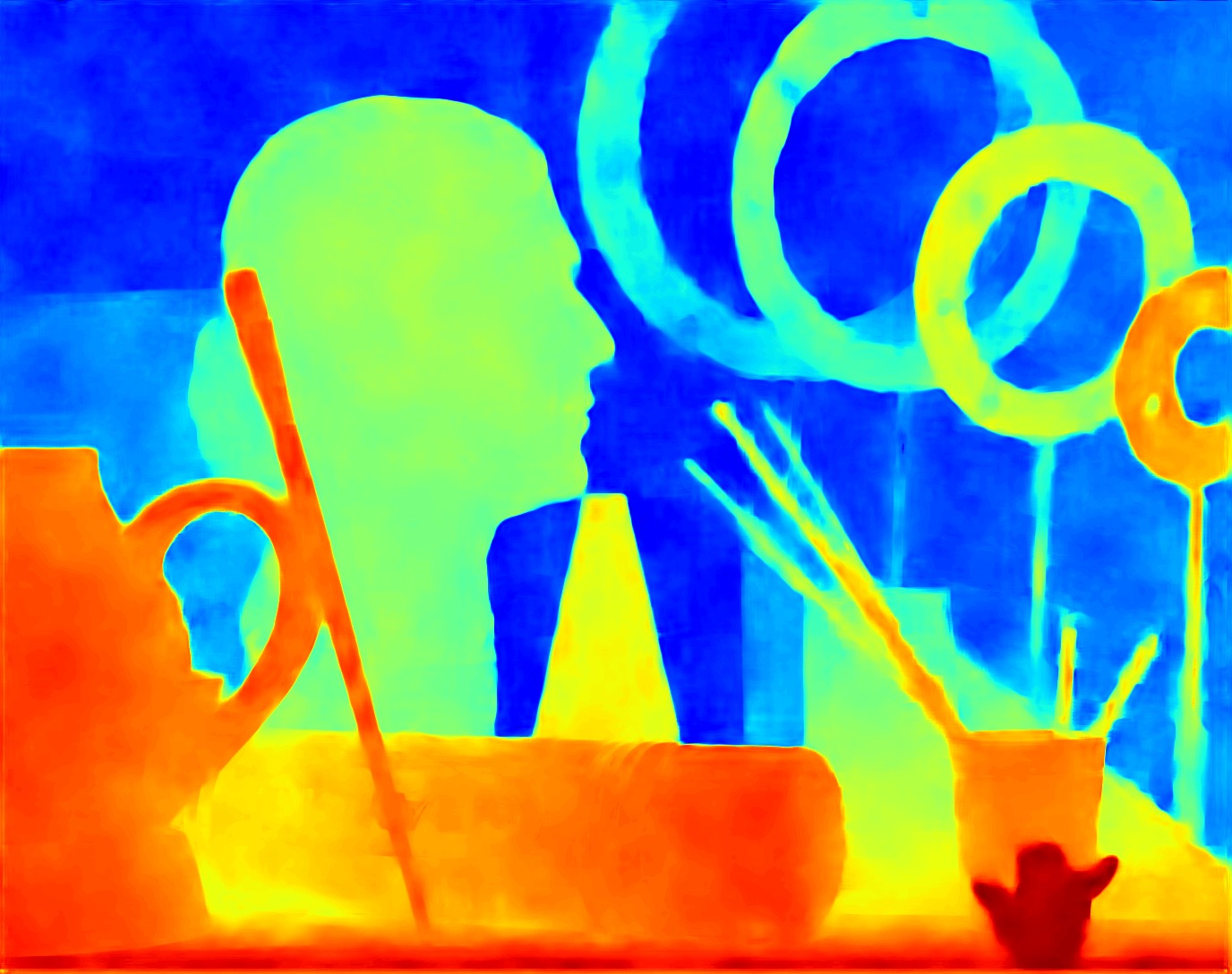} 
  }
  \begin{minipage}[b]{0.0195\linewidth}
%  \vspace*{1cm}
  \includegraphics[width=\linewidth]{images/vert_colorbar}
  \end{minipage}
   \caption{Visual comparison of noisy depth map upsampling~($8  \times$) on the art sequence in the noisy Middlebury dataset~\citep{park2011}: (a)~Input, (b)~DMSG~\citep{hui2016depth}, (c)~PAC~\citep{su2019pixel}, (d)~DJFR~\citep{li2017joint}, and (e)~DKN.}
  \label{fig:noisy-middlebury}
\vspace{-0.3cm}
\end{figure*}

\begin{figure*}[t]
  \centering
  \subfloat[Intensity.]{
  \begin{minipage}[b]{0.19\linewidth}
  	\includegraphics[width=\linewidth]{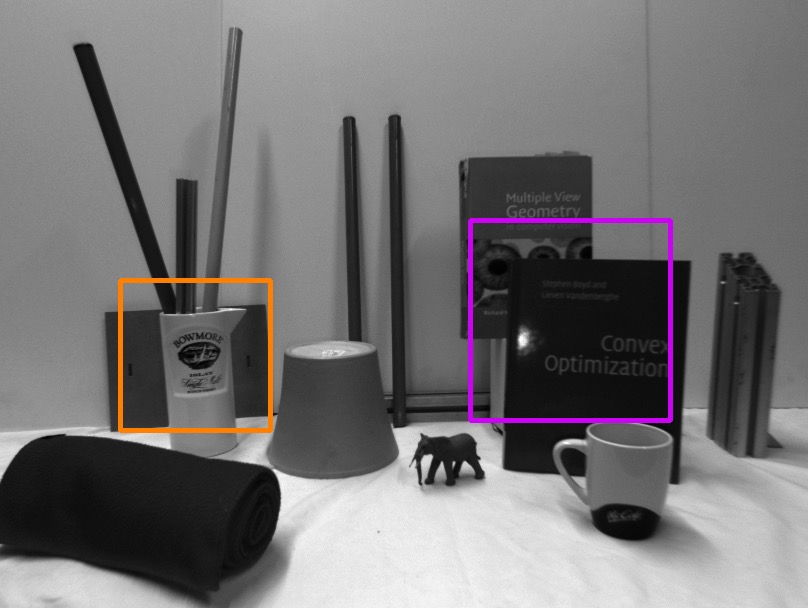} \\
  	\includegraphics[width=0.49\linewidth]{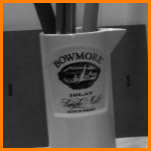}
  	\includegraphics[width=0.49\linewidth]{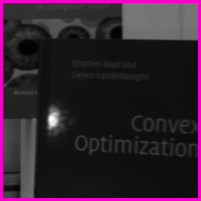}
  	\end{minipage}
  }
  \subfloat[LR ToF.]{
  \begin{minipage}[b]{0.19\linewidth}
  	\includegraphics[width=\linewidth]{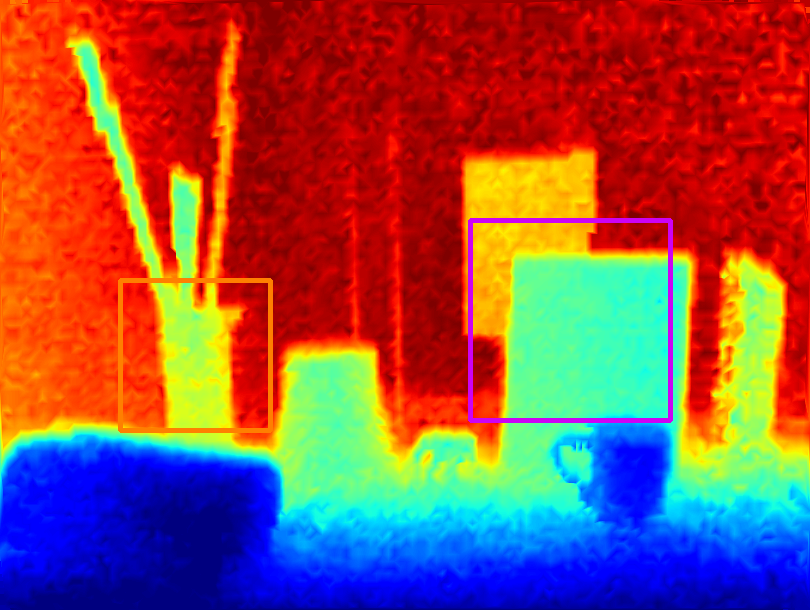} \\
  	\includegraphics[width=0.49\linewidth]{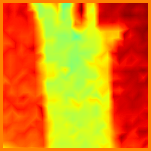}
  	\includegraphics[width=0.49\linewidth]{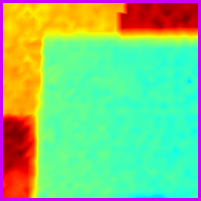}
  \end{minipage}
  }
  \subfloat[TGV.]{
  \begin{minipage}[b]{0.19\linewidth}
  	\includegraphics[width=\linewidth]{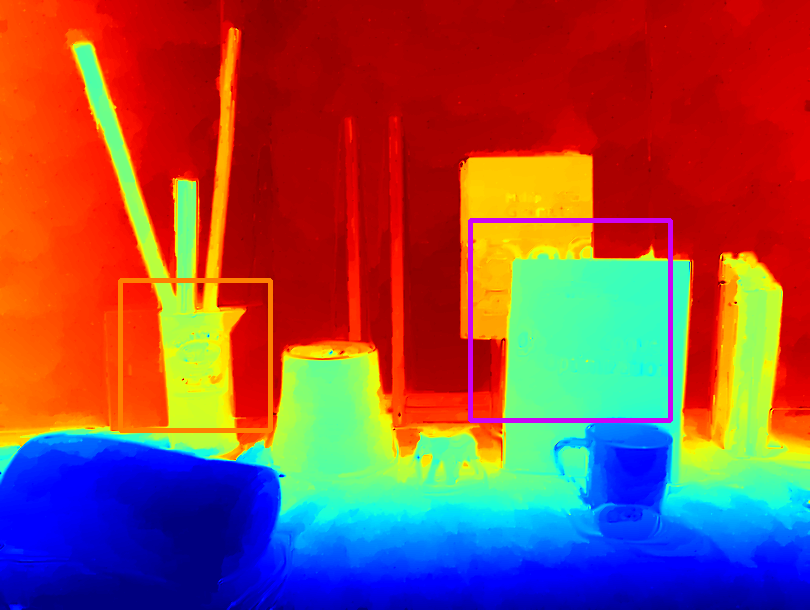} \\
  	\includegraphics[width=0.49\linewidth]{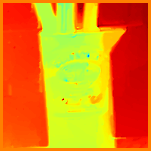}
  	\includegraphics[width=0.49\linewidth]{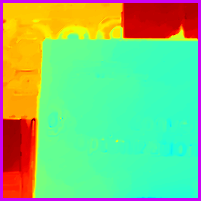}
  \end{minipage}
  }
  \subfloat[DKN.]{
  \begin{minipage}[b]{0.19\linewidth}
  	\includegraphics[width=\linewidth]{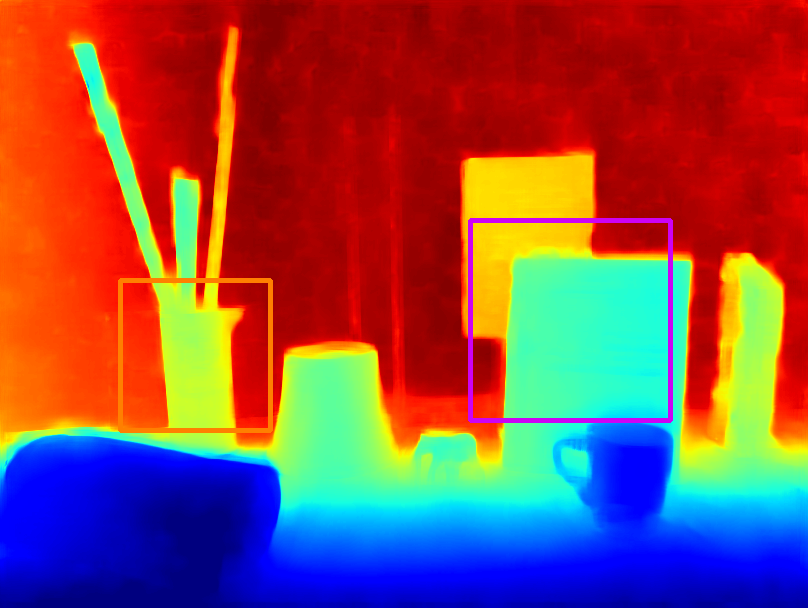} \\
  	\includegraphics[width=0.49\linewidth]{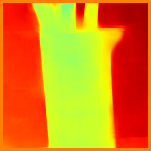}
  	\includegraphics[width=0.49\linewidth]{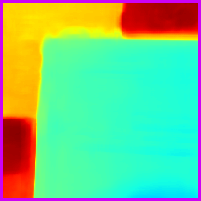}
  \end{minipage}
  }
  \subfloat[GT.]{
  \begin{minipage}[b]{0.19\linewidth}
  	\includegraphics[width=\linewidth]{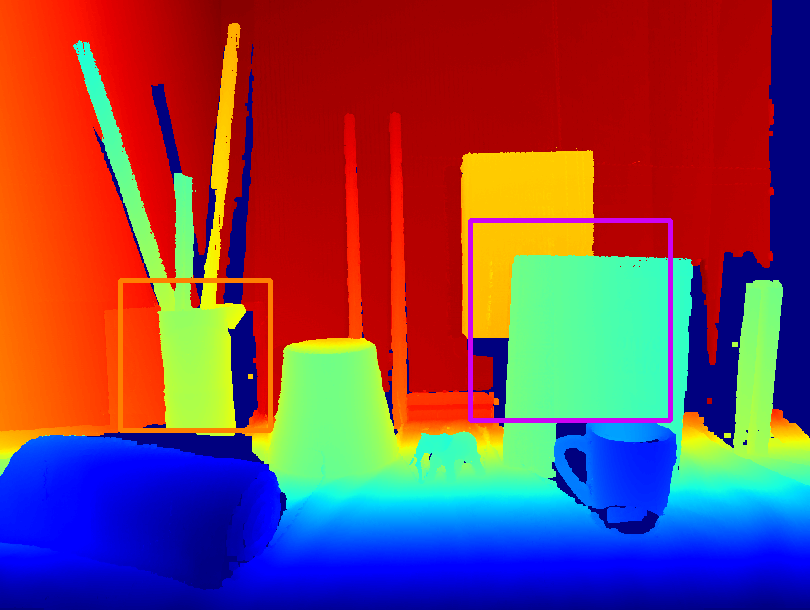} \\
  	\includegraphics[width=0.49\linewidth]{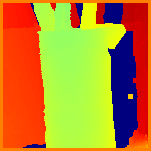}
  	\includegraphics[width=0.49\linewidth]{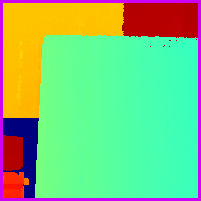}
  \end{minipage}
  }
  \begin{minipage}[b]{0.0305\linewidth}
%  \vspace*{1cm}
  \includegraphics[width=\linewidth]{images/vert_colorbar}
  \end{minipage}
   \caption{Visual comparison of noisy depth map upsampling on the books sequence in the ToFMark dataset~\citep{ferstl2013}: (a)~Intensity, (b)~LR ToF, (c)~TGV~\citep{ferstl2013}, (d)~DKN, and (e)~GT.}
  \label{fig:tofmark}
\vspace{-0.2cm}
\end{figure*}

We show in Fig.~\ref{fig:limitation} a failure example on the Lu dataset~\citep{lu2014depth}. As the color images in the dataset are captured in low-light conditions, the color boundaries become less reliable, and our model trained without the color image guidance,~DKN$^\dagger$, outperforms DKN. Specifically, the percentage of the image that DKN$^\dagger$ outperforms DKN~(8$\times$) is as follows: 43\%~(13/30), 100\%~(6/6), 2\%~(10/449), and 27\%~(238/864) for the Middlebury, Lu, NYU v2, and Sintel datasets, respectively. This suggests that using guidance images does not \emph{always} give better results. \emph{In general}, RGB guidance images provide structural information (e.g., gradients along occluding edges) useful in depth map upsampling, but they may also contain unrelated details (e.g., gradients along texture edges) that at time degrade performance.

\begin{table}[t]
\centering
\caption{Quantitative comparison of noisy depth map upsampling in terms of average RMSE on the ToFMark dataset~\citep{ferstl2013}. $\dagger$: The model trained with a synthetic dataset.}
\label{table:tofmark}
\addtolength{\tabcolsep}{-3.0pt}
%\scriptsize
\begin{tabular}[c]{l c c c}
\midrule
\multicolumn{1}{c}{Methods} & Books & Devil & Shark \\
\midrule
NN & 30.46 & 27.53 & 38.21 \\
Bilinear & 29.11 & 25.34 & 36.34 \\
JBU~\citep{kopf2007joint} & 27.82 & 25.30 & 34.79 \\
GF~\citep{he2013guided} & 27.11 & 23.45 & 33.26 \\
TGV~\citep{ferstl2013} & 24.00 & 23.19 & 29.89 \\
PDN$^\dagger$~\citep{riegler16gdsr} & \underline{23.74} & \underline{20.47} & \underline{28.81} \\
\midrule

DKN w/o Res. & \textbf{23.45} & \textbf{19.97} & \textbf{27.91} \\ %_20190923072511 epoch8

\midrule
\end{tabular}
\vspace{-0.4cm}
\end{table}

\begin{table}[t]
\centering
\caption{Quantitative comparison on saliency map upsampling in terms of weighted F-scores~\citep{margolin2014evaluate}. We use 5,168 images from the DUT-OMRON dataset~\citep{yang2013saliency}.}
%\scriptsize
\label{table:saliency}
\addtolength{\tabcolsep}{-4.5pt}
\begin{tabular}[c]{l c}
\midrule
\multicolumn{1}{c}{Methods} & Weighted-Fscore \\
\midrule
Bicubic Int. & 0.886 \\
GF~\citep{he2013guided} & 0.890 \\
SDF~\citep{ham2018robust} & 0.885 \\
DMSG~\citep{hui2016depth} & 0.938 \\
DJFR~\citep{li2017joint} & 0.925 \\ 
PAC~\citep{su2019pixel} & 0.943 \\
\midrule
DKN & \underline{0.944} \\ 
FDKN & \textbf{0.961} \\
  \midrule
  \end{tabular}
\vspace{-0.5cm}
\end{table}

\begin{figure*}[t]
  \centering
  \footnotesize	
  \subfloat[RGB image.]{
    \begin{minipage}[b]{0.123\linewidth}
      \includegraphics[width=\linewidth]{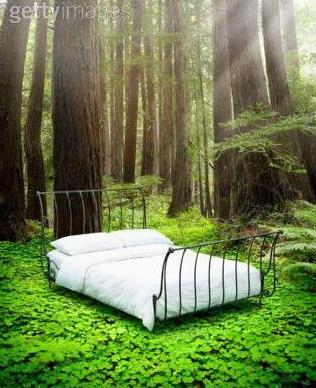}\vspace{0.05cm}
      \includegraphics[width=\linewidth]{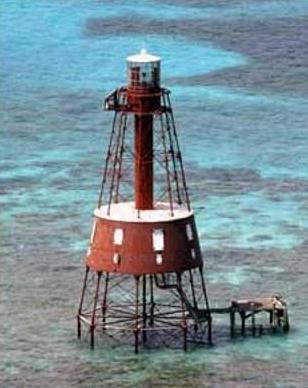} 
    \end{minipage}
  }\hspace{-0.2cm}  
  \subfloat[Bicubic Int.]{
    \begin{minipage}[b]{0.123\linewidth}
      \includegraphics[width=\linewidth]{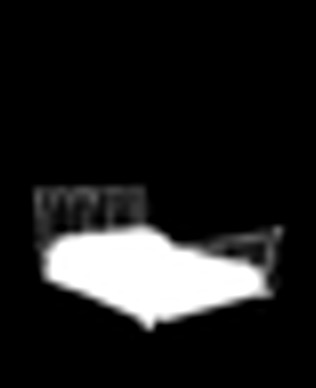}\vspace{0.05cm}
      \includegraphics[width=\linewidth]{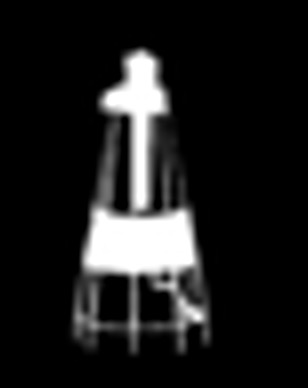} 
    \end{minipage}
  }\hspace{-0.2cm}  
  \subfloat[DMSG.]{
    \begin{minipage}[b]{0.123\linewidth}
      \includegraphics[width=\linewidth]{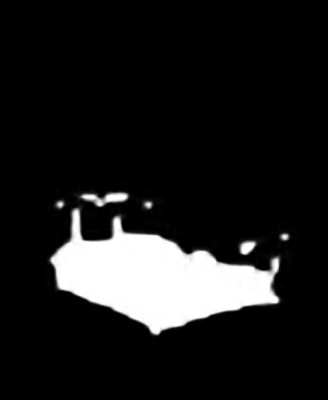}\vspace{0.05cm}
      \includegraphics[width=\linewidth]{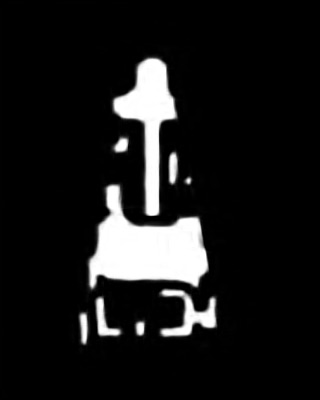} 
    \end{minipage}
  }\hspace{-0.2cm}
  \subfloat[PAC.]{
    \begin{minipage}[b]{0.123\linewidth}
      \includegraphics[width=\linewidth]{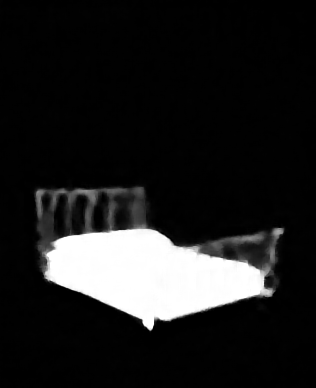}\vspace{0.05cm}
      \includegraphics[width=\linewidth]{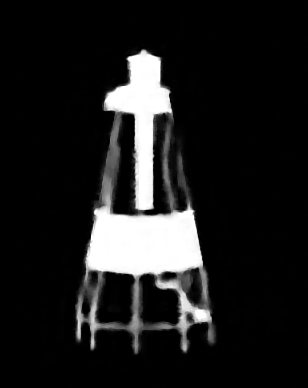} 
    \end{minipage}
  }\hspace{-0.2cm}  
  \subfloat[DJFR.]{
    \begin{minipage}[b]{0.123\linewidth}
      \includegraphics[width=\linewidth]{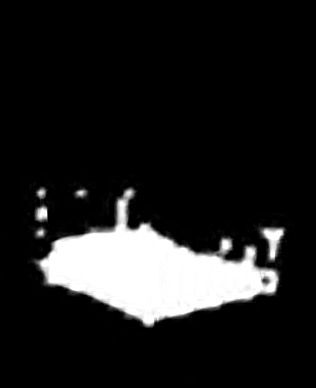}\vspace{0.05cm}
      \includegraphics[width=\linewidth]{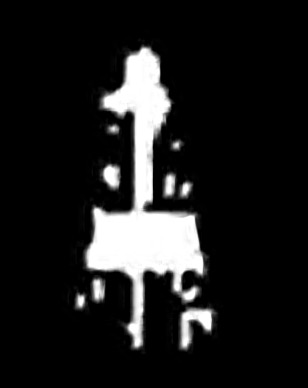} 
    \end{minipage}
  }\hspace{-0.2cm}  
  \subfloat[DKN.]{
    \begin{minipage}[b]{0.123\linewidth}
      \includegraphics[width=\linewidth]{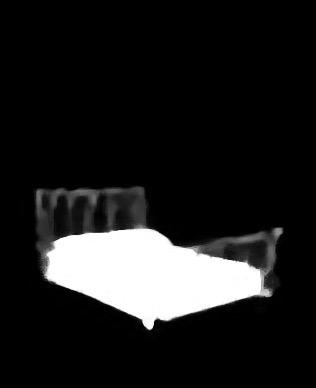}\vspace{0.05cm}
      \includegraphics[width=\linewidth]{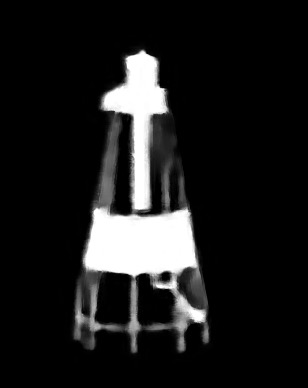} 
    \end{minipage}
  }\hspace{-0.2cm} 
  \subfloat[FDKN.]{
    \begin{minipage}[b]{0.123\linewidth}
      \includegraphics[width=\linewidth]{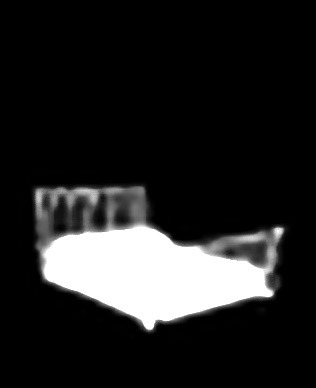}\vspace{0.05cm}
      \includegraphics[width=\linewidth]{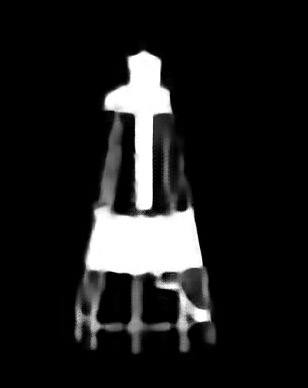} 
    \end{minipage}
  }\hspace{-0.2cm} 
  \subfloat[Ground truth.]{
    \begin{minipage}[b]{0.123\linewidth}
      \includegraphics[width=\linewidth]{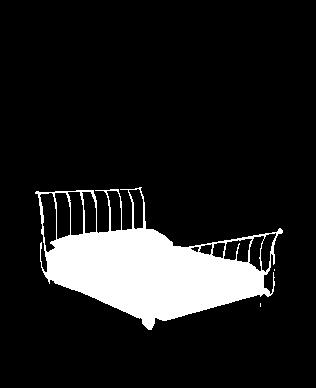}\vspace{0.05cm}
      \includegraphics[width=\linewidth]{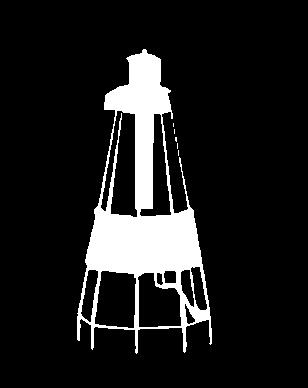} 
    \end{minipage}
  }\hspace{-0.2cm}  
  \caption{Visual comparison of saliency map upsampling~($8 \times$) on the DUT-OMRON dataset~\citep{yang2013saliency}: (a)~an RGB image, (b)~Bicubic Int., (c)~DMSG~\citep{hui2016depth}, (d)~PAC~\citep{su2019pixel}, (e)~DJFR~\citep{li2017joint}, (f)~DKN, (g)~FDKN, and (h)~ground truth.}
  \vspace{-.4cm}
  \label{fig:saliency}
\end{figure*}

\noindent{\textbf{Noisy depth map upsampling.}}
To show the robustness of our models on noisy data, we train our models using pairs of noisy low-resolution/ground-truth depth images in the NYU v2~\citep{silberman2012indoor} dataset. We simulate noisy low-resolution depth images following the protocol of~\citep{riegler16gdsr}. The other experimental settings are the same as in Section~\ref{exp:detail}. We compare our models with the state of the art in Table~\ref{table:noise-middlebury} including CNN-based methods~\citep{hui2016depth,li2017joint,su2019pixel,riegler16gdsr} on the noisy Middlebury dataset. To obtain the results of DMSG~\citep{hui2016depth}, DJFR~\citep{li2017joint}, and PAC~\citep{su2019pixel}, we retrain the models using the same image pairs as ours. The results of PDN are taken from~\citep{riegler16gdsr}\footnote{PDN uses synthetic depth maps created by a 3D renderer. The authors provide the source code online but despite our best efforts, we have not been able to retrain the corresponding models in the same setting as ours. 
We thus simply indicate the original results from~\citep{riegler16gdsr}}. In Table~\ref{table:tofmark}, we compare our model with other methods~\citep{kopf2007joint,he2013guided,ferstl2013,riegler16gdsr} on the ToFMark dataset, where all numbers are taken from~\citep{riegler16gdsr}. From these tables, we can see that our models outperform the state of the art, demonstrating that they are quite effective to handle synthetic and real noisy data, even better than PDN~\citep{riegler16gdsr} that requires many iterations to compute the primal-dual algorithm in testing. Figures~\ref{fig:noisy-middlebury} and~\ref{fig:tofmark} show visual comparisons on the noisy Middlebury and ToFmark datasets, respectively. We can see that our models suppress the noise, while preserving sharp depth boundaries~(e.g.,~thin sticks and small holes in Fig.~\ref{fig:noisy-middlebury}) and being robust to texture-copying artifacts~(e.g.,~texts in a vase and books in Fig.~\ref{fig:tofmark}).

\begin{figure*}[t]
  \centering
\footnotesize	
  \subfloat[{Guidance.}]{
    \begin{minipage}[b]{0.122\linewidth}
      \includegraphics[width=\linewidth,frame]{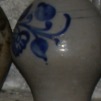}\vspace{0.05cm}
      \includegraphics[width=\linewidth,frame]{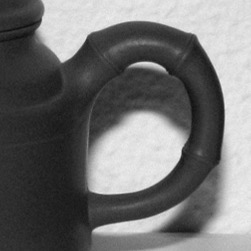}
    \end{minipage}
  }\hspace{-0.2cm}
  \subfloat[{Target.}]{
    \begin{minipage}[b]{0.122\linewidth}
      \includegraphics[width=\linewidth,frame]{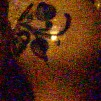}\vspace{0.05cm}
      \includegraphics[width=\linewidth,frame]{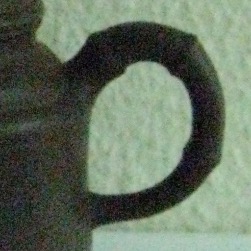}
    \end{minipage}
  }\hspace{-0.2cm}
  \subfloat[GF.]{
    \begin{minipage}[b]{0.122\linewidth}
      \includegraphics[width=\linewidth,frame]{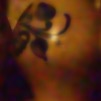}\vspace{0.05cm}
      \includegraphics[width=\linewidth,frame]{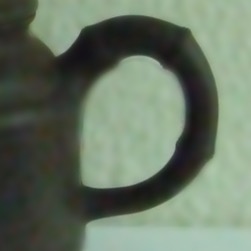}
    \end{minipage}
  }\hspace{-0.2cm}
  \subfloat[SDF.]{
    \begin{minipage}[b]{0.122\linewidth}
      \includegraphics[width=\linewidth,frame]{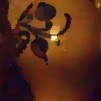}\vspace{0.05cm}
      \includegraphics[width=\linewidth,frame]{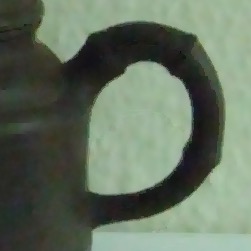}
    \end{minipage}
  }\hspace{-0.2cm}
  \subfloat[Yan.]{
    \begin{minipage}[b]{0.122\linewidth}
      \includegraphics[width=\linewidth,frame]{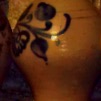}\vspace{0.05cm}
      \includegraphics[width=\linewidth,frame]{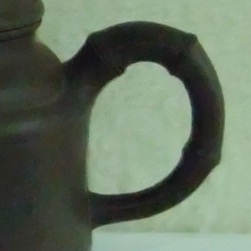}
    \end{minipage}
  }\hspace{-0.2cm}
  \subfloat[DJF.]{
    \begin{minipage}[b]{0.122\linewidth}
      \includegraphics[width=\linewidth,frame]{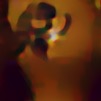}\vspace{0.05cm}
      \includegraphics[width=\linewidth,frame]{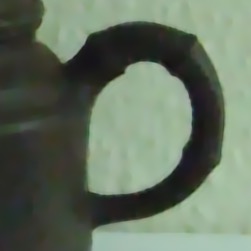}
    \end{minipage}
  }\hspace{-0.2cm}
  \subfloat[{DKN}.]{
    \begin{minipage}[b]{0.122\linewidth}
      \includegraphics[width=\linewidth,frame]{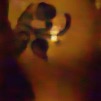}\vspace{0.05cm}
      \includegraphics[width=\linewidth,frame]{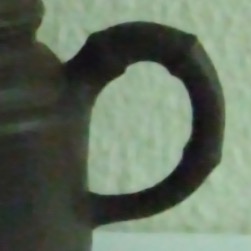}
    \end{minipage}
  }\hspace{-0.2cm}
  \subfloat[{FDKN}.]{
    \begin{minipage}[b]{0.122\linewidth}
      \includegraphics[width=\linewidth,frame]{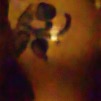}\vspace{0.05cm}
      \includegraphics[width=\linewidth,frame]{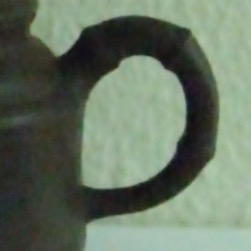}
    \end{minipage}
  }\hspace{-0.2cm}
  \caption{Examples of cross-modality noise reduction for (top) flash/non-flash denoising and (bottom) RGB/NIR denoising: (a-b)~Guidance and target images, (c)~GF~\citep{he2013guided}, (d)~SDF~\citep{ham2018robust}, (e)~Yan~\citep{yan2013cross}, (f)~DJF~\citep{li2016deep}, (g)~DKN, and (h)~FDKN. Our models preserve textures while smoothing noise. GF and DJF tend to over-smooth the textures. Artifacts are clearly visible in the results of SDF. The method of~\citep{yan2013cross}, specially designed for this task, gives the best results.} 
  \vspace{-.3cm}
  \label{fig:noise-reduction}
\end{figure*}

\noindent{\textbf{Saliency map upsampling.}} 
To evaluate the generalization ability of our models for depth map upsampling on other tasks, we apply them trained with the NYU v2 dataset to saliency map upsampling without fine-tuning. We downsample saliency maps~($\times 8$) in the DUT-OMRON dataset~\citep{yang2013saliency}, and then upsample them under the guidance of high-resolution color images. We show in Table~\ref{table:saliency} a comparison of weighted F-measure \citep{margolin2014evaluate} between upsampled images and the ground truth. Figure~\ref{fig:saliency} shows examples of the upsampling results by the state of the art and our models. The results show that our models outperform others including CNN-based methods~\citep{li2017joint,hui2016depth,su2019pixel}.

\begin{figure*}[h!]
  \centering
  \footnotesize	
  \subfloat[\scriptsize{Input image.}]{
    \begin{minipage}[b]{0.251\linewidth} 
      \includegraphics[width=\linewidth, frame]{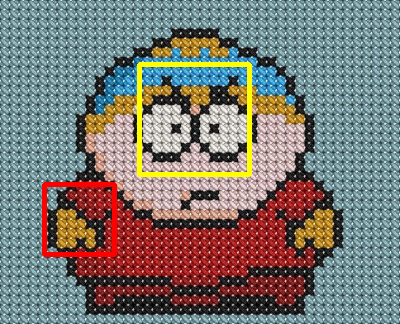}\vspace{0.05cm}
      \includegraphics[width=\linewidth, frame]{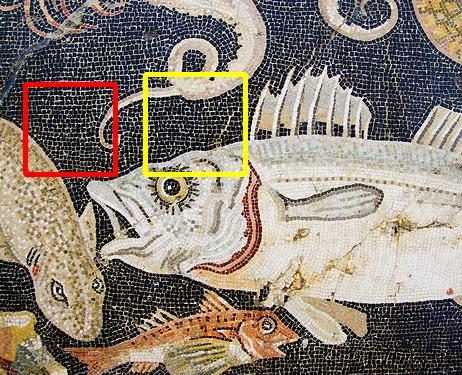}
    \end{minipage}
  }
  \subfloat[\scriptsize{RGF.}]{
    \begin{minipage}[b]{0.1\linewidth} 
      \includegraphics[width=\linewidth, frame]{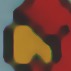} 
      \includegraphics[width=\linewidth, frame]{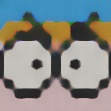}\vspace{0.05cm}
      \includegraphics[width=\linewidth, frame]{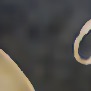}
      \includegraphics[width=\linewidth, frame]{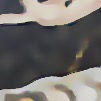} 
    \end{minipage}
  }\hspace{-0.18cm}
  \subfloat[\scriptsize{RTV.}]{
    \begin{minipage}[b]{0.1\linewidth} 
      \includegraphics[width=\linewidth, frame]{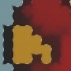} 
      \includegraphics[width=\linewidth, frame]{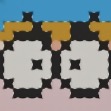}\vspace{0.05cm}
      \includegraphics[width=\linewidth, frame]{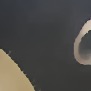}
      \includegraphics[width=\linewidth, frame]{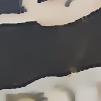} 
    \end{minipage}
  }\hspace{-0.18cm}
  \subfloat[\scriptsize{Cov.}]{
    \begin{minipage}[b]{0.1\linewidth} 
      \includegraphics[width=\linewidth, frame]{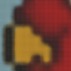} 
      \includegraphics[width=\linewidth, frame]{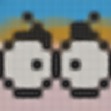}\vspace{0.05cm}
      \includegraphics[width=\linewidth, frame]{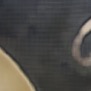}
      \includegraphics[width=\linewidth, frame]{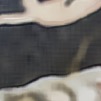} 
    \end{minipage}
  }\hspace{-0.18cm}
  \subfloat[\scriptsize{SDF.}]{
    \begin{minipage}[b]{0.1\linewidth} 
      \includegraphics[width=\linewidth, frame]{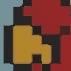} 
      \includegraphics[width=\linewidth, frame]{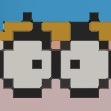}\vspace{0.05cm}
      \includegraphics[width=\linewidth, frame]{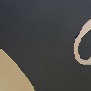}
      \includegraphics[width=\linewidth, frame]{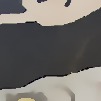} 
    \end{minipage}
  }\hspace{-0.18cm}
  \subfloat[\scriptsize{DJF.}]{
    \begin{minipage}[b]{0.1\linewidth} 
      \includegraphics[width=\linewidth, frame]{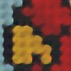} 
      \includegraphics[width=\linewidth, frame]{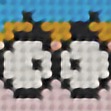}\vspace{0.05cm}
      \includegraphics[width=\linewidth, frame]{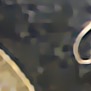}
      \includegraphics[width=\linewidth, frame]{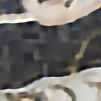} 
    \end{minipage}
  }\hspace{-0.18cm}
  \subfloat[\scriptsize{DKN}.]{
    \begin{minipage}[b]{0.1\linewidth} 
      \includegraphics[width=\linewidth, frame]{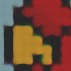} 
      \includegraphics[width=\linewidth, frame]{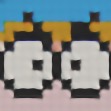}\vspace{0.05cm}
      \includegraphics[width=\linewidth, frame]{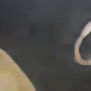}
      \includegraphics[width=\linewidth, frame]{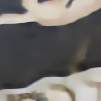} 
    \end{minipage}
  }\hspace{-0.18cm}
  \subfloat[\scriptsize{FDKN}.]{
    \begin{minipage}[b]{0.1\linewidth} 
      \includegraphics[width=\linewidth, frame]{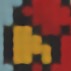} 
      \includegraphics[width=\linewidth, frame]{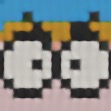}\vspace{0.05cm}
      \includegraphics[width=\linewidth, frame]{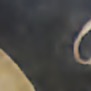}
      \includegraphics[width=\linewidth, frame]{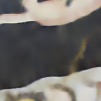} 
    \end{minipage}
  }\hspace{-0.18cm}
  \caption{Visual comparison of texture removal for regular (top) and irregular (bottom) textures: (a)~an input image, (b)~RGF~\citep{zhang2014rolling}, (c)~RTV~\citep{xu2012structure}, (d)~Cov~\citep{karacan2013structure}, (e)~SDF~\citep{ham2018robust}, (f)~DJF~\citep{li2016deep}, (g)~DKN, and (h)~FDKN.}
  \label{fig:texture_removal}
\vspace{-0.3cm}
\end{figure*}

\vspace{-0.5cm}
\subsection{Cross-modality image restoration and texture removal experiments}~\label{exp:other}
\vspace{-0.5cm}

\vspace{-0.7cm}
\subsubsection{Experimental details}
\vspace{-0.3cm}
Following~\citep{li2016deep,li2017joint}, we use depth denoising as a proxy task for cross-modality image restoration and texture removal, since ground-truth datasets for these tasks are not available. We train our models for denoising depth images with RGB/D image pairs from the NYU v2 dataset~\citep{silberman2012indoor}. The models for depth noise removal are similarly trained to those for joint image upsampling in Section~\ref{exp:detail} under the guidance of high-resolution RGB images but with 4k iterations. Noisy depth images are synthesized by adding Gaussian noise with zero mean and variance of $0.005$. The models are then applied to the tasks of cross-modality image restoration and texture removal without fine-tuning. For comparison, average RMSE for GF~\citep{he2013guided}, Yan~\citep{yan2013cross}, SDF~\citep{ham2018robust}, DJF~\citep{li2016deep}, and DKN on depth noise removal are 5.34, 12.53, 7.56, 2.63, and 2.46, respectively, in the test split of the NYU v2 dataset, showing that our model again outperforms the others. We do not use the residual connection for noise removal tasks, since we empirically find that it does not help in this case. We only show qualitative results in these tasks, since ground truth is not available. All previous works~(e.g.,~\citep{li2016deep,yan2013cross,li2017joint,he2013guided}) we are aware of for these tasks offer qualitative results only.

\vspace{-0.3cm}
\subsubsection{Results}
\vspace{-0.4cm}
\noindent{\textbf{Cross-modality image restoration.}}
For flash/non-flash denoising, we set the flash and non-flash images as guidance and target ones, respectively. Similarly, we restore the color image guided by the flash NIR image in RGB/NIR denoising. Examples for flash/non-flash and RGB/NIR restoration are shown in Fig.~\ref{fig:noise-reduction}. Qualitatively, our models outperform other state-of-the-art methods~\citep{li2016deep,ham2018robust,he2013guided}. For example, GF~\citep{he2013guided} using guidance images only cannot deal with gradient reversal in flash NIR images, resulting in smoothed edges. SDF~\citep{ham2018robust} and DJF~\citep{li2016deep} use both guidance and target images. However, SDF tends to enhance edges, giving over-sharpened results, while DJF overly smooths images. In contrast, our models preserve edges while smoothing noise without artifacts. This demonstrates that our models trained with RGB/D images can generalize well for others with different modalities. The method of~\citep{yan2013cross}, specially designed for this task, gives the best results.

\begin{figure*}[t]
\vspace{-0.1cm}
  \centering
  \footnotesize	
  \subfloat[RGB image.]{
    \begin{minipage}[b]{0.16\linewidth}
      \includegraphics[width=\linewidth]{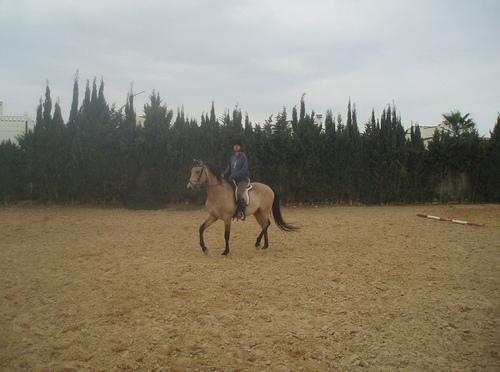}\vspace{0.05cm}
      \includegraphics[width=\linewidth]{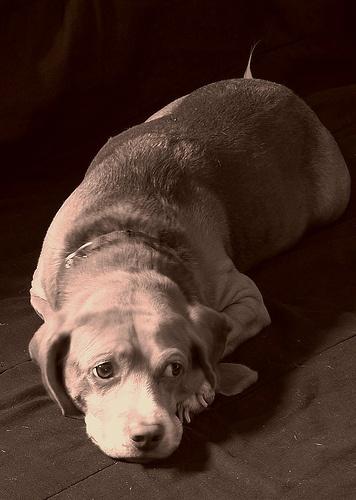}
    \end{minipage}
  }\hspace{-0.2cm}  
  \subfloat[Baseline.]{
    \begin{minipage}[b]{0.16\linewidth}
      \includegraphics[width=\linewidth]{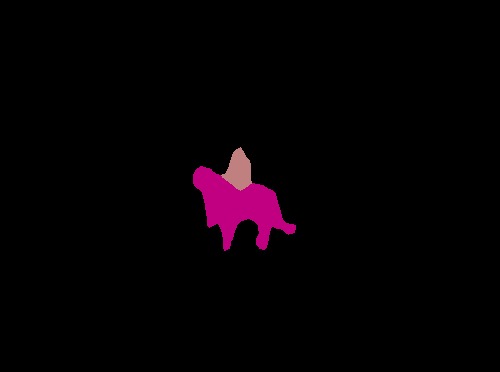}\vspace{0.05cm}
      \includegraphics[width=\linewidth]{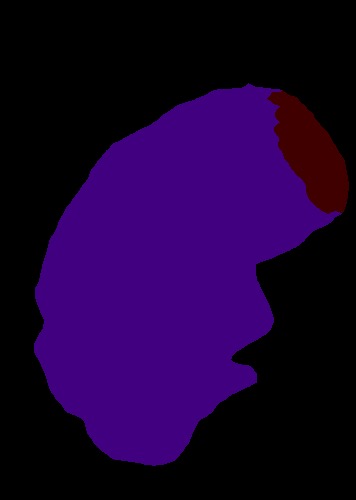}
    \end{minipage}
  }\hspace{-0.2cm}  
  \subfloat[DenseCRF.]{
    \begin{minipage}[b]{0.16\linewidth}
      \includegraphics[width=\linewidth]{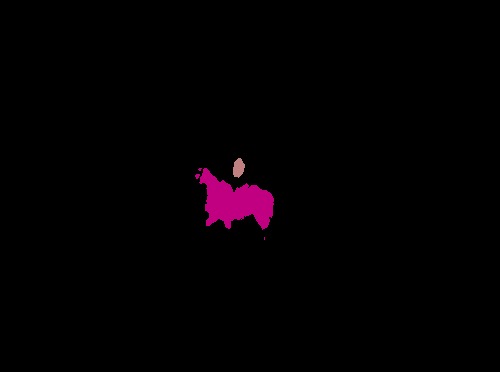}\vspace{0.05cm}
      \includegraphics[width=\linewidth]{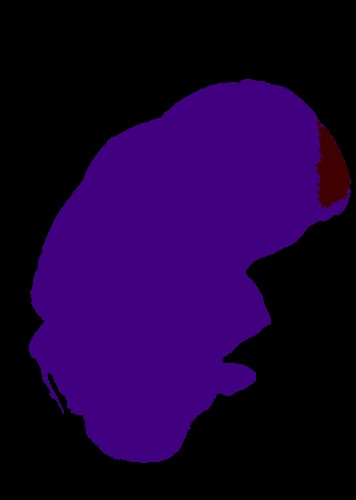}
    \end{minipage}
  }\hspace{-0.2cm}  
  \subfloat[DGF.]{
    \begin{minipage}[b]{0.16\linewidth}
      \includegraphics[width=\linewidth]{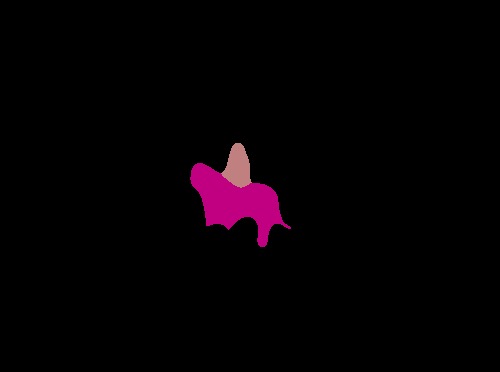}\vspace{0.05cm}
      \includegraphics[width=\linewidth]{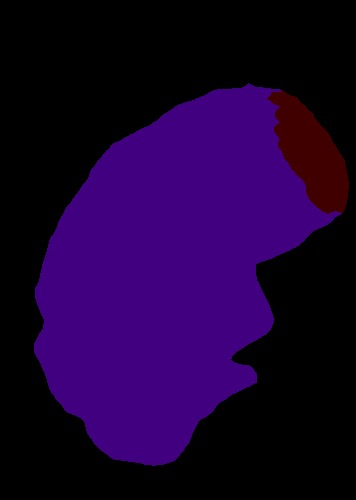}
    \end{minipage}
  }\hspace{-0.2cm}  
  \subfloat[FDKN.]{
    \begin{minipage}[b]{0.16\linewidth}
      \includegraphics[width=\linewidth]{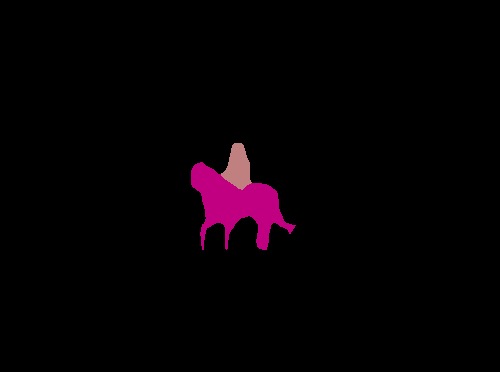}\vspace{0.05cm}
      \includegraphics[width=\linewidth]{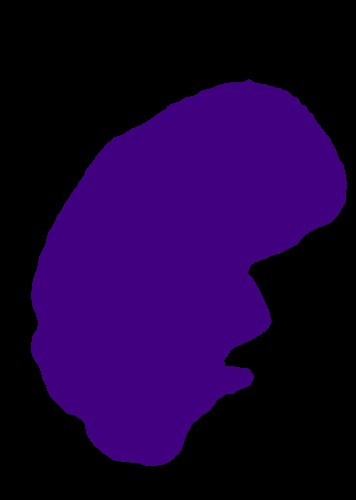} 
    \end{minipage}
  }\hspace{-0.2cm}
  \subfloat[Ground truth.]{
    \begin{minipage}[b]{0.16\linewidth}
      \includegraphics[width=\linewidth]{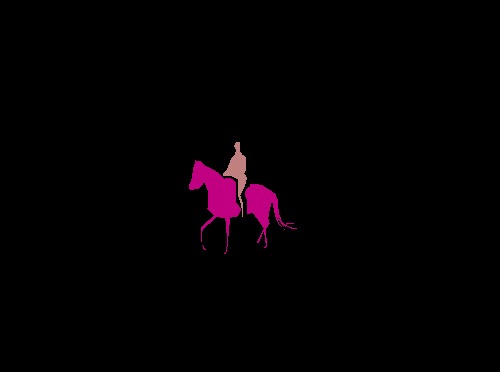}\vspace{0.05cm}
      \includegraphics[width=\linewidth]{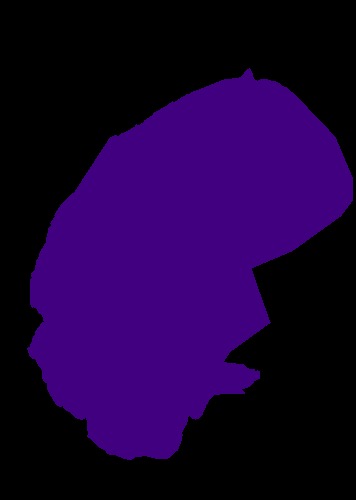} 
    \end{minipage}
  }\hspace{-0.2cm}    
  \caption{Visual comparison of semantic segmentation on the validation set of Pascal VOC 2012 benchmark~\citep{pascalvoc2012}: (a)~an RGB image, (b)~baseline~\citep{chen2018deeplab}, (c)~denseCRF~\citep{krahenbuhl2011efficient}, (d)~DGF~\citep{wu2018fast}, (e)~FDKN, and (f)~ground truth. Compared to the state of the art, our model shows better ability to improve the localization accuracy of object boundaries and refine incorrectly labeled segments.}
  \label{fig:segmentation}
%\vspace{-0.4cm}
\end{figure*}

\noindent{\textbf{Texture removal.}}
We set the textured image itself for guidance and target, and apply our models repeatedly to remove small-scale textures. We show examples in Fig.~\ref{fig:texture_removal}. Our models outperform DJF~\citep{li2016deep} trained for denoising depth images, suggesting that a generalization ability is better. In particular, they remove textures without artifacts while maintaining other high-frequency structures such as image boundaries and corners. Our models give results comparable to other methods including RTV~\citep{xu2012structure} and Cov~\citep{karacan2013structure} specially designed for texture removal. A plausible explanation of why networks trained for denoising depth images work for texture removal is that textures can be considered as patterned noise. Repeatedly applying our networks thus removes them. A similar finding can be found in~\citep{li2016deep}. We empirically find that 4 iterations are enough to get satisfactory results, and use the same number of iterations for all experiments.

\begin{table}[t]
\centering
\caption{Quantitative comparison on semantic segmentation in terms of average IoU. We use 1,449 images from the validation set in the Pascal VOC~2012 benchmark~\citep{pascalvoc2012}.}
\label{table:segmentation}
\addtolength{\tabcolsep}{-3.0pt}
%\scriptsize
\begin{tabular}[c]{l c}
\midrule
\multicolumn{1}{c}{Methods} & Mean IoU \\
\midrule
Baseline~\citep{chen2018deeplab} & 70.69 \\
DenseCRF~\citep{krahenbuhl2011efficient} & 71.98 \\
DGF~\citep{wu2018fast} & \underline{72.96} \\ 
\midrule
FDKN & \textbf{73.60} \\
\midrule
  \end{tabular}
\vspace{-0.45cm}
\end{table}

\vspace{-0.6cm}
\subsection{Semantic segmentation experiments}\label{exp:ss}
\vspace{-0.4cm}
CNNs commonly use max pooling and downsampling to achieve invariance, but this degrades localization accuracy especially at object boundaries~\citep{long2015fully}. DeepLab~\citep{chen2018deeplab} overcomes this problem using probabilistic graphical models. It applies a fully connected CRF~\citep{krahenbuhl2011efficient} to the response of the final layer of a CNN. Zheng et al. interpret CRFs as recurrent neural networks which are then plugged into as a part of a CNN~\citep{zheng2015conditional}, making it possible to train the whole network end-to-end. Recently, Wu et al. have proposed a layer to integrate guided filtering~\citep{he2013guided} into CNNs~\citep{wu2018fast}. Instead of using CRFs~\citep{chen2018deeplab,krahenbuhl2011efficient,zheng2015conditional} or guided image filtering~\citep{he2013guided}, we apply the FDKN to the response of the final layer of DeepLab~v2~\citep{chen2018deeplab} for semantic segmentation. %Note that FDKN is fully differentiable and the whole network is trained end-to-end. 

\vspace{-0.2cm}
\subsubsection{Experimental details}
\vspace{-0.1cm}
Following the experimental protocol of~\citep{wu2018fast}, we plug FDKN into DeepLab-v2~\citep{chen2018deeplab}, which uses ResNet-101~\citep{he2016residual} pretrained for ImageNet classification, as a part of CNNs for semantic segmentation, instead of applying a fully connected conditional random field~(CRF)~\citep{krahenbuhl2011efficient} to refine segmentation results. That is, we integrate DeepLab-v2 and our model and train the whole network end-to-end, avoiding an offline post-processing using CRFs. We use the Pascal VOC 2012 dataset~\citep{pascalvoc2012} that contains $1,464$, $1,449$, and $1,456$ images for training, validation and test, respectively. Following~\citep{chen2018deeplab,wu2018fast}, we augment the training dataset by the annotations provided by~\citep{BharathICCV2011}, resulting in $10,582$ images, and use $1,449$ images in the validation set for evaluation. We train the network using a softmax log loss with a batch size of 1 for 20k iterations. The SGD optimizer with momentum of $0.9$ is used. As learning rate, we use the scheduling method of~\citep{chen2018deeplab} with learning rate of $2.5\times10^{-4}$ and $2.5\times 10^{-3}$ for DeepLab-v2 and FDKN, respectively. We upsample 21-channel outputs~(20 object classes and background) of DeepLab-v2 before a softmax layer using a high-resolution color image. We apply the FDKN separately in each channel.

%\footnote{We use a PyTorch version available online:~\url{https://github.com/isht7/pytorch-deeplab-resnet}}

\begin{table*}[http]
\centering
%\vspace{-.2cm}
\caption{Average RMSE comparison~(DKN/FDKN) of different components and size of kernels~(from $3 \times 3$ to $25 \times 25$). From the third row, we can see that aggregating pixels from a $15 \times 15$ window is enough. We thus restrict the maximum range of offset locations to $15 \times 15$. For example, results for $7 \times 7$ in the forth row are computed using $49$ pixels sparsely sampled from a $15 \times 15$ window. We omit the results for $15\times 15$, $19\times 19$ and $25\times 25$ kernels, since they are equal to or beyond the maximum range of offset locations. For each network, numbers in bold indicate the best performance and underscored ones are the second best.}
%\vspace{-.1cm}
\label{table:ablation-archicture}
\addtolength{\tabcolsep}{-3.0pt}
\newcolumntype{C}[1]{>{\centering\arraybackslash}p{#1}}
\begin{tabular}[c]{c c c c c C{0.8cm}@{/} C{0.8cm} C{0.8cm}@{/} C{0.8cm} C{0.8cm}@{/} C{0.8cm} C{0.8cm}@{/} C{0.8cm} C{0.8cm}@{/} C{0.8cm} C{0.8cm}@{/} C{0.8cm}} 
\midrule
\multicolumn{2}{c}{Weight learning}  & \multicolumn{2}{c}{Offset learning} & \multirow{2}{*}{Res.} & \multicolumn{2}{c}{\multirow{2}{*}{$3 \times 3$}} & \multicolumn{2}{c}{\multirow{2}{*}{$5 \times 5$}} & \multicolumn{2}{c}{\multirow{2}{*}{$7 \times 7$}} & \multicolumn{2}{c}{\multirow{2}{*}{$15 \times 15$}} & \multicolumn{2}{c}{\multirow{2}{*}{$19 \times 19$}} & \multicolumn{2}{c}{\multirow{2}{*}{$ 25 \times 25$}}\\
\cmidrule(lr){1-2} \cmidrule(lr){3-4}  RGB  & Depth  &  RGB  & Depth   \\ 
\cmidrule{1-17}
$\checkmark$ & & & & & 5.92 & 6.05 & 5.52 & 5.73 & 5.43 & 5.67 & 5.59 & 5.74 & 5.82 & 5.81 & 6.21 &5.99 \\
& $\checkmark$ & & & & 5.24 & 5.30 & 4.36 & 4.47 & 4.09 & 4.24 & 4.09 & 4.17 & 4.11 & 4.18 & 4.15 &4.21 \\
$\checkmark$ & $\checkmark$ &  & & & 5.03 & 5.14 & 3.90 & 4.16 & 3.48 & 3.80 & 3.32 & 3.66 & 3.33 & 3.66 & 3.39 & 3.72 \\
$\checkmark$ & & $\checkmark$ & & & 5.37 & 5.18 & 5.38 & 5.09 & 5.40 & 5.07 & \multicolumn{2}{c}{--} & \multicolumn{2}{c}{--} & \multicolumn{2}{c}{--} \\
& $\checkmark$ & &$\checkmark$ & & 4.06 & 4.13 & 4.09 & 4.13 & 4.13 & 4.14 & \multicolumn{2}{c}{--} & \multicolumn{2}{c}{--} & \multicolumn{2}{c}{--} \\
$\checkmark$ &$\checkmark$ & $\checkmark$ & $\checkmark$& & 3.36 & 3.67 & 3.32 & 3.65 & 3.33 & 3.66 & \multicolumn{2}{c}{--} & \multicolumn{2}{c}{--} & \multicolumn{2}{c}{--} \\
$\checkmark$ &$\checkmark$ & $\checkmark$ & $\checkmark$&$\checkmark$ & 3.26 & 3.58 & \underline{3.21} & \underline{3.53} & \textbf{3.19} & \textbf{3.52} & \multicolumn{2}{c}{--} & \multicolumn{2}{c}{--} & \multicolumn{2}{c}{--} \\
\midrule
\end{tabular}
%\vspace{-.4cm}
\end{table*}

%\vspace{-.4cm}
\begin{figure*}[t]
  \centering
  \footnotesize	
  \subfloat[]{
    \begin{minipage}[b]{0.162\linewidth} 
      \includegraphics[width=\linewidth]{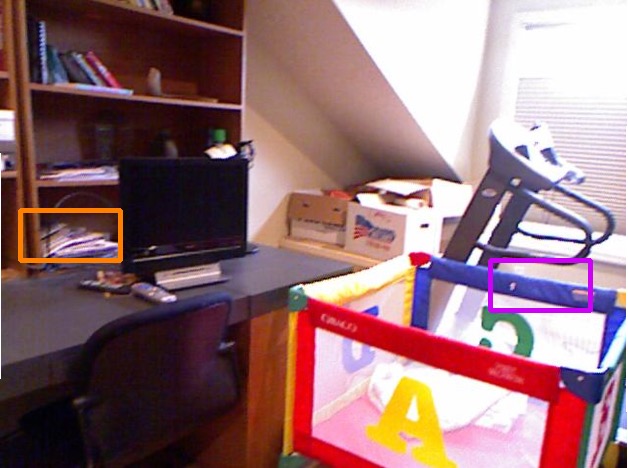}
    \end{minipage}
  }\hspace{-0.15cm}
  \subfloat[]{
    \begin{minipage}[b]{0.115\linewidth} 
      \includegraphics[width=\linewidth]{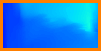}\vspace{0.05cm}
      \includegraphics[width=\linewidth]{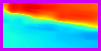}
    \end{minipage}
  }\hspace{-0.22cm}
  \subfloat[]{
    \begin{minipage}[b]{0.115\linewidth} 
      \includegraphics[width=\linewidth]{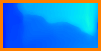}\vspace{0.05cm}
      \includegraphics[width=\linewidth]{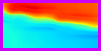}
    \end{minipage}
  }\hspace{-0.22cm}
  \subfloat[]{
    \begin{minipage}[b]{0.115\linewidth} 
      \includegraphics[width=\linewidth]{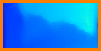}\vspace{0.05cm}
      \includegraphics[width=\linewidth]{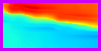}
    \end{minipage}
  }\hspace{-0.22cm}
  \subfloat[]{
    \begin{minipage}[b]{0.115\linewidth} 
      \includegraphics[width=\linewidth]{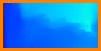}\vspace{0.05cm}
      \includegraphics[width=\linewidth]{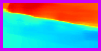}
    \end{minipage}
  }\hspace{-0.22cm}
  \subfloat[]{
    \begin{minipage}[b]{0.115\linewidth} 
      \includegraphics[width=\linewidth]{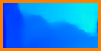}\vspace{0.05cm}
      \includegraphics[width=\linewidth]{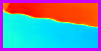}
    \end{minipage}
  }\hspace{-0.22cm}
  \subfloat[]{
    \begin{minipage}[b]{0.115\linewidth} 
      \includegraphics[width=\linewidth]{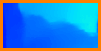}\vspace{0.05cm}
      \includegraphics[width=\linewidth]{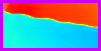}
    \end{minipage}
  }\hspace{-0.22cm}
  \subfloat[]{
    \begin{minipage}[b]{0.115\linewidth} 
      \includegraphics[width=\linewidth]{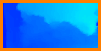}\vspace{0.05cm}
      \includegraphics[width=\linewidth]{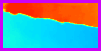}
    \end{minipage}
  }\hspace{-0.1cm}  
  \begin{minipage}[b]{0.0155\linewidth}
%  \vspace*{1cm}
  \includegraphics[width=\linewidth]{images/vert_colorbar}
  \end{minipage}
  \caption{Visual comparison of different networks~(DKN) for depth upsampling~($8\times$) using the kernel of size $3\times3$. (a) an RGB image. Results for weight regression using (b)~RGB, (c)~depth and (d)~RGB and depth images. Results for weight and offset regression using (e)~RGB, (f)~RGB and depth images, and (g)~RGB and depth images with the residual connection. (h)~ground truth.}
  \label{fig:ablation-network}
%  \vspace{-.2cm}
\end{figure*}

\begin{table}[t]
\centering
\caption{Quantitative comparison of different components for weight regression for filter kernels of size~$3\times 3$.}
\label{table:weight_regression}
\vspace{-.1cm}
\label{table:ablation-activation}
\addtolength{\tabcolsep}{-3.0pt}
\newcolumntype{C}[1]{>{\centering\arraybackslash}p{#1}}
\begin{tabular}[c]{c c c c c} 
\midrule
\multicolumn{2}{c}{Activation function} & \multirow{2}{*}{Mean subtraction} & \multirow{2}{*}{DKN} & \multirow{2}{*}{FDKN} \\
\cmidrule(lr){1-2} 
Softmax & Sigmoid \\
\midrule
& & & 3.74 & 3.98 \\
& $\checkmark$ &  & 5.72 & 5.89 \\
$\checkmark$ & & $\checkmark$ & \underline{3.49} & \underline{3.65} \\
&$\checkmark$ &  $\checkmark$ & \bf{3.26} & \bf{3.58} \\ 
\midrule
\end{tabular}
%\vspace{-.2cm}
\end{table}

\vspace{-0.3cm}
\subsubsection{Results}
\vspace{-0.2cm}
We show in Table~\ref{table:segmentation} mean intersection-over-union~(IoU) scores for the validation set in the Pascal VOC 2012 benchmark~\citep{pascalvoc2012}. To the baseline method~(DeepLab~v2~w/o CRF~\citep{chen2018deeplab}), we add CRF~\citep{krahenbuhl2011efficient}, guided filtering~\citep{wu2018fast}, or FDKN layers. This table shows that our model quantitatively yields better accuracy in terms of mean IoU than other state-of-the-art methods. Examples of semantic segmentation are shown in Fig.~\ref{fig:segmentation}. Our model outperforms other methods qualitatively as well: It improves the localization of object boundaries~(first row) and refines incorrect labels~(second row).

\vspace{-.6cm}
\section{Ablation study}\label{sec:discussion}
\vspace{-.2cm}
In this section, we conduct an ablation analysis on different components in our models, and show the effects of different parameters for depth map upsampling~($8 \times$) on the NYU v2 dataset. We also discuss other issues including kernel visualization, runtime, training loss, and upscaling factors for training and testing.

\begin{table*}[t]
\centering
%\vspace{-.3cm}
\caption{Quantitative comparison of using different number of channels for feature extraction. The results of DKN and FDKN are separated by ``/". We denote by $n_i$~($i=1,2$) the number of channels in the final two layers for feature extraction. For each network, numbers in bold indicate the best performance and underscored ones are the second best.}
\vspace{-.1cm}
\addtolength{\tabcolsep}{-3.0pt}
\newcolumntype{C}[1]{>{\raggedright\arraybackslash}p{#1}}
\begin{tabular}{l C{0.65cm}@{/} C{0.65cm} C{0.65cm}@{/} C{0.65cm} C{0.65cm}@{/} C{0.65cm} C{0.65cm}@{/} C{0.65cm} C{0.65cm}@{/} C{0.65cm}}
\midrule
\multicolumn{1}{c}{$n_1$} & \multicolumn{2}{c}{128} & \multicolumn{2}{c}{256} & \multicolumn{2}{c}{256} & \multicolumn{2}{c}{256} & \multicolumn{2}{c}{256} \\  
\cmidrule(lr){2-11} 
\multicolumn{1}{c}{$n_2$} & \multicolumn{2}{c}{128} & \multicolumn{2}{c}{128} & \multicolumn{2}{c}{256} & \multicolumn{2}{c}{512} & \multicolumn{2}{c}{1024}\\
\cmidrule{1-11}
RMSE & 3.26 & ~3.58 & 3.22 & ~3.51 & 3.20 & ~3.49 & \underline{3.17} & \underline{~3.46} & \textbf{3.15} & \textbf{~3.42} \\

Runtime~(s) & \textbf{0.17} & \textbf{~0.01} & \underline{0.20} & \textbf{~0.01} & 0.22 & \textbf{~0.01} & 0.27 & \underline{~0.02} & 0.36 & \underline{~0.02} \\ 

Number of parameter~(M) & \textbf{1.1} & \textbf{~0.6} & \underline{1.7} & \underline{~1.1} & 2.3 & ~1.7 & 3.5 & ~2.9 & 5.8 & ~5.9\\
Model size~(MB) & \textbf{4.5} & \textbf{~2.8} & \underline{6.7} & \underline{~4.5} & 9.0 & ~7.3 & 14.0 & ~13.0 & 23.0 & ~24.2 \\

\midrule
\end{tabular}
\label{table:ablation-featurechannel}
\vspace{-.2cm}
\end{table*}
\noindent{\textbf{Network architecture.}}
We show the average RMSE for six variants of our models in Table~\ref{table:ablation-archicture}. The baseline models learn kernel weights from the guidance images only. From the second row, we can see that our models trained using the target images only give better results than the baseline, indicating that using the guidance images only is not enough to fully exploit common structures. The third row demonstrates that constructing kernels from both guidance and target images boosts performance. For example, the average RMSE of DKN decreases from $5.92$ to $5.03$ for the $3\times 3$ kernel. The fourth and fifth rows show that learning the offsets significantly boosts the performance of our models. The average RMSE of DKN trained using the guidance or target images only decreases from $5.92$ to $5.37$ and from $5.24$ to $4.06$, respectively, for the $3\times 3$ kernel.  The last two rows demonstrate that the effect of learning kernel weights and offsets from both inputs is significant, and combining all components including the residual connection gives the best results. Figure~\ref{fig:ablation-network} shows a visual comparison of using different networks for depth upsampling. Note that learning to predict the spatial offsets is important because (1)~learning spatially-variant kernels for individual pixels would be very hard otherwise, unless using much larger kernels to achieve the same neighborhood size, which would lead to an inefficient implementation, and (2) contrary to current architectures including DJF~\citep{li2016deep}, PAC~\citep{su2019pixel} and DMSG~\citep{hui2016depth}, this allows sub-pixel information aggregation. 

Our models use a two-stream network to extract feature maps from the guidance and target images. We can regress the weights and offsets with concatenated guidance and target images passed to a single network. This model, however, gives worse errors than our two-stream DKN. In particular, the RMSE increases from 3.26 to 3.50 for filter kernels of size~$3 \times 3$. This suggests that the different feature maps from the two-stream architecture are better to estimate the kernel weights and offsets. A similar finding is noted in DJF~\citep{li2016deep}.

Our models use sigmoid and mean subtraction layers for weight regression. We could use a softmax layer that makes all elements larger than 0 and smaller 1 as in the sigmoid layer. The sigmoid and mean subtraction layer can be replaced with a single linear one. We compare in Table~\ref{table:weight_regression} the performance of DKN and FDKN with softmax or sigmoid layers, and when using mean subtraction or not. This demonstrates that 1)~the softmax layer does not perform as well as the sigmoid layer, and 2) constraints on weight regression using sigmoid and mean subtraction layers give better results.

\begin{table}[t]
\centering
\caption{Runtime comparison for images of size~$640 \times 480$ on the NYU v2~\citep{silberman2012indoor} dataset. $\dagger$: Our models trained with the depth map only without any guidance.}
\label{table:runtime}
\addtolength{\tabcolsep}{-3.0pt}
%\scriptsize
\begin{tabular}[c]{l c c}
\midrule
\multicolumn{1}{c}{Methods} & GPU Times(s) & CPU Times (s) \\
\midrule
MRF~\citep{diebel2006application} & -- & 0.69 \\
GF~\citep{he2013guided} & -- & \textbf{0.14} \\
JBU~\citep{kopf2007joint} & -- & \underline{0.31} \\
TGV~\citep{ferstl2013} & -- & 33 \\
Park~\citep{park2011} & -- &17 \\
SDF~\citep{ham2018robust} & -- & 25 \\
FBS~\citep{barron2016fast} & -- & 0.37 \\
\midrule
DMSG~\citep{hui2016depth} & 0.04 & 1.4 \\
DJFR~\citep{li2017joint} & \textbf{0.01} & 1.3 \\
PAC~\citep{su2019pixel} & \underline{0.03} & 1.4 \\
DKN$^\dagger$ & 0.09 & 3.2 \\
FDKN$^\dagger$ & \textbf{0.01} & 0.9 \\
DKN & 0.17 & 5.4 \\
FDKN & \textbf{0.01} & 1.0 \\  
\midrule
  \end{tabular}
 \vspace{-.1cm}
\end{table}

\noindent{\textbf{Kernel size.}}
Table~\ref{table:ablation-archicture} also compares the performances of networks with different size of kernels. We enlarge the kernel size gradually from $3 \times 3$ to $25 \times 25$ and compute the average RMSE. From the third row, we observe that the performance improves until size of $15 \times 15$. Increasing size further does not give additional performance gain. This indicates that aggregating pixels from a $15 \times 15$ window is enough for the task. For offset learning, we restrict the maximum range of the sampling position to $15\times 15$ for all experiments. That is, the filtering results from the third to last rows are computed by aggregating 9, 25 or 49 samples sparsely chosen from a $15 \times 15$ window. The last row of Table~\ref{table:ablation-archicture} suggests that our final models also benefit from using more samples. The RMSE for DKN decreases from $3.26$ to $3.19$ at the cost of additional runtime. For comparison, DKN with kernels of size $3 \times 3$, $5 \times 5$ and $7 \times 7$ take 0.17, 0.18 and 0.19 seconds, respectively, with a Nvidia Titan XP. A $3 \times 3$ size offers a good compromise in terms of RMSE and runtime and this is what we have used in all experiments.

\noindent{\textbf{Feature channels.}}
In Table~\ref{table:ablation-featurechannel}, we compare the effects of the number of feature channels in terms of RMSE, runtime, the number of network parameters, and model size. We use our DKN and FDKN models including the residual connection and a fixed size of~$3 \times 3$ kernels. We vary the number of channels $n_i$~($i=1,2$) in the final two layers for feature extraction (see Tables~\ref{table:architecture} and~\ref{table:fastarchitecture}). The table shows that using more channels for feature extraction helps improve performance, but requires more runtime and a large number of parameters to be learned. For example, DKN takes twice more time for a (modest)~$0.11$ RMSE gain. Consequently, we choose the number of feature channels $n_1=128$ and $n_2=128$ for both models.

\begin{figure}[t]
\centering
\includegraphics[width=0.95\linewidth]{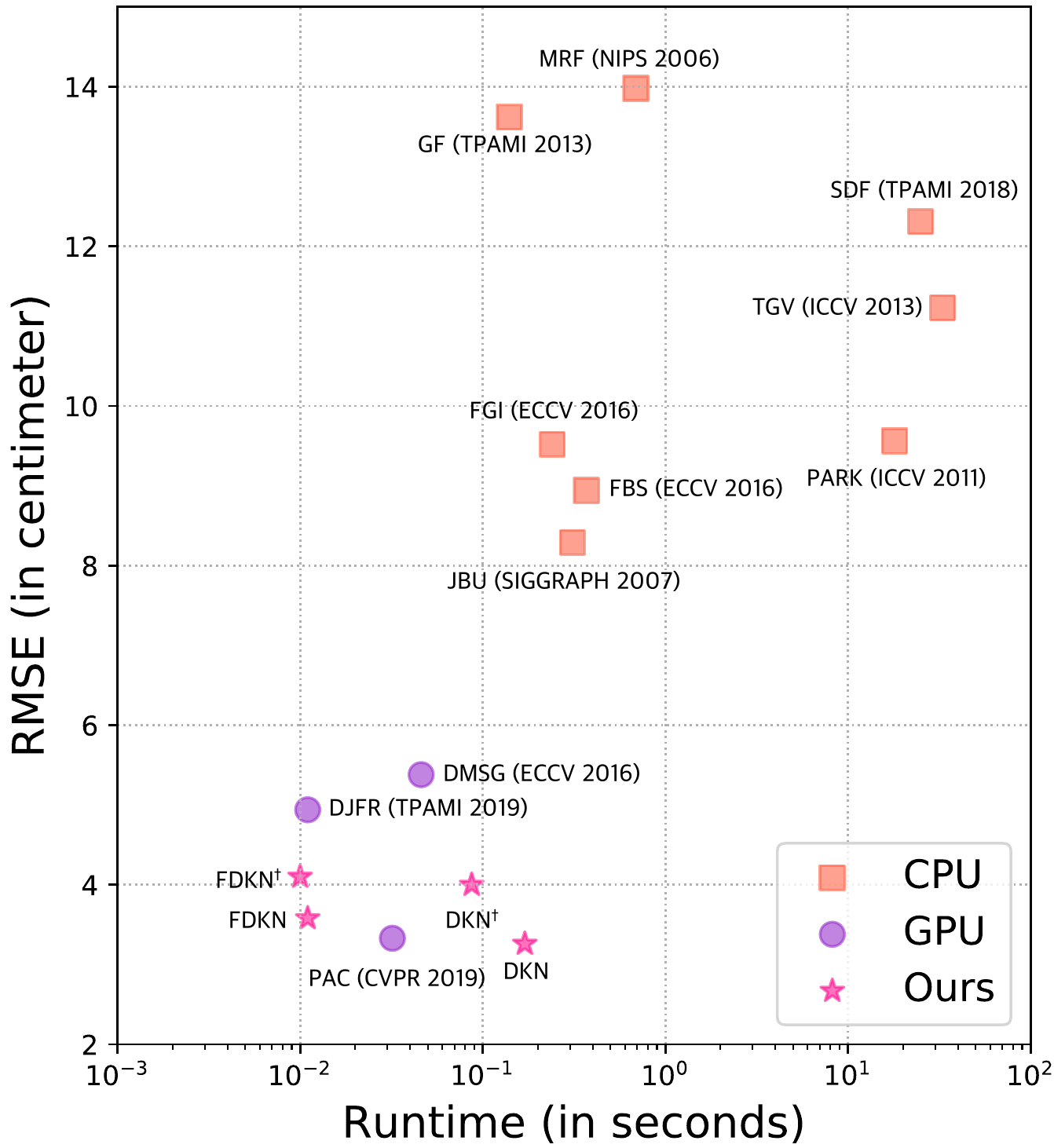}
\caption{Runtime and root mean squared errors~(RMSE) comparison of upsampled depth maps~($8 \times$) on the NYU v2~\citep{silberman2012indoor} dataset. $\dagger$: Our models trained with the depth map only without any guidance.}
\label{fig:timegraph}
\vspace{-.1cm}
\end{figure}

\noindent{\textbf{DownConv for DKN.}}
We empirically find that extracting features from large receptive fields is important to incorporate context for weight and offset learning. For example, reducing the size from $51 \times 51$ to $23 \times 23$ causes an increase of the average RMSE from $3.26$ to $5.00$ for the $3\times 3$ kernel. The DKN without DownConv layers can be implemented in a single forward pass, but requires more parameters~($1.6$M vs. $1.1$M for DKN) to maintain the same receptive field size, with a total number of convolutions increasing from $0.6$M to $1$M at each pixel. We may use dilated convolutions~\citep{yu2016multi} that support large receptive fields without loss of resolution. When using the same receptive field size as $51 \times 51$, the average RMSE for dilated convolutions increases from $3.26$ to $4.30$ for the $3\times 3$ kernel. We can also use deconvolutional layers, similar to the U-Net architecture~\citep{ronneberger2015u} for upsampling. They, however, produce checkerboard artifacts~\citep{odena2016deconvolution}, which may degrade the performance. The resampling technique~(Fig.~\ref{fig:resampling}) thus appears to be the preferable alternative.

\begin{table}[t]
\centering
\caption{RMSE comparison of using different upscaling factors for training and testing on depth map upsampling.}
\label{table:limitation}
\addtolength{\tabcolsep}{-2.0pt}
\begin{tabular}[c]{c r r r} 
\midrule
Train/Test& \multicolumn{1}{c}{$4\times$}  & \multicolumn{1}{c}{$8\times$} & \multicolumn{1}{c}{$16\times$}\\
\cmidrule{1-4}
$4\times$ & {\textbf{1.62}} & 6.70 &  11.24 \\
$8\times$ & 3.93 & {\textbf{3.26}} & 10.53  \\ 
$16\times$ & 9.04  & 8.61 & {\textbf{6.51}} \\ 
\midrule
\end{tabular}
%\vspace{-.2cm}
\end{table}

\noindent{\textbf{Training loss.}}
Average RMSEs for L1, L2 and L1+perceptual losses are 3.58, 3.69, and 3.66, respectively, for the task of depth upsampling~($8 \times $) on the NYU v2 dataset. The L1 loss is robust to outliers, preserving depth boundaries better than the L2 one. The perceptual loss, typically using a network~\citep{Simonyan15} pretrained for the ImageNet classification, does not provide a gain on the RMSE performance.

\noindent{\textbf{Runtime.}} 
Our network consists of two parts:~1) feature extraction and weight \& offset regression, and 2) weighted average~(Fig.~\ref{fig:overview}). The second part takes less than 0.001s for all models. The first part for DKN, DKN$^\dagger$, FDKN, FDKN$^\dagger$ takes 0.165s, 0.086s, 0.011s, 0.01s respectively for images of size~$640 \times 480$. For comparison with other methods, table~\ref{table:runtime} shows runtime on the same machine. We report the GPU runtime for DMSG~\citep{hui2016depth}, DJFR~\citep{li2017joint}, PAC~\citep{su2019pixel}, and our models with a Nvidia Titan XP. DKN is slower than DMSG~\citep{hui2016depth}, PAC~\citep{su2019pixel} and DJFR~\citep{li2017joint}, but yields a significantly better RMSE~(Fig.~\ref{fig:timegraph} and Table~\ref{table:depth-upsampling}). FDKN runs about $17 \times$ faster than the DKN, as fast as DJFR, but with significantly higher accuracy. We also report the CPU runtime with an Intel i5 3.3 GHz, demonstrating that FDKN is as fast or faster than other CNN-based methods, even on CPUs.

\noindent{\textbf{Upscaling factors for training and testing.}}
Table~\ref{table:limitation} compares the average RMSE on the NYU dataset~\citep{silberman2012indoor}, when the scale factors for training and test are different. It shows that the performance is degraded. This may be handled by a scale augmentation technique during training~\citep{kim2016accurate}. A visual comparison is shown in Fig.~\ref{fig:limitation-scalefactor}.

\begin{figure}[t]
  \centering
  \footnotesize		
  \subfloat[RGB image.]{
    \begin{minipage}[b]{0.29\linewidth} 
      \includegraphics[width=\linewidth]{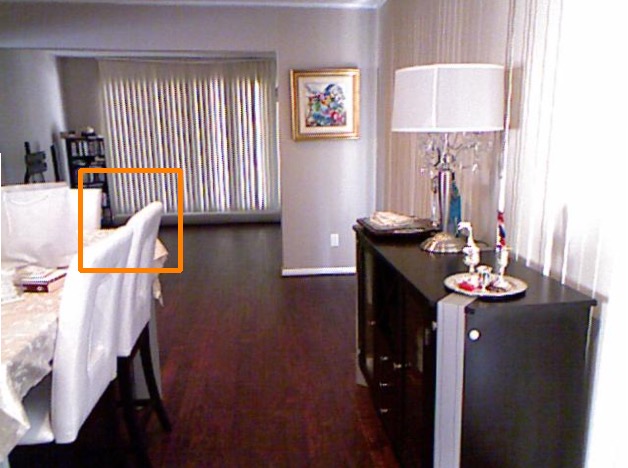}
    \end{minipage}
  }\hspace{-0.2cm}
  \subfloat[$\times 4$.]{
    \begin{minipage}[b]{0.215\linewidth} 
      \includegraphics[width=\linewidth]{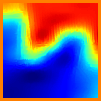}
    \end{minipage}
  }\hspace{-0.15cm}
  \subfloat[$\times 8$.]{
    \begin{minipage}[b]{0.215\linewidth} 
      \includegraphics[width=\linewidth]{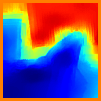}
    \end{minipage}
  }\hspace{-0.15cm}
  \subfloat[$\times 16$.]{
    \begin{minipage}[b]{0.215\linewidth} 
      \includegraphics[width=\linewidth]{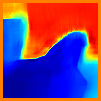}
    \end{minipage}
  }
  \begin{minipage}[b]{0.027\linewidth}
  \includegraphics[width=\linewidth]{images/vert_colorbar}
  \end{minipage}
  \caption{Visual comparison of upsampled depth images for DKN when the scale factors for training~$(\times 16)$ and test~($\times 4, \times 8, \times 16$) are different.}

  \label{fig:limitation-scalefactor}
\end{figure}

\begin{figure*}[t]
  \centering
  \subfloat{
    \begin{minipage}[b]{0.3\linewidth}
      \includegraphics[width=\linewidth]{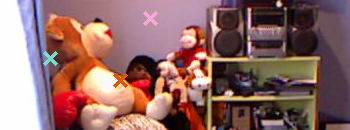} 
      \includegraphics[width=\linewidth]{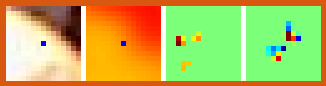} 
    \end{minipage}
  }\hspace{-0.2cm}
  \subfloat{
    \begin{minipage}[b]{0.3\linewidth}
      \includegraphics[width=\linewidth]{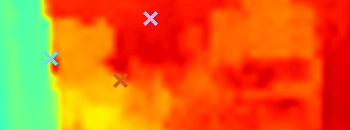} 
      \includegraphics[width=\linewidth]{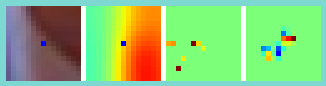} 
    \end{minipage}
  }\hspace{-0.2cm}
  \subfloat{
    \begin{minipage}[b]{0.3\linewidth}
      \includegraphics[width=\linewidth]{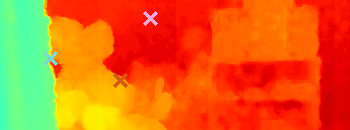} 
      \includegraphics[width=\linewidth]{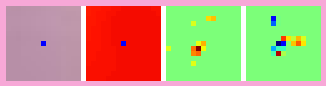} 
    \end{minipage}
  }\hspace{-0.2cm}
  \subfloat{
    \begin{minipage}[b]{0.046\linewidth}
      \includegraphics[width=\linewidth]{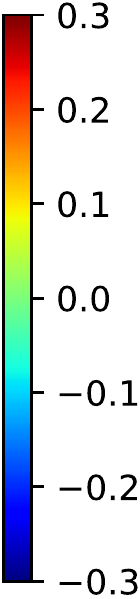}
    \end{minipage}
  }\hspace{-0.2cm}
  \caption{Visualization of filter kernels. Top: (From left to right) an RGB image, a low-resolution depth image, and an upsampling result by DKN. Bottom: (From left to right) Snippets of RGB images, low-resolution depth images, and kernels learned w/o and w/ the residual connection. The center positions in the RGB and depth images are denoted by blue dots. The kernel weights are plotted with a heat map.~(Best viewed in color.)}
%  \vspace{-.2cm}
  \label{fig:visualkernel}
\end{figure*}

\begin{table}[t]
\centering
\caption{RMSE comparison by varying the number of training data on depth map upsampling~($8 \times$).}
\label{table:trainingnumber}
\addtolength{\tabcolsep}{-3.0pt}
\begin{tabular}[c]{l l c c c c c c c}
\midrule
\multicolumn{1}{c}{Datasets} & \multicolumn{1}{c}{Methods}& 10 & 50 & 100 & 200 & 500 & 700 & 1000 \\
\cmidrule(lr){1-9}
\multirow{2}{*}{NYU v2} & DMSG & 7.40 &6.32 &5.97 &5.64 &5.41 &5.35 &5.38 \\
& FDKN & 5.01 & 4.28 & 3.98 & 3.77 & 3.61 & 3.60 & 3.58\\
& DKN & 4.73 & 3.97 & 3.62 & 3.33 & 3.27 & 3.25 & 3.26 \\
\cmidrule(lr){1-9}
\multirow{2}{*}{Sintel}  & DMSG & 8.62 & 8.08 &7.65 &7.46 &7.27 &7.21 &7.24\\
& FDKN & 6.04 & 5.37 & 5.16 & 5.05 & 4.98 & 4.98 & 4.96 \\
& DKN & 5.79 & 5.24 & 5.01 & 4.85 & 4.82 & 4.74 & 4.77 \\
  \midrule
  \end{tabular}
%  \vspace{-.3cm}
\end{table}

\noindent{\textbf{Kernel prediction vs. direct regression.}}
Our model has several advantages over current CNN-based approaches that directly regress the filtering output. First, the direct regression may overfit the particular characteristics of training data, especially when the number of training samples is small. In contrast, weighted averaging smooths the output and acts as a regularizer, suggesting that our model is not seriously affected by the number of training samples. To demonstrate this, we evaluate the average RMSE performance in Table~\ref{table:trainingnumber}, when varying the size of the training data. We train the DKN, FDKN and DMSG~\citep{hui2016depth}, where the filtering output is directly regressed from input images, for depth map upsampling~($8 \times$) while gradually increasing the number of training samples from $10$ to $1,000$ in the NYU v2 dataset~\citep{silberman2012indoor}. We test them in the same configuration as in Table~\ref{table:depth-upsampling}. 

Table~\ref{table:trainingnumber} shows that our models are more robust to the size of training data and generalize better to other images~(e.g.,~on the Sintel dataset~\citep{butler2012sintel}) outside the training dataset than the direct regression approach, even with more learnable parameters~(1.1M for DKN and 0.6M for FDKN vs. 0.43M for DMSG~\citep{hui2016depth}). In particular, the DKN trained with only $10$ images outperforms the state of the art by a significant margin for all test datasets~(see Table~\ref{table:depth-upsampling}). Second, the kernels learned by direct regression are defined implicitly and hard to visualize. In contrast, our method learns sparse kernels~(i.e.,~\emph{where} to aggregate) explicitly. We can interpret and visualize why kernels learned by our model give smooth results while preserving edges~(Fig.~\ref{fig:visualkernel}), and this also gives a clue for tuning hyper-parameters. For examples, we can reduce the maximum range of offset locations~(i.e.,~the size of the filter kernel) and the number of weights~(i.e.,~the total number of samples to aggregate), when the weights are concentrated on central parts of the kernels, and a few of them are highly confident, respectively. Note that sparsely aggregating sub-pixel information is not feasible for direct regression approaches~(e.g.,~DMSG~\citep{hui2016depth} and DJF~\citep{li2016deep}). Finally, our model can be applied to any tasks requiring an explicit weighted averaging processing beyond (joint)~image filtering, as confirmed for the task of semantic segmentation in Section~\ref{exp:ss}.

%\vspace{-0.4cm}
\noindent{\textbf{Kernel visualization.}}
We show in Fig.~\ref{fig:visualkernel} some examples of $3\times 3$ filter kernels estimated by the DKN with/without the residual connection. Although the sampling positions are fractional, we plot them on a discrete regular grid using bilinear interpolation for the purpose of visualization. Corresponding kernel weights are also interpolated. We observe three things: (1)~The learned kernels are spatially adaptive and edge-aware. For example, the kernels learned without the residual connection aggregate depth values that are similar to that at the center position. Note that nearby pixels have lower weights than non-neighboring ones especially at depth boundaries, as they are blurred in the low-resolution depth image. This suggests that structural details are also related to further away pixels.  A similar finding is noted in nonlocal means~\citep{buades2005non}.~(2) They can handle the case when the structures from the guidance and target images are not consistent as shown in the second example. (3)~The kernels learned with the residual connection are orientation-selective and look like high-pass filters. For example, the kernels from the first and second examples can extract diagonal and vertical edges, respectively. 

\vspace{-.3cm}
\section{Conclusion}
\vspace{-.2cm}
We have presented a CNN architecture for joint image filtering that is generic and applicable to a great variety of applications. Instead of regressing the filtering results directly from the network, we use spatially-variant weighted averages where the set of neighbors and the corresponding kernel weights are learned end-to-end in a dense and local manner. We have also presented an efficient implementation that gives much faster runtime than the brute-force one. A fast version further achieves an additional $17 \times$ speed-up without much~(if any) loss in performance. Our models generalize well to images that have different modalities from the training dataset, as demonstrated by our experiments. Finally, we have shown that the weighted averaging process with sparsely sampled $3 \times 3$ kernels is sufficient to set new state-of-the-art results on several tasks. In future work, we will explore network architectures for sparse-to-dense interpolation such as depth completion~\citep{tang2019learning} and optical flow propagation~\citep{revaud2015epicflow}.

\begin{figure*}[t]
\centering
\includegraphics[width=0.9\linewidth]{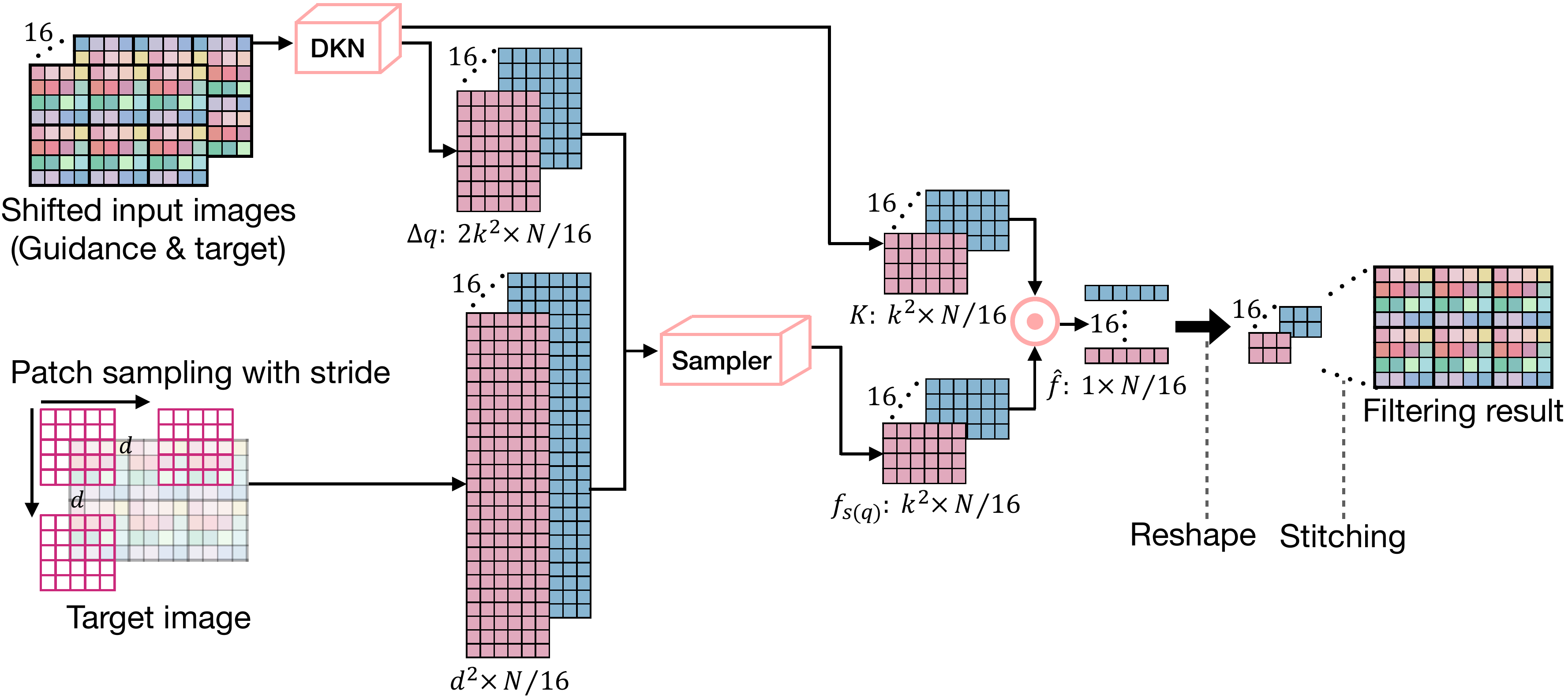}
\caption{Efficient implementation using a shift-and-stitch approach. We denote $d$ by the maximum range of the sampling location~${{\bf{s}}({\bf{q}})}$. We shift the input images and compute the filtering result for each shifted input. We then stitch them up to get a filtering result that has the same resolution as the inputs. Our approach makes it possible to reuse the storage for kernel weights, offsets, and resampled pixels. See text for details.~(Best viewed in color.)}
\label{fig:efficient_implementation}
\end{figure*}

%\appendix
%\vspace{-0.3cm}
\begin{appendix}
\section{Appendix: Efficient implementation}\label{app:implementation}
 We use the shift-and-stitch approach~\citep{long2015fully,niklaus2017video} that stitches the network outputs from shifted versions of the input~(Fig.~\ref{fig:efficient_implementation}). We can obtain the same result as the pixel-wise implementation in 16 forward passes. We first shift input images $x$ pixels to the left and $y$ pixels up, once for every $(x,y)$ where $\{ (x,y) | 0 \leq x,y \leq 3 \}$, and obtain a total of 16 shifted inputs. Each shifted input goes through the network that gives the kernel weights $K$ and the offsets $\Delta {\bf{q}}$ of size $k^2 \times N / 16$ and $2k^2 \times N / 16$, respectively. The next step is to obtain image values~$f_{{\bf{s}}({\bf{q}})}$ using the sampling function~${{\bf{s}}({\bf{q}})}$ from the target image. To this end, starting from every location $(x,y)$ in the target image, we sample patches of size~$d \times d$ with stride 4 in each dimension, each of which gives the output of size~$d^2 \times N/16$. The patch size corresponds to the maximum range of the sampling position~${{\bf{s}}({\bf{q}})}$. For an efficient implementation, we restrict the range~(e.g., to~$15\times 15$ in our experiment). We then sample $k^2$~pixels using the sampling position~${{\bf{s}}({\bf{q}})}$ from patches of size~$d \times d$, obtaining~$f_{{\bf{s}}({\bf{q}})}$~of size~$k^2 \times N / 16$~for each shifted input. To compute a weighted average, we apply element-wise multiplication between the kernel weights~$K$~and the corresponding sampled pixels~$f_{\bf{s}({\bf{q}})}$~of size~$k^2 \times N / 16$~followed by column-wise summation, resulting in an output of size $1 \times N / 16$. Finally, we stitch 16 outputs of size $1 \times N / 16$ into a single one to get the final output. Note that one can stitch kernel weights and offsets first and then compute a weighted average, but this requires a large amount of memory. We stitch instead the outputs after the weighted average, and reuse the storage for kernel weights, offsets, and sampled pixels. 

	\end{appendix}

\begin{acknowledgements}
 The authors would like to thank Yijun Li for helpful discussion. This work was supported in part by Samsung Research Funding \& Incubation Center for Future Technology (SRFC-IT1802-06), the Louis Vuitton/ENS chair on artificial intelligence, the Inria/NYU collaboration agreement, and the French government under management of Agence Nationale de la Recherche as part of the ``Investissements d'avenir" program, reference ANR-19-P3IA-0001 (PRAIRIE 3IA Institute).
\end{acknowledgements}


% TODO: bib style need to be changed
\bibliographystyle{spbasic}      % basic style, author-year citations
\bibliography{egbib} % name your BibTeX data base

\end{document}